\documentclass[10pt,twocolumn,letterpaper]{article}

\usepackage[pagenumbers]{cvpr} %

\usepackage{xcolor}

\definecolor{sharedfill}{HTML}{DCE8F7}   %
\definecolor{privatefill}{HTML}{FBE3CF}  %
\definecolor{frozenfill}{HTML}{ECECEC}   %
\definecolor{clevr4}{HTML}{004C99}

\newcommand{\fillbox}[2]{%
\mathord{\setlength{\fboxsep}{1pt}\colorbox{#1}{$\texttt{#2}$}}}
\newcommand{\Encoder}{\fillbox{frozenfill}{Encoder}}
\newcommand{\AdapterS}{\fillbox{sharedfill}{Adapter}}
\newcommand{\AdapterW}{\fillbox{privatefill}{Adapter}}
\newcommand{\DecoderA}{\fillbox{privatefill}{Decoder}}
\newcommand{\DecoderW}{\fillbox{privatefill}{Decoder}}

\usepackage{multirow}
\usepackage[table]{xcolor}
\usepackage{graphicx}
\usepackage{subcaption}
\usepackage{wrapfig}

\definecolor{ourspink}{RGB}{253,235,244}
\definecolor{ourpurple}{RGB}{234, 224, 245}
\definecolor{upgreen}{RGB}{55,160,55}
\definecolor{downred}{RGB}{190,0,0}

\newcommand{\up}{\textcolor{upgreen}{$\bm{\uparrow}$}}
\newcommand{\down}{\textcolor{downred}{$\bm{\downarrow}$}}
\newcommand{\optarrow}{\textcolor{gray}{$\boldsymbol{\swarrow}$}}

\usepackage{amsfonts} 
\usepackage{mathrsfs}

\DeclareRobustCommand{\loss}{%
  {\fontencoding{OT1}\fontfamily{cmtt}\fontseries{m}\selectfont AnchorNCE}%
}

\DeclareRobustCommand{\method}{%
  {\fontencoding{OT1}\fontfamily{cmtt}\fontseries{m}\selectfont FAnchor}%
}

\usepackage[cmttb]{bold-extra}

\DeclareRobustCommand{\methodb}{%
  {\fontencoding{OT1}\fontfamily{cmtt}\bfseries\selectfont FAnchor}%
}

\DeclareRobustCommand{\lossb}{%
  {\fontencoding{OT1}\fontfamily{cmtt}\bfseries\selectfont AnchorNCE}%
}

\usepackage{lipsum}

\usepackage[most]{tcolorbox}
\newtcolorbox{takeaway}{
  colback=ourpurple!35,
  colframe=ourpurple!70!violet,
  boxrule=1.2pt,
  arc=2pt,
  left=4pt, right=4pt, top=3pt, bottom=3pt
}

\usepackage{amsmath,mathtools,amsfonts,bm, amssymb}

\def\eqref#1{equation~\ref{#1}}

\def\1{\bm{1}}

\def\rva{{\mathbf{a}}}

\def\rvd{{\mathbf{d}}}
\def\rve{{\mathbf{e}}}
\def\rvf{{\mathbf{f}}}
\def\rvg{{\mathbf{g}}}
\def\rvh{{\mathbf{h}}}
\def\rvu{{\mathbf{i}}}

\def\rvs{{\mathbf{s}}}

\def\rvu{{\mathbf{u}}}
\def\rvv{{\mathbf{v}}}
\def\rvw{{\mathbf{w}}}
\def\rvx{{\mathbf{x}}}

\def\rvz{{\mathbf{z}}}

\def\rmX{{\mathbf{X}}}

\def\vtheta{{\bm{\theta}}}

\DeclareMathAlphabet{\mathsfit}{\encodingdefault}{\sfdefault}{m}{sl}
\SetMathAlphabet{\mathsfit}{bold}{\encodingdefault}{\sfdefault}{bx}{n}

\def\sR{{\mathbb{R}}}

\newcommand{\Ls}{\mathcal{L}}

\newtheorem{theorem}{Theorem}

\newtheorem{proposition}[theorem]{Proposition}

\newtheorem{definition}{Definition}

\newtheorem{assumption}{Assumption}

\usepackage{acronym}
\acrodef{nn}[NN]{Neural Network}
\acrodef{poe}[PoE]{Product-of-Experts}
\acrodef{moe}[MoE]{Mixture-of-Experts}
\acrodef{sota}[SOTA]{state-of-the-art}
\acrodef{cl}[CL]{contrastive learning}
\acrodef{vae}[VAE]{variational autoencoder}
\acrodef{moe}[MoE]{Mixture of Experts}
\acrodef{slln}[SLLN]{Strong Law of Large Numbers}
\acrodef{dct}[DCT]{Dominated Convergence Theorem}
\acrodef{mni}[MNI]{Minimum Necessary Information}
\acrodef{vlm}[VLM]{language-aligned vision model}

\def\vtheta{{\boldsymbol{\theta}}}
\def\vphi{{\boldsymbol{\phi}}}
\def\sR{{\mathbb{R}}}

\definecolor{cvprblue}{rgb}{0.21,0.49,0.74}
\usepackage[pagebackref,breaklinks,colorlinks,allcolors=cvprblue]{hyperref}

\def\confName{CVPR}
\def\confYear{2027}

  \title{\emph{Aligned, yet distinct}: Shared–Private Factorization of Multimodal Representations with Formal Guarantees}

\author{
Huyen Vo$^{1,2}$\thanks{Equal contribution. 
} \quad
Mar\'{i}a Mart\'{i}nez-Garc\'{i}a$^{1}$\footnotemark[1] \quad
Jonas Fischer$^{3}$ \quad
Isabel Valera$^{1,2}$\\[4pt]
{ $^{1}$Department of Computer Science, Saarland University, Germany}\\
{ $^{2}$Max Planck Institute for Software Systems, Saarland Informatics Campus, Germany}\\
{ $^{3}$Max Planck Institute for Informatics, Saarland Informatics Campus, Germany}
}

\begin{document}
\maketitle
\addtocontents{toc}{\protect\setcounter{tocdepth}{-10}} %

\begin{abstract}

Factorizing multimodal representations into shared and private (\ie, modality-specific) information requires capturing \emph{all and only} cross-modal invariant information in the shared component, while preserving the remainder in private components.
We introduce \method{}, an end-to-end framework that achieves this factorization using lightweight adapters on pretrained embeddings and two complementary objectives.
 First, \loss{ } aligns shared representations to a common anchor defined by their arithmetic mean, scaling linearly with the number of modalities. %
 Importantly,  it formally guarantees  block-identifiability of the shared information, and allows for a fine-grained control of private-to-shared  leakage with a single hyperparameter.
 Second, the private representations are optimized using a Matryoshka-like reconstruction loss with two terms: one that prioritizes reconstruction from the shared anchor alone, and another that refines the reconstruction by capturing modality-specific information in the private representations.
Across controlled and real-world benchmarks, \method{} achieves competitive or superior information preservation--leakage trade-offs over state-of-the-art methods, while improving robustness by nearly halving the retrieval rate of adversarial samples on RoCOCO. %
Our code for this work is available at \href{https://github.com/vothuckhanhhuyen/fanchor}{github.com/vothuckhanhhuyen/fanchor}.

\end{abstract}

\section{Introduction}

Self-supervised \ac{cl} underpins multimodal models such as CLIP~\citep{radford2021learning} and ALIGN~\citep{jia2021scaling}, which align representations from large-scale image--text pairs. 
By capturing information shared across modalities,
these methods implicitly assume \textit{multi-view redundancy}: that shared information suffices for downstream tasks \citep{Sridharan2008AnIT, tosh2021contrastive, tsai2021selfsupervised}. Relatedly,  the \textit{Platonic representation hypothesis} \citep{huh2024position} suggests that independently learned representations converge toward a common structure \citep{maniparambil2024vision, schnaus2026shared}. Yet, modalities also contain %
modality-specific information relevant
for prediction and generation, but not captured by shared representations \citep{tsai2018learning, liang2023factorized, tjandrasuwita2025understanding}. This motivates factorizing multimodal representations into shared and private components %
to distinguish their contributions to downstream tasks \citep{rizou2026repercent} and improve robustness to modality-specific perturbations~\citep{van2023aspnet, wang2025an}. %

Effective factorization requires shared representations to capture \emph{all and only} information common across modalities, leaving modality-specific information to private representations. Violating either 
causes \emph{shared-to-private} or \emph{private-to-shared leakage}, respectively. Yet, identifying 
information shared across \emph{all} modalities is challenging. Pairwise contrastive objectives~\citep{tian2020contrastive} scale quadratically with the number of modalities, while fixed-anchor approaches~\citep{zhu2024languagebind, girdhar2023imagebind, ruan2023accommodating} may overlook information shared among non-anchor modalities~\citep{jeong2024anchors}. Geometric approaches based on embedding volumes or areas~\citep{cicchetti2025gramian, cicchetti2026triangle} likewise either incur quadratic costs or support only a fixed number of modalities.

Existing \emph{self-supervised factorization} methods use infor\-mation-theoretic objectives~\citep{liang2023factorized, wang2025an, rizou2026repercent} and rely on task-preserving data augmentations. While the \ac{mni} criterion seeks to capture exactly the shared information~\citep{fischer2020conditional}, entanglement with private information imposes a \emph{trade-off between information preservation and leakage reduction}~\citep{wang2025an, rizou2026repercent}. Moreover, they are restricted to two modalities~\citep{liang2023factorized, wang2025an} or pairwise comparisons~\citep{rizou2026repercent}, failing to capture globally shared information.

\vspace{0.07in}
\noindent \textbf{Contributions.} We introduce \methodb{} (\textbf{F}actorized \textbf{Anchor}NCE), an end-to-end framework that factorizes pretrained multimodal embeddings into shared and private representations using lightweight adapters and two complementary objectives. Our contributions are threefold:
\begin{enumerate}[(1)]
    \item \textbf{Shared representations.} We propose \loss{}, a contrastive objective that aligns all modalities with a common anchor defined by their arithmetic mean. It scales linearly with the number of modalities and controls private-to-shared information leakage through a single hyperparameter (\Cref{sec:shared}).
    \item \textbf{Private representations.} We introduce a Matryoshka-inspired reconstruction objective that first prioritizes reconstruction from the shared anchor and then refines it using modality-specific information from the private representations (\Cref{sec:private}).
    \item \textbf{Theoretical guarantees.} We derive an asymptotic decomposition of \loss{} into alignment and uniformity terms and establish block-identifiability of shared information for any number of modalities (\Cref{sec:formal_guarantees}).
\end{enumerate}
\vspace{0.055in}
\noindent We empirically validate these contributions (\Cref{sec:exp_results}). On Clevr-4 and VGGSounder, \method{} achieves competitive or superior trade-offs between information preservation and leakage reduction compared to \ac{sota} factorization methods. On RoCOCO image-to-text retrieval, its shared representations learned with \loss{} retrieve adversarial samples 2.5--3.4 times less frequently than CLIP with fine-tuned projection heads, demonstrating improved robustness. Beyond two modalities, \loss{} remains competitive with \ac{sota} multimodal contrastive objectives while scaling linearly with the number of modalities.

\section{Background and Related work}
\begin{figure*}[t]
  \centering
  \includegraphics[width=\textwidth]{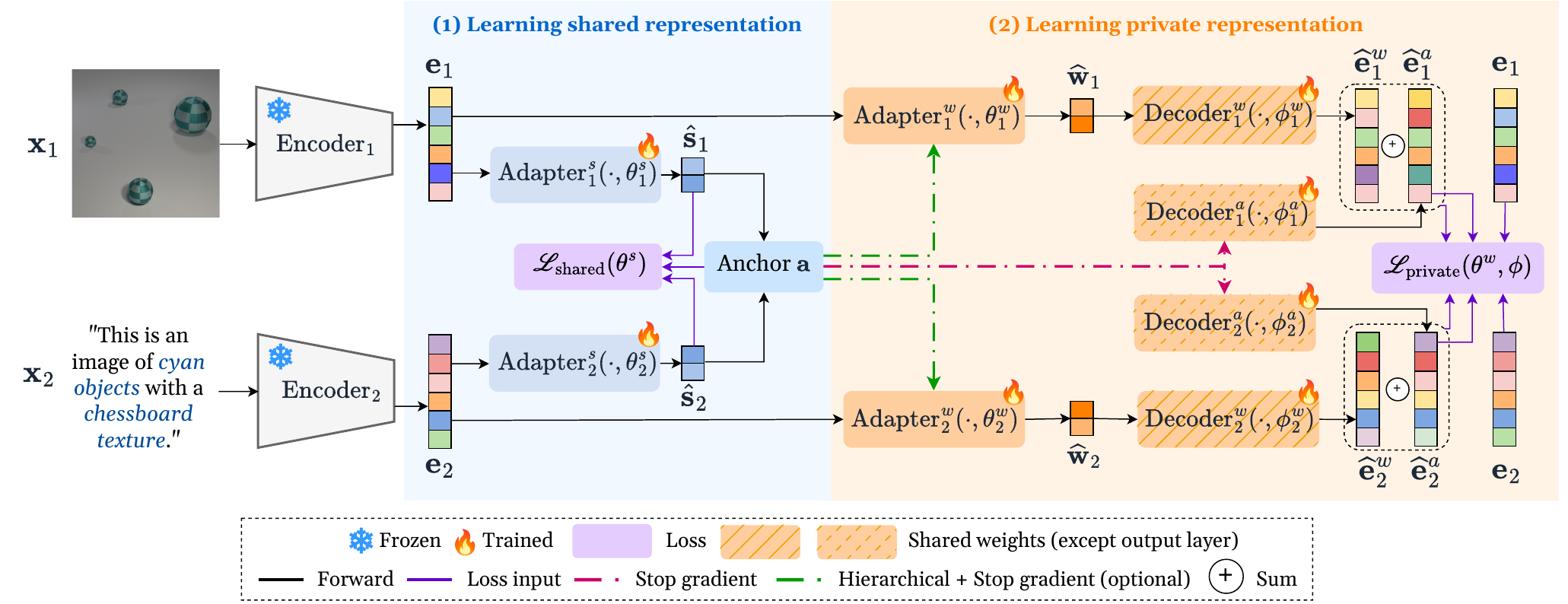}
    \caption{\method{} \emph{overview}.
    Frozen pretrained encoders map the inputs $\rvx_m$, $m \in \{1,2\}$ (image and text), to embeddings $\rve_m$. 
    \textbf{(1) Shared representations.} A shared adapter maps $\rve_m$ to a shared representation $\hat{\rvs}_m$, learned via our contrastive loss $\mathcal{L}_\text{shared} = \mathcal{L}_\text{AnchorNCE}$ in~\cref{eq:anchorloss}. 
    \textbf{(2) Private representations.} A private adapter maps $\rve_m$ to a private representation $\hat{\rvw}_m$, learned via our Matryoshka-like reconstruction loss $\mathcal{L}_\text{private} = \mathcal{L}_\text{rec}$ in~\cref{eq:loss_rec}. 
    Ideally, both shared representations should capture \emph{cyan color} and \emph{chessboard texture}, the image's private representation \emph{sphere shape}, and the text's should remain \emph{empty}.
    The architecture \emph{generalizes} to $M \geq 2$ modalities.
    }
  \label{fig:architecture}
  \vspace{-0.1in} %
\end{figure*}

\subsection{Contrastive learning and identifiability}
\label{sec:background}

\ac{cl} methods learn representations that capture information shared, or invariant, across positive pairs. Thus, the choice of positive and negative pairs determines what information is encoded in the representations. Let $\mathbf{X}^i=(\rvx^i_1,\ldots,\rvx^i_M)$ denote a sample of $M$ modalities. For $m\neq n$, positive pairs $(\rvx^i_m,\rvx^i_n)$ are drawn from the joint distribution $p(\rvx_m,\rvx_n)$, representing the same object or scene, while negative pairs $(\rvx^i_m,\rvx^j_n)$ represent different objects or scenes~\citep{ident_multimodal}. %

A popular objective for \ac{cl} is the InfoNCE loss \citep{infonce}, which, for two modalities, is defined as
\begin{equation*}
\begin{aligned}
&\Ls_{\text{InfoNCE}} (\rvg_1, \rvg_2)\\ &=
\mathop{\mathbb{E}}_{\{\mathbf{X}^i\}_{i=1}^{B}\sim p_{\text{data}}}
\left[
-\frac{1}{B}\sum_{i=1}^{B} \log \frac{ e^{\text{sim}\left(\rvg_1(\rvx_1^i), \rvg_2(\rvx_2^i)\right)/{\tau}}}{\sum_{j=1}^{B}e^{\text{sim}\left(\rvg_1(\rvx_1^i), \rvg_2(\rvx_2^j)\right)/\tau}} 
\right],
\end{aligned}
\end{equation*}
where $\rvg_1$ and $\rvg_2$ are modality-specific encoders, ${\tau>0}$ is a temperature parameter, $B$ is the batch size, and $\text{sim}(\cdot,\cdot)$ is a similarity metric (e.g., cosine similarity) \citep{infonce, hypersphere}. A common variant in multimodal settings is the symmetrized InfoNCE loss, which averages the objective in both directions: $\Ls_{\text{SymInfoNCE}}(\rvg_1, \rvg_2) = \tfrac{1}{2} \Ls_{\text{InfoNCE}} (\rvg_1, \rvg_2) + \tfrac{1}{2}\Ls_{\text{InfoNCE}} (\rvg_2, \rvg_1)$ \citep{radford2021learning, zhang2022contrastive, ident_multimodal}.

\citet{hypersphere} showed that the InfoNCE loss decomposes into two terms: (i) an \emph{alignment} term (the numerator), which pulls positive pairs together, and (ii) a \emph{uniformity} term (the denominator), which prevents representation collapse. 
\citet{ident_unimodal} further showed that, as the number of negative samples grows, the uniformity term can be interpreted as a nonparametric estimator of the negative entropy of the learned representations. Consequently, \citet{ident_unimodal}, \citet{ident_multimodal}, and \citet{ident_multiview} interpret InfoNCE as aligning positive pairs while approximately maximizing representation entropy. According to this interpretation, for two modalities, symmetric InfoNCE asymptotically behaves like the objective:
\begin{equation}
\begin{aligned}
\Ls_{\text{SymAlignEnt}} = &\mathop{\mathbb{E}}_{\mathbf{X}^i\sim p_{\text{data}}}\left[\|\rvg_1(\rvx^i_1)-\rvg_2(\rvx^i_2)\|_2^2\right]\\ &-\tfrac{1}{2}\left[H(\rvg_1(\rvx_1)) + H(\rvg_2(\rvx_2)) \right].
\end{aligned}
\label{eq:symalignent}
\end{equation}
\noindent \textbf{Block-identifiability of shared factors.} Building on this interpretation, \cite{ident_unimodal, ident_multimodal, ident_multiview} studied the identifiability of shared factors, also known as \emph{content}, under this asymptotic loss in unimodal \citep{ident_unimodal} and multimodal \citep{ident_multimodal, ident_multiview} \ac{cl}. They showed that, under suitable generative assumptions, shared factors are \emph{block-identifiable}: the content information can be isolated, without identifying individual latent factors. \citet{ident_multimodal} extended the unimodal results of \citet{ident_unimodal} to two modalities ($M=2$), while \citet{ident_multiview} generalized them to $M>2$ by considering arbitrary subsets of observed modalities. However, both multimodal approaches rely on pairwise objectives, whose number of terms grows quadratically with $M$.

\subsection{Contrastive learning beyond two modalities}

A natural extension of \ac{cl} beyond two modalities is to contrast all pairs, which relates every modality directly but scales quadratically with $M$~\citep{tian2020contrastive}. Other methods instead align all modalities with a fixed anchor, such as images~\citep{girdhar2023imagebind}, language~\citep{zhu2024languagebind}, or audio~\citep{ruan2023accommodating}, achieving linear scaling in $M$ but potentially missing information shared among non-anchor modalities. %
More recent works replace pairwise similarity in InfoNCE with a single score over all modalities~\citep{saporta2024contrasting, cicchetti2025gramian, cicchetti2026triangle}. Among these, GRAM~\citep{cicchetti2025gramian} and TRIANGLE~\citep{cicchetti2026triangle} use the volume and area spanned by modality embeddings, respectively, but still contrast these scores against chosen anchor modalities. 
CentroBind~\citep{jeong2024anchors} instead avoids this choice by using the mean of all modality embeddings as an anchor. It applies a symmetric InfoNCE loss between each modality and this anchor, promoting high entropy in each encoder's outputs, as in~\cref{eq:symalignent}.%

\subsection{Multimodal representation factorization}

Self-supervised factorization methods learn shared and private representations by optimizing mutual information, estimated with contrastive bounds such as InfoNCE~\citep{liang2023factorized, wang2025an, rizou2026repercent}. 
They rely on \emph{data augmentations}, which provide the learning signal for private representations and must therefore preserve task-relevant or modality-specific information.
FactorCL~\citep{liang2023factorized} combines lower and upper bounds to separate task-relevant from task-irrelevant information. DisentangledSSL~\citep{wang2025an} establishes optimality criteria for entangled shared and private information, a setting common in real-world data. In this case, the \ac{mni} criterion~\citep{fischer2020conditional} is unattainable, \ie, shared information cannot be preserved without capturing modality-specific information, leading to \emph{a trade-off between information preservation and leakage reduction}~\citep{wang2025an}. RePercENT~\citep{rizou2026repercent} extends this framework beyond two modalities by learning shared and private representations for pairs of modalities, capturing pairwise rather than globally shared information. 
DisentangledSSL learns both representations sequentially, whereas its JointOpt variant, like FactorCL and RePercENT, learns them jointly.

\section{\method{}: Shared--private factorization}

In this section, we introduce \method{}, an end-to-end \emph{scalable} framework that factorizes pretrained multimodal embeddings into shared and private representations using lightweight adapters and two complementary learning objectives. We describe the architecture and factorization process, followed by the objectives for learning shared and modality-specific representations, respectively.

As illustrated in \Cref{fig:architecture}, \method{} factorizes embeddings extracted by pretrained foundation models. For each modality ${m \in \{1, \dots, M\}}$, a frozen encoder $\Encoder_m$ maps the input $\rvx_m$ to an embedding ${\rve_m = \Encoder_m(\rvx_m)}$. Following \citet{rizou2026repercent}, \emph{we assume that these encoders are sufficiently expressive for their embeddings to preserve most of the relevant modality information}. Therefore, we treat each embedding $\rve_m$ as a composite representation and use lightweight, modality-specific adapters to split it into shared and private (modality-specific) representations, ${\rve_m \mapsto (\hat{\rvs}_m, \hat{\rvw}_m)}$. We jointly train these adapters by minimizing the following objective:
\begin{equation}
    \mathcal{L}(\theta^s, \theta^w, \phi)
    = \underbrace{\mathcal{L}_{\text{shared}}(\theta^s)}_{\text{\Cref{sec:shared}}}
    + \underbrace{\mathcal{L}_{\text{private}}(\theta^w, \phi)}_{\text{\Cref{sec:private}}},
\label{eq:general_loss}
\end{equation}
whose components are detailed in the following sections.

\subsection{Learning shared representations}\label{sec:shared}

A good factorization requires learning shared representations that isolate the information that is invariant across modalities. To this end, for each modality $m$, a shared-representation adapter $\AdapterS_{m}^{s}$ projects the pretrained embedding $\rve_m$ into a common embedding space:
\begin{equation}
    \hat{\rvs}_m = \AdapterS_{m}^{s}(\rve_m; \vtheta_m^s).
\end{equation}
With the pretrained encoders frozen, we optimize only the shared adapter parameters ${\vtheta^s = (\vtheta_m^s)_{m=1}^M}$ by minimizing $\mathcal{L}_{\text{shared}}$ in~\cref{eq:general_loss}. \ac{cl} objectives are a natural choice for $\mathcal{L}_{\text{shared}}$, as they provably identify information shared across positive pairs \citep{ident_unimodal, ident_multimodal, ident_multiview}. However, common multimodal objectives, such as symmetric InfoNCE \citep{radford2021learning, ident_multimodal}, rely on pairwise alignment losses, whose computational cost grows quadratically with $M$.
To overcome this limitation, we propose an alternative objective: \loss{}.

Inspired by multimodal \acp{vae} \citep{mmvae, mmvae+}, which integrate information from multiple modalities into a shared representation, we aggregate the outputs $\hat{\rvs}_m$ of all modality-specific adapters 
via a \ac{moe}. This \emph{aggregated} representation serves as a common \emph{anchor} for \ac{cl}, aligning each modality's shared representation with an embedding that integrates information from all modalities. This enables simultaneous alignment without relying on a fixed single-modality anchor~\citep{girdhar2023imagebind, zhu2024languagebind, ruan2023accommodating} or separate pairwise losses~\citep{tian2020contrastive}.
Formally, \loss{} is defined as
\begin{equation}
\begin{aligned}
&\Ls_{\text{AnchorNCE}}= \\ &
\mathop{\mathbb{E}}_{\{\mathbf{X}^i\}_{i=1}^B\sim p_{\text{data}}} \left[-\frac{1}{MB}\sum_{i=1}^B\sum_{m=1}^{M} \log \frac{e^{(\hat{\rvs}_m^i)^{\top}\rva^i/\tau}}{\left(\sum_{j=1}^{B}e^{
(\rva^{i})^{\top}\rva^{j}/\tau}\right)^{\lambda}}\right],
\end{aligned}
\label{eq:anchorloss}
\end{equation}
\noindent where $B$ is the batch size, ${\tau>0}$ is the InfoNCE temperature \citep{infonce}, and ${\lambda>0}$ controls the trade-off between alignment and uniformity \citep{hypersphere}. We adopt this formulation since it is widely used in the \ac{cl} literature \citep{hypersphere}. We normalize our representations, making the dot product equivalent to cosine similarity, as it has been shown to promote better representations \citep{wang2017normface, chen2020simple}. Specifically, the anchor is defined as $\rva^i = \tilde{\rva}^i/\|\tilde{\rva}^i\|_2$, where ${\tilde{\rva}^i=\tfrac{1}{M}\sum_{m=1}^{M}\hat{\rvs}_m^i}$. We also assume that each adapter outputs unit-norm representations. 

\vspace{0.055in}
\noindent\textbf{Remark. } \citet{jeong2024anchors} also employ anchor-based alignment, but use symmetric InfoNCE to contrast each modality's shared representations with the anchor, \emph{implicitly} encouraging high entropy in the modality-specific representation spaces (see~\cref{eq:symalignent}) \citep{hypersphere, ident_unimodal, ident_multimodal, ident_multiview}. %
In contrast, \loss{} aligns each modality with the anchor while \emph{explicitly promoting high entropy in the anchor space}, as proved in Section~\ref{sec:formal_guarantees}.
This encourages shared representations to align with a common, non-collapsed anchor that captures information shared across modalities. An empirical comparison of the two objectives is provided in Appendix~\ref{app:exp_details}. %

\subsection{Learning private representations}\label{sec:private}

Similarly to previous section, for each modality $m$, we use a private adapter $\AdapterW_{m}^{w}$ to map the pretrained embedding $\rve_m$ into a modality-specific private representation
\begin{equation}
    \hat{\rvw}_m = \AdapterW_{m}^{w}(\rve_m; \vtheta_m^w).
\label{eq:adapter_p}
\end{equation}
We learn the private representations by reconstructing $\rve_m$ from both the shared anchor $\rva$ and $\hat{\rvw}_m$~\citep{bousmalis2016domain}. However, reconstruction alone does not prevent it from encoding shared information.
To address this, we draw on the coarse-to-fine principle of Matryoshka representation learning~\citep{kusupati2022matryoshka} and introduce a \emph{Matryoshka-like reconstruction objective}: a \emph{coarse} reconstruction is first obtained from the shared anchor and then \emph{refined} with the modality-specific information from the private representation. Specifically, we use two separate decoders, $\DecoderA_{m}^{a}$ and $\DecoderW_{m}^{w}$, whose outputs contribute \emph{additively} to the reconstruction of $\rve_m$:
\begin{align}
    \hat{\rve}_m^{a} &=
        \DecoderA_m^{a}(
        \operatorname{sg}(\rva); \vphi_m^{a}), \label{eq:decoder_a}\\
    \hat{\rve}_m^{w} &=
        \DecoderW_m^{w}(
        \hat{\rvw}_m; \vphi_m^{w}), \label{eq:decoder_w}\\
    \hat{\rve}_m &=
        \hat{\rve}_m^{a} +
        \hat{\rve}_m^{w}, \label{eq:decoder_sum}
\end{align}
where ${\operatorname{sg}(\cdot)}$ denotes the stop-gradient operator, which prevents the decoder gradients from reaching the shared adapter parameters $\vtheta^s$. 
The reconstruction loss combines the coarse and refined levels:
\begin{equation}
    \mathcal{L}_{\mathrm{rec}}(\vtheta^w, \vphi)
    = \frac{1}{M} \sum_{m=1}^{M}
    \mathbb{E}\Big[
        \|\hat{\rve}_m^{a}
        - \rve_m^\mathrm{tgt}\|_2^2
        + \|\hat{\rve}_m
        - \rve_m^\mathrm{tgt}\|_2^2
    \Big],
\label{eq:loss_rec}
\end{equation}
where ${\textstyle \rve_m^\mathrm{tgt} = \rve_m / \|\rve_m\|_2}$ is the normalized pretrained embedding,  ${\vtheta^w = (\vtheta_m^w)_{m=1}^M}$ the private adapter parameters and ${\vphi = (\vphi_m^{a}, \vphi_m^{w})_{m=1}^{M}}$ the decoder parameters. Unlike prior methods that require carefully designed augmentations to preserve task-relevant or modality-specific information~\citep{liang2023factorized, wang2025an, rizou2026repercent}, our objective requires \emph{no augmentations}.

To improve parameter efficiency, the anchor decoders
share all parameters except their final output layers, which remain modality-specific, and the same holds for the private decoders.
All adapters and decoders are lightweight MLPs.

\vspace{0.055in}
\noindent\textbf{Hierarchical conditioning.} Existing methods reduce redundancy between shared and private representations through explicit orthogonality regularization~\citep{wang2025an, rizou2026repercent}. %
Instead, we consider an \emph{architectural} approach that encourages this separation by conditioning the private adapter in~\cref{eq:adapter_p} on the shared anchor $\rva$
\begin{equation}     
    \hat{\rvw}_m = \AdapterW_{m}^{w}\big([\rve_m; \operatorname{sg}(\rva)]; \vtheta_m^w\big),
\end{equation}
where $\operatorname{sg}(\cdot)$, as in~\cref{eq:decoder_a}, denotes the stop gradient operator. %
This conditioning encourages $\hat{\rvw}_m$ to capture the residual modality-specific information in $\rve_m$ that the anchor does not already explain. Similar conditioning mechanisms have been explored in hierarchical latent-variable models~\citep{sonderby2016ladder, vahdat2020nvae}. Closer to our setting, this conditioning has been used to capture dependencies between shared and private latents given the observation~\citep{wang2025an} and shown to improve shared--private disentanglement in multimodal \acp{vae}~\citep{vo2026hlder}. We refer to this conditioned variant as \method{}~+~Cond.

\section{\loss{}: formal guarantees}\label{sec:formal_guarantees}

We analyze the theoretical properties of our proposed objective for multimodal \ac{cl}, \loss. We first formalize the assumptions, then derive the objective's asymptotic form and establish identifiability guarantees extending those of InfoNCE  to an arbitrary number of modalities. 

\subsection{Assumptions on the data generating process}\label{sec:gen_process}

Following previous identifiability analysis \citep{ident_unimodal, ident_multimodal, ident_multiview}, we formalize the multimodal data generation process as a latent variable model. Specifically, we assume a continuous latent variable $\rvz$ taking values in %
${\mathcal{Z}\subseteq\sR^{d}}$, with probability density $p_{\rvz}$, which contains all the information required to generate observations across modalities. We further assume that $\rvz$ can be decomposed into: (i) an invariant component $\rvs$, which captures information shared across all modalities, and (ii) $M$ modality-specific (private) components $\rvw_m$. %

Each modality-specific observation $\rvx_m$, taking values in ${\mathcal{X}_m\subseteq\sR^{D_m}}$, is then generated by applying a modality-specific mixing function ${\rvf_m:\mathcal{Z}_m\rightarrow\mathcal{X}_m}$ to its corresponding latent representation ${\rvz_m=(\rvs_m,\rvw_m)}$, which takes values in ${\mathcal{Z}_m \subseteq \mathbb{R}^{d_z}}$:
\begin{equation}\label{eq:gen_model}
\rvz \sim p_{\rvz}, \qquad \rvx_m = \rvf_m(\rvz_m).
\end{equation}
We assume that the shared latent components are identical across modalities, \ie ${\rvs_m=\rvs_n}$ almost everywhere for all ${m\neq n}$, where ${m,n\in\{1,\dots,M\}}$. Furthermore, each mixing function $\rvf_m$ is assumed to be a smooth invertible mapping with a smooth inverse (\ie a diffeomorphism) between the latent representation and observation spaces. In general, each modality has a distinct mixing function, reflecting the different ways in which the shared and private latent components generate the observations.

\begin{figure*}[t]
    \centering
    \includegraphics[width=\textwidth]{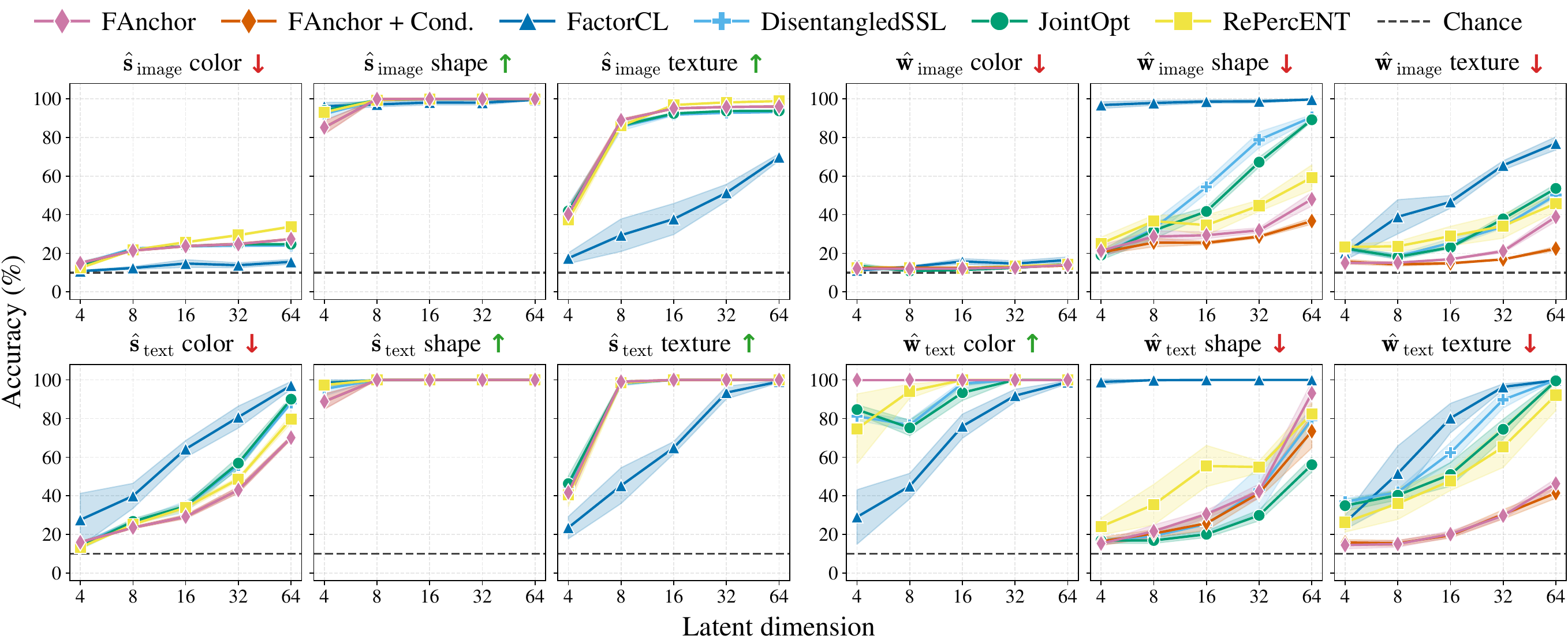}
    \caption{\textbf{Factorization on Clevr-4 (Image without color, Text).} Linear-probe accuracy (\%) for predicting color, shape, and texture from the shared (left) and private (right) representations across latent dimensions. \up/\down: higher/lower is better; dashed lines: chance (10\%).}
    \label{fig:clevr4_image_no_color_text}
    \vspace{-0.1in} %
\end{figure*}

\subsection{Identifiability of shared factors}

Following \cite{ident_unimodal,ident_multimodal, ident_multiview}, we first formalize block-identifiability in the multimodal setting with $M$ modalities, each associated with a mixing function $\rvf_m$ and encoder $\rvg_m$. %
Rather than recovering individual factors of variation, we aim to isolate the invariant information shared across modalities from the modality-specific information. 
As in \citet{ident_multimodal} and \citet{ident_multiview}, we allow for distinct modality-specific latent variables and data-generating mechanisms. We therefore aim to establish the block-identifiability of shared information across modalities under \loss{}, which scales linearly with $M$.
\begin{definition}[Block-identifiability] \label{def:blockident}
We say that the true shared information $\rvs = \rvf_1^{-1}(\rvx_1)_{1:d_s}= \ldots=\rvf_M^{-1}(\rvx_M)_{1:d_s}$ is block-identified by a function ${\rvg_m: \mathcal{X}_m \rightarrow \mathcal{Z}_m}$, with ${m \in \{1,\ldots,M\}}$, if there exists an invertible function ${\rvh_m:\sR^{d_s}\rightarrow\sR^{d_s}}$, s.t. for the inferred shared information ${\hat{\rvs}_m=\rvg_m(\rvx_m)_{1:d_s}}$ it holds that ${\hat{\rvs}_m=\rvh_m(\rvs)}$. In other words, the inferred shared information from each modality $\hat{\rvs}_m$ contains all and only the information about $\rvs$.    
\end{definition}
\vspace{-0.02in}

Identifiability guarantees for InfoNCE rely on its asymptotic behavior as the number of negative samples approaches infinity~\cite{ident_unimodal,ident_multimodal,ident_multiview}. Following \citet{hypersphere}, we show that \loss{} asymptotically decomposes into \emph{alignment} and \emph{uniformity} terms. The following proposition extends their result, establishing that optimizing \loss{} requires both, with uniformity enforced directly on the aggregated anchors. The proof follows Theorem 1 in \cite{hypersphere} and is provided in Appendix~\ref{app_proof_thm1}.

\begin{proposition}[Asymptotics of $\Ls_{\text{AnchorNCE}}$]\label{thm:asymptotics} 
For fixed $\tau > 0$ and ${\lambda > 0}$, as the number of negative samples ${K\rightarrow\infty}$, our proposed contrastive loss converges to
\begin{equation}
    \begin{aligned}
        \lim_{K\rightarrow\infty} &\left(\Ls_{\text{\emph{AnchorNCE}}} - \lambda\log{K}\right) = \\
        &-\frac{1}{M\tau}\mathop{\mathbb{E}}_{\mathbf{X}^i\sim p_{\text{\emph{data}}}}\left[ \sum_{m=1}^{M}\rvg_m(\rvx_m^i)^{\top} \rva^i \right]\\
        &+\lambda\mathop{\mathbb{E}}_{\mathbf{X}^i\sim p_{\text{\emph{data}}}}\left[\log \mathop{\mathbb{E}}_{\mathbf{X}^j\sim p_{\text{\emph{data}}}}\left[e^{ (\rva^i)^{\top}\rva^j/\tau}\right]\right],
    \end{aligned}
    \label{eq:thm1}
\end{equation}
where $\rva^i = \tilde{\rva}^i / \|\tilde{\rva}^i\|_2$ with $\tilde{\rva}^i = \tfrac{1}{M}\sum_{m=1}^{M} \rvg_m(\rvx_m^i)$, assuming that all modality-specific encoders output unit-norm representations. Here, $\mathbf{X}^i$ denotes the positive sample and $\mathbf{X}^j$ a negative sample drawn independently of $\mathbf{X}^i$, assuming that the event ${\mathbf{X}^j = \mathbf{X}^i}$ has probability zero. Moreover:
    \begin{enumerate}
        \item The first term is minimized iff all the modality-specific encoders are perfectly aligned.
        \item If modality-specific encoders exist such that the anchors are perfectly uniform, they form the exact minimizers of the second term.
        \item For the convergence in~\cref{eq:thm1}, the absolute deviation from the limit decays on $\mathcal{O}(K^{-1/2})$.

    \end{enumerate}
\end{proposition}
\vspace{-0.02in}

Then, in the limit of infinite negative samples, our proposed loss behaves as
\begin{equation}\label{eq:anchoralignmaxent}
\begin{aligned}
&\mathcal{L}_{\text{AnchorAlignEnt}}\\ &= \mathop{\mathbb{E}}_{\mathbf{X}^i\sim p_{\text{data}}}\left[\frac{1}{M}\sum_{m=1}^M\|\rvg_m(\rvx^i_m) - \tilde\rva^i\|_2^2\right] - \lambda H(\tilde{\rva}),
\end{aligned}
\end{equation}
interpreting the uniformity term as a nonparametric entropy estimator~\citep{ident_unimodal}. Following prior work~\citep{ident_unimodal, ident_multimodal, ident_multiview},~\cref{eq:anchoralignmaxent} operates in the unnormalized embedding space, which in~\cref{eq:anchorloss} is equivalent to replacing the dot product with cosine similarity. 
While our experiments use the finite-sample objective $\mathcal{L}_{\text{AnchorNCE}}$, our block-identifiability results in Proposition~\ref{thm:identifiability} are derived for the asymptotic loss $\mathcal{L}_{\text{AnchorAlignEnt}}.$ %
The proof, provided in Appendix~\ref{app_proof_thm2}, follows the proof outlines of Theorem 4.4 in~\citet{ident_unimodal} and Theorem 1 in~\citet{ident_multimodal}.

\begin{proposition}[Identifying shared information]\label{thm:identifiability}
Assume the data generating process described in Section~\ref{sec:gen_process}, where the different modalities are generated from~\cref{eq:gen_model}, with ${p_{\rvz_{m}|\rvz_{n}}}$ as defined in Assumption~\ref{content_invariance} (Appendix~\ref{app_proof_thm2}). Further, let ${p_{\rvz}}$ be a smooth and continuous density on $\mathcal{Z}$ with ${p_\rvz(\rvz)>0}$ almost everywhere. Let ${\rvg_m: \mathcal{X}_m\rightarrow(0,1)^{d_s}}$, ${m\in\{1,\ldots,M\}}$, be smooth functions that minimize $\mathcal{L}_{\text{AnchorAlignEnt}}$ as defined in~\cref{eq:anchoralignmaxent}. Then ${\rvg_1, \ldots, \rvg_M}$ block-identify the true shared variables in the sense of Definition \ref{def:blockident}.
\end{proposition}
\vspace{-0.02in}

\section{Experiments} \label{sec:exp_results}

In this section, we empirically evaluate \method{} and \loss{}. Specifically, we assess (i) the effectiveness of \method{} in factorizing shared and private information %
(\Cref{sec:exp_factor}), (ii)~the robustness of shared representations to adversarial samples in the retrieval gallery (\Cref{sec:exp_robust}), and (iii)~the ability of \loss{} to learn shared representations across more than two modalities (\Cref{sec:exp_multi_align}).

\vspace{0.055in}
\noindent \textbf{Baselines.} We compare our methods against two groups of \ac{sota} baselines. For factorization and robustness, we consider FactorCL~\citep{liang2023factorized}, DisentangledSSL and its JointOpt variant~\citep{wang2025an}, and RePercENT~\citep{rizou2026repercent}. For multimodal alignment, we consider GRAM~\citep{cicchetti2025gramian} and TRIANGLE~\citep{cicchetti2026triangle}. Unless stated otherwise, we report results for the best hyperparameter configuration of each method, averaged over 5 seeds.

\subsection{Evaluation of shared--private factorization}
\label{sec:exp_factor}

Following prior work~\citep{wang2025an,rizou2026repercent}, we evaluate the shared--private factorization with linear probes, trained separately on the shared and private representations to predict shared and modality-specific targets, \ie, attributes or labels. Ideally, each target should be predictable only from its corresponding representation. Thus, we aim to preserve relevant information while minimizing leakage. In real-world settings, where shared and modality-specific information are often entangled, these goals involve an inherent \emph{trade-off}.

\subsubsection{Controlled setting: Clevr-4} 
\label{sec:clevr4}
\begin{figure}[t]
    \centering
    \includegraphics[width=0.475\textwidth]{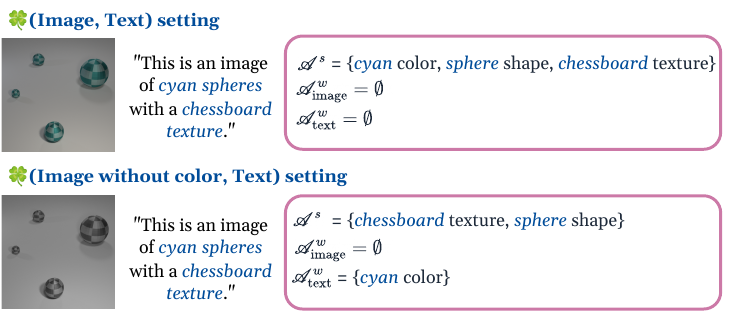}
    \caption{Two Clevr-4 settings with shared attributes $\mathscr{A}^s$ and private attributes $\mathscr{A}^w_m$ for each modality $m$. Removing an attribute from one modality makes it private to the other.} 
    \label{fig:clevr4_exp}
\vspace{-0.1in} %
\end{figure}

Clevr-4~\citep{vaze2023no} is a synthetic image dataset in which all objects in an image share the same \emph{shape}, \emph{color}, and \emph{texture}, each sampled independently from 10 classes (we omit the object count). We then pair each image with a template-based caption describing these attributes. To control which information is shared across modalities, we consider four settings, obtained by optionally removing color from images (via grayscale conversion) and/or shape from captions. \Cref{fig:clevr4_exp} illustrates two of these settings.

\vspace{0.055in}
\noindent \textbf{Setup.} %
We use frozen pretrained \texttt{CLIP-RN50} image and text encoders~\citep{radford2021learning}. %
To assess the effect of representation capacity, we set the shared and private dimensions equal across methods and vary them over $\{4, 8, 16, 32, 64\}$. For each model, we train separate linear probes on the shared $\hat{\mathbf{s}}_m$ and private representations $\hat{\mathbf{w}}_m$, $m \in \{\text{image}, \text{text}\}$, to predict \emph{color}, \emph{shape}, and \emph{texture}. We report classification accuracy (\%). For our methods, we set $\lambda = 1$ in \loss{} and provide an ablation over $\lambda$ in Appendix~\ref{app:exp_details}.

\vspace{0.055in}
\noindent \textbf{Results.} \Cref{fig:clevr4_image_no_color_text} shows results for a setting where \emph{shape} and \emph{texture} are shared across modalities, while \emph{color} is private to text. Shared representations learned by \method{} capture both shape and texture with nearly 100\% accuracy for a number of dimensions ${>8}$, matching the strongest baselines, while FactorCL fails to capture texture. Our private text representations also capture color with 100\% accuracy at all tested dimensions, 
whereas baselines need more capacity.
Meanwhile, our private representations exhibit less shared-to-private leakage in nearly all cases. At dimension $64$, for example, all baselines predict shared texture from $\hat{\mathbf{w}}_{\text{text}}$ with over 90\% accuracy, 
compared with at most 46\% for ours. \method{} + Cond. further reduces this leakage without altering the shared representations, supporting the use of conditional private adapters. Generally, leakage into both representations tends to increase with dimensionality, suggesting that excess capacity allows representations to capture information beyond their intended latent factors.

\begin{table*}[t]
\centering
\caption{\textbf{Robustness of shared representations.} Image-to-text retrieval on clean MS-COCO 5K and RoCOCO with adversarial captions added to the retrieval gallery. \up/\down: higher/lower is better. Best in \textbf{bold}, second-best \underline{underlined}.}
\vspace{-0.05in} 
\label{tab:rococo_i2t}
\footnotesize
\setlength{\tabcolsep}{2.5pt}
\begin{tabular}{r c ccc ccc ccc ccc}
\toprule
& \textbf{COCO 5K}
& \multicolumn{3}{c}{\textbf{Rand-voca}}
& \multicolumn{3}{c}{\textbf{Same-concept}}
& \multicolumn{3}{c}{\textbf{Diff-concept}}
& \multicolumn{3}{c}{\textbf{Danger}} \\
\cmidrule(lr){2-2}
\cmidrule(lr){3-5}
\cmidrule(lr){6-8}
\cmidrule(lr){9-11}
\cmidrule(lr){12-14}
\textbf{Image to Text}
& \textbf{R@1}\up
& \textbf{R@1}\up & \textbf{drop}\down & \textbf{RSMS}\down
& \textbf{R@1}\up & \textbf{drop}\down & \textbf{RSMS}\down
& \textbf{R@1}\up & \textbf{drop}\down & \textbf{RSMS}\down
& \textbf{R@1}\up & \textbf{drop}\down & \textbf{RSMS}\down \\
\midrule
\rowcolor{ourpurple}
\methodb \, (ours)
& \underline{50.71}
& \textbf{46.32} & \textbf{8.66} & \textbf{9.47}
& \textbf{45.90} & \textbf{9.48} & \textbf{11.08}
& \textbf{47.60} & \textbf{6.13} & \textbf{7.14}
& \textbf{47.52} & \textbf{6.28} & \textbf{6.77} \\
\midrule
CLIP {\scriptsize(zero-shot)}~\citep{radford2021learning}
& 50.12
& 39.14 & 21.91 & 28.52
& 39.04 & 22.11 & 29.28
& 40.28 & 19.63 & 26.48
& 41.30 & 17.60 & 22.98 \\

CLIP {\scriptsize(proj. FT)}~\citep{radford2021learning}
& \textbf{58.57}
& \underline{44.29} & 24.38 & 30.10
& \underline{45.84} & 21.73 & 27.27
& \underline{47.28} & 19.28 & 24.30
& \underline{47.46} & 18.97 & 22.90 \\
\midrule

FactorCL~\citep{liang2023factorized}
& 22.23
& 16.78 & 24.51 & 29.15
& 15.55 & 30.03 & 35.97
& 17.16 & 22.78 & 27.97
& 17.27 & 22.30 & 24.70 \\

DisentangledSSL~\citep{wang2025an}
& 48.04
& 41.44 & \underline{13.72} & \underline{15.62}
& 41.62 & \underline{13.35} & \underline{15.90}
& 43.82 & \underline{8.77} & \underline{10.76}
& 42.94 & \underline{10.62} & \underline{12.59} \\

JointOpt~\citep{wang2025an}
& 50.46
& 43.38 & 14.03 & 15.96
& 43.57 & 13.65 & 16.37
& 45.86 & 9.10 & 11.12
& 44.49 & 11.83 & 12.81 \\

RePercENT~\citep{rizou2026repercent}
& 34.06
& 25.43 & 25.33 & 29.96
& 25.49 & 25.16 & 30.10
& 26.86 & 21.13 & 25.60
& 26.18 & 23.13 & 26.98 \\
\bottomrule
\end{tabular}
\vspace{-0.05in} %
\end{table*}

\subsubsection{Real-world setting: VGGSounder}
\label{sec:vggsounder}

VGGSounder~\citep{zverev2025vggsounder} is a multi-label audio-visual classification benchmark that re-annotates the VGGSound~\citep{chen2020vggsound} test set with 309 classes. Human annotators labeled each class as audible, visible, or both, defining three tasks: two private tasks on audio-only and video-only labels, and a shared task on audio-video labels. Unlike Clevr-4, shared and modal\-ity-specific information may be entangled. For example, a clip has the audio-only label ``wind noise'' and the video-only label ``wind rustling leaves'', yet the moving leaves provide a visual cue to the audible wind, making information labeled as audio-only partly recoverable from video.

\begin{figure}[t]
    \centering
    \includegraphics[width=0.475\textwidth]{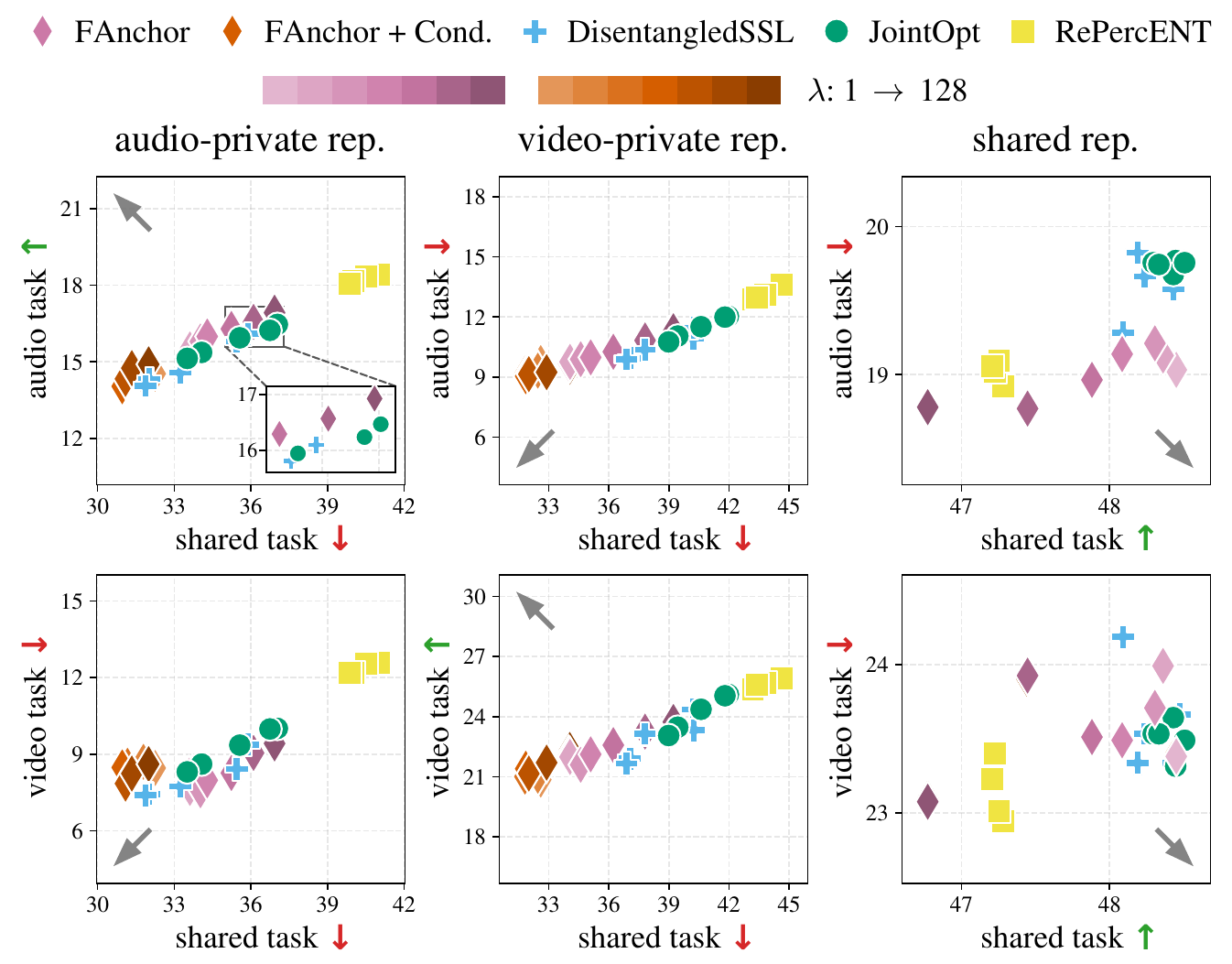}
    \caption{\textbf{Factorization on VGGSounder.} Linear-probe F1@1 (\%) on shared and private tasks from audio-private (left), video-private (middle), and shared (right) representations. Each point is one hyperparameter configuration; color gradients indicate $\lambda$ from 1 to 128 for ours. \up/\down: higher/lower is better; \optarrow: optimal regions.}
    \label{fig:av_single_f1}
    \vspace{-0.15in} %
\end{figure}

\vspace{0.055in}
\noindent \textbf{Setup.} We use frozen pretrained \texttt{PE-AV-Base} audio and video encoders~\citep{vyas2026pushing}.
All methods are trained on VGGSound and evaluated on VGGSounder. We train linear probes on the private representations $\hat{\mathbf{w}}_m$, $m \in \{\text{audio}, \text{video}\}$, and on the shared representation, \ie, the concatenation $[\hat{\mathbf{s}}_{\text{audio}}, \hat{\mathbf{s}}_{\text{video}}]$ for the baselines and the anchor $\mathbf{a}$ for ours. Each probe predicts all 309 classes and is evaluated on the three tasks using F1@1 (\%)~\citep{zverev2025vggsounder}. To illustrate the preservation--leakage trade-off, we report multiple hyperparameter configurations, varying ${\lambda \in \{1,4,8,16,32,64,}$ $128\}$ for our methods (baseline settings in Appendix~\ref{app:exp_details_1}).

\vspace{0.055in} \noindent \textbf{Results.}  %
\Cref{fig:av_single_f1} plots shared- and private-task F1@1 for each linear probe. For the private representations (left, middle), all models lie along a diagonal: higher private-task performance comes with more shared-to-private leakage, suggesting these tasks also rely on shared factors. Our methods show lower leakage, thus isolating private and shared components better than the baselines. The hyperparameter $\lambda$, which balances alignment and uniformity, moves our models along this diagonal, directly controlling this trade-off. At comparable leakage, \method{} achieves slightly higher audio-task F1@1 on the audio-private representation than DisentangledSSL and JointOpt, while RePercENT reaches the highest private-task F1@1 at the cost of greater leakage.

For the shared representations (right), \method{}~+~Cond.\ and \method{} 
yield identical results, as 
conditioning does not affect the shared adapters. 
Both remain competitive on the shared task while leaking less audio-task information than DisentangledSSL and JointOpt, whereas RePercENT lags behind.
Here, $\lambda$ mainly controls shared-task performance, while private-task leakage varies only slightly.

\subsection{Robustness of shared representations ($M=2$)}
\label{sec:exp_robust}

A shared space may encourage abstract representations that rely less on modality-specific details, such as individual words or image regions, and are thus more robust to their perturbation. Prior work reports such robustness to sensor noise in multimodal action segmentation with a shared--private architecture~\citep{van2023aspnet}.
We thus evaluate the robustness of \method's shared representations on cross-modal retrieval with adversarial samples added to the retrieval gallery.

\vspace{0.055in}
\noindent \textbf{RoCOCO.} RoCOCO~\citep{park2024rococo} is a robustness benchmark built on the MS-COCO 5K test set~\citep{lin2014microsoft} that adds adversarial captions and images to the retrieval galleries. Adversarial captions replace a single word of the original caption with a random (Rand-voca), a same-concept, a different-concept (Diff-concept), or a danger-related word (Danger). 
Adversarial images are described in Appendix~\ref{app:exp_details_1}.

\vspace{0.055in}
\noindent \textbf{Setup.} We use frozen pretrained \texttt{CLIP-ViT-B/32} image and text encoders~\citep{radford2021learning}. All methods are trained on the MS-COCO training set and evaluated on its 5K test set and RoCOCO. Besides the factorization baselines, we compare against CLIP, evaluated \emph{zero-shot} and with fine-tuned projection heads. We report Recall@1 (R@1, \%), its relative drop (\%) from MS-COCO to RoCOCO, and the Recall Score of Manipulated Samples (RSMS)~\citep{park2024rococo}, \ie, the percentage of queries for which an adversarial sample is retrieved at rank~1. For ours, we set $\lambda = 4$ in \loss{}.

\vspace{0.055in}
\noindent \textbf{Results.}  
\Cref{tab:rococo_i2t} reports image-to-text retrieval results. On clean MS-COCO, \method{} ranks second in R@1 (50.7\%),
behind only CLIP with fine-tuned projection heads (58.6\%), and outperforms all factorization baselines. On RoCOCO, it achieves the highest R@1 and the lowest relative drop and RSMS in every setting: its drop stays below 10\% and its RSMS below 12\%, whereas both CLIP variants lose at least 17.6\% of their R@1 and reach an RSMS of at least 22.9\%. DisentangledSSL and JointOpt also improve robustness over both CLIP variants, but their RSMS remains 3.6 to 6.5 points above our method. Similar trends also hold for text-to-image retrieval (see Appendix~\ref{app:exp_details}).

Our gains highlight the effectiveness of \loss{} over both CLIP variants, which are trained with the symmetric InfoNCE loss, whose theoretical guarantees are analogous for two modalities~\citep{ident_multimodal}.
Despite its higher R@1 on clean MS-COCO, fine-tuned CLIP reaches an RSMS 2.5 to 3.4 times higher than ours on RoCOCO. A similar gap appears when we train our adapters with the same loss (see Appendix~\ref{app:exp_details}). This agrees with \citet{park2024rococo}, who show that many \acp{vlm} degrade substantially at any pre-training scale, as they tend to rely on specific words or image regions rather than the overall semantic context. Altogether, \method{}'s shared representations are the most robust among all methods.

\subsection{Scaling to $M>2$ modalities}
\label{sec:exp_multi_align}

Finally, we evaluate \loss{} independently of the factorization and compare it beyond two modalities with the geometric contrastive objectives GRAM~\citep{cicchetti2025gramian} and TRIANGLE~\citep{cicchetti2026triangle}, which score all modalities jointly.

\vspace{0.055in}
\noindent \textbf{VAST-150K.} This subset of VAST-27M~\citep{chen2023vast}, released by \citet{cicchetti2025gramian}, contains 150K open-domain video clips (frames and audio) paired with automatically generated captions, yielding three modalities: video, audio, and text.

\begin{table}[t]
\centering
\caption{{\boldmath\textbf{Alignment beyond two modalities.}} Zero-shot text-to-video (T2V) and video-to-text (V2T) retrieval (R@1, \%) on MSR-VTT, ActivityNet, and VATEX, with videos given as frames and audio. 
\up: higher is better. Best in \textbf{bold}, second-best \underline{underlined}.}
\vspace{-0.05in}
\label{tab:vast}
\footnotesize
\setlength{\tabcolsep}{1.5pt}
\begin{tabular}{rcccccc}
\toprule
&
\multicolumn{2}{c}{\textbf{MSR-VTT}}
& \multicolumn{2}{c}{\textbf{ActivityNet}}
& \multicolumn{2}{c}{\textbf{VATEX}} \\
\cmidrule(lr){2-3}
\cmidrule(lr){4-5}
\cmidrule(lr){6-7}
\textbf{Method}
& \textbf{T2V}\up & \textbf{V2T}\up
& \textbf{T2V}\up & \textbf{V2T}\up
& \textbf{T2V}\up & \textbf{V2T}\up \\
\midrule
\rowcolor{ourpurple}
\lossb \, (ours)
& \textbf{52.86} & \textbf{51.34}
& \underline{56.48} & 50.92
& \textbf{79.64} & \underline{74.74} \\
\midrule
VAST~\citep{chen2023vast}
& 49.20 & 46.90
& 51.80 & 47.80
& 75.70 & 73.00 \\
\midrule

GRAM~\citep{cicchetti2025gramian}
& \underline{52.58} & 50.14
& \textbf{59.28} & \underline{52.72}
& \underline{79.28} & 74.42 \\

TRIANGLE~\citep{cicchetti2026triangle}
& 49.86 & \underline{50.86}
& 56.06 & \textbf{53.44}
& 77.56 & \textbf{74.78} \\
\bottomrule
\end{tabular}
\vspace{-0.1in} %
\end{table}

\vspace{0.055in}
\noindent \textbf{Setup.} 
We use pretrained VAST encoders~\citep{chen2023vast}: \texttt{BERT-B} (text)~\citep{devlin2019bert}, \texttt{BEATs} (audio)~\citep{pmlr-v202-chen23ag}, and \texttt{EVA-CLIP ViT-G} (video)~\citep{sun2023eva}, without fusion layers. All methods fine-tune the full model on VAST-150K using their respective contrastive objectives and are evaluated zero-shot on MSR-VTT~\citep{xu2016msr}, ActivityNet~\citep{caba2015activitynet}, and VATEX~\citep{wang2019vatex}. We measure text-to-video (T2V) and video-to-text (V2T) retrieval using Recall@1 (R@1,\%).
As a reference, we report the pretrained VAST model with fusion layers, which we do not fine-tune since prior work reports no improvement~\citep{cicchetti2026triangle}.
For ours, we set $\lambda=16$ and compute text-to-anchor similarity, where the anchor aggregates video and audio embeddings.

\vspace{0.055in}

\noindent \textbf{Results.} \Cref{tab:vast} shows that models fine-tuned with \loss{} outperform pretrained VAST on all datasets in both T2V and V2T retrieval, suggesting that \loss{} further aligns the modality embeddings towards a shared multimodal space.
We compare \loss{} against GRAM and TRIANGLE using two-sided Student's $t$-tests across 5 seeds (${p < 0.05}$; standard deviations in Appendix~\ref{app:exp_details}). \loss{} significantly outperforms TRIANGLE in T2V on all datasets, and GRAM in V2T on MSR-VTT, while other differences on MSR-VTT and VATEX are not significant.
On ActivityNet, GRAM significantly outperforms \loss{} in both tasks, and TRIANGLE does so in V2T. 

Overall, \loss{} achieves retrieval performance competitive with \ac{sota} multimodal contrastive objectives while scaling linearly with the number of modalities: it requires ${O(Md)}$ operations per sample, compared with ${O(M^2d)}$ for GRAM, which computes all pairwise inner products. In contrast, TRIANGLE is limited to three modalities. These properties make \loss{} an effective and scalable objective for joint multimodal alignment.

\section{Conclusion}

We proposed \method{}, a scalable framework that factorizes pretrained multimodal embeddings into shared and private representations using lightweight adapters trained with novel contrastive and Matryoshka-like reconstruction losses. We formally guarantee that the former block-iden\-ti\-fies the shared information for any number of modalities.
We evaluated \method{} on synthetic and real-world datasets for factorization quality, robustness, and multimodal alignment.
A limitation is that typical real-world downstream tasks largely depend on shared factors, making it hard to isolate
 the contribution of the private representations.
 Generative tasks are a natural way to further analyze 
 their role~\citep{ramesh2022hierarchical, mmvae+, zhang2026disentanglement}, as generation should depend on private information and our reconstruction loss already includes a decoder. Extending \method{} to token-level features for dense prediction, such as segmentation~\citep{zhou2022extract, Chang_2019_CVPR, wang2023multi}, also provides an interesting avenue for future work.

\section*{Acknowledgements}

This work has been supported by the project \textit{“Society-Aware Machine Learning: The paradigm shift demanded by society to trust machine learning”}, funded by the European Union and led by Isabel Valera (ERC-2021-STG, SAML, 101040177). Huyen Vo acknowledges her membership in the CS@Max Planck doctoral program. Isabel Valera and Mar\'{i}a Mart\'{i}nez-Garc\'{i}a also acknowledge the stimulating research environment of the GRK 2853/1 “\emph{Neuroexplicit Models of Language, Vision, and Action}”, funded by the Deutsche Forschungsgemeinschaft (DFG, German Research Foundation) under project number 471607914. Views and opinions expressed are, however, those of the author(s) only and do not necessarily reflect those of the aforementioned funding agencies. Neither of the aforementioned parties can be held responsible for them.

{
    \small
    \bibliographystyle{ieeenat_fullname}
    \bibliography{main}
}

\clearpage

\clearpage
\onecolumn
\setcounter{page}{1}
\appendix

{\centering
\Large
\textbf{\thetitle}\\
\vspace{0.5em}Supplementary Material \\
\vspace{1.0em}
}

\addtocontents{toc}{\protect\setcounter{tocdepth}{3}}  %

\tableofcontents

\vspace{0.08in}

\section{Proofs}
\label{app:proofs}

\subsection{Proof of Proposition 1}
\label{app_proof_thm1}

\begin{assumption}[Well-defined (normalized) anchors]\label{ass:anchor}
Let $\tilde{\rva}^i=\tfrac{1}{M}\sum_{m=1}^{M}\rvg_m(\rvx^i_m)$ denote the
unnormalized anchor, so that ${\rva^i=\tilde{\rva}^i/\|\tilde{\rva}^i\|_2}$.
We assume that $\|\tilde{\rva}^i\|_2>0$ holds under $p_{\text{data}}$,
i.e.,
\begin{equation*}
    \mathbb{P}_{\rmX^i\sim p_{\text{data}}}\!\left(\|\tilde{\rva}^i\|_2=0\right)=0.
\end{equation*}
\end{assumption}

\setcounter{theorem}{0}
\begin{proposition}[Asymptotics of $\Ls_{\text{AnchorNCE}}$] \label{thm1_ap}
For fixed $\tau > 0$ and $\lambda > 0$, as the number of negative samples $K\rightarrow\infty$, our proposed (normalized) contrastive loss converges to
\begin{equation}
    \begin{aligned}
        &\lim_{K\rightarrow\infty} \left(\Ls_{\text{\emph{AnchorNCE}}}(\rvg_1, \ldots, \rvg_M) - \lambda\log{K} \right)\\
        &= \lim_{K\rightarrow\infty} \left( \mathop{\mathbb{E}}_{\substack{ \rmX^i\sim p_{\text{\emph{data}}} \\ \{\rmX^{j}\}_{j=1}^{K}\overset{\scriptscriptstyle\mathrm{i.i.d.}}{\sim} p_{\text{\emph{data}}}}}
         \left[-\frac{1}{M}\sum_{m=1}^{M} \log \frac{e^{\rvg_m(\rvx^i_m)^{\top}\rva^i / \tau}}{ \left(e^{(\rva^i)^{\top}\rva^i/\tau} + \sum_{j=1}^{K}e^{(\rva^i)^{\top}\rva^{j}/\tau}\right)^{\lambda}}\right] - \lambda\log{K} \right) \\
        &=-\frac{1}{M\tau}\mathop{\mathbb{E}}_{\substack{\rmX^i\sim p_{\text{\emph{data}}}}}\left[ \sum_{m=1}^M\rvg_m(\rvx^i_m)^{\top} \rva^i \right] +
        \lambda\mathop{\mathbb{E}}_{\substack{ \rmX^i\sim p_{\text{\emph{data}}} }}
        \left[\log \mathop{\mathbb{E}}_{\substack{ \rmX^{j}\sim p_{\text{\emph{data}}} }}
        \left[e^{(\rva^i)^{\top}\rva^j/\tau}\right]\right],
    \end{aligned}
    \tag{\ref{eq:thm1}}
\end{equation}
where $\rva^i = \tilde{\rva}^i / \|\tilde{\rva}^i\|_2$ with $\tilde{\rva}^i = \tfrac{1}{M}\sum_{m=1}^{M} \rvg_m(\rvx_m^i)$, assuming that all modality-specific encoders output unit-norm representations. Here, $\mathbf{X}^i$ denotes the positive sample and $\mathbf{X}^j$ a negative
sample drawn independently of $\mathbf{X}^i$, assuming that the event $\mathbf{X}^j = \mathbf{X}^i$ has probability zero. Moreover:

    \begin{enumerate}
        \item The first term is minimized iff all the modality-specific encoders are perfectly aligned.
        \item If modality-specific encoders exist such that the anchors are perfectly uniform, they form the exact minimizers of the second term.
        \item For the convergence in  \cref{eq:thm1}, the absolute deviation from the limit decays on $\mathcal{O}(K^{-1/2})$.
    \end{enumerate}
\end{proposition}

\emph{Proof of Proposition 1.} We first show the convergence of our proposed loss, and then the relations between the two limiting terms and the alignment and uniformity properties. We follow the proof outline of Theorem 1 in \citet{hypersphere}.\\

\textbf{Proof of convergence in \cref{eq:thm1}.}

Note that for any positive sample $\rmX^i = (\rvx^i_1, \ldots, \rvx^i_M)$ with $\rvx^i_m \in \mathbb{R}^{D_m}$ and negative samples $\{\rmX^{j}\}_{j=1}^{K} \overset{\scriptscriptstyle\mathrm{i.i.d.}}{\sim} p_{\text{data}}$, we have
\begin{equation}
\begin{aligned}
    &\lim_{K\rightarrow\infty} \lambda\log{\left(\frac{1}{K}e^{(\rva^i)^{\top}\rva^i/\tau} + \frac{1}{K}\sum_{j=1}^{K}e^{(\rva^i)^{\top}\rva^{j}/\tau}\right)}  = \lambda\log \mathop{\mathbb{E}}_{\substack{ \rmX^{j}\sim p_{\text{data}} }}
    \left[ e^{(\rva^i)^{\top}\rva^{j}/\tau} \right] \quad \text{almost surely,}
\end{aligned} \label{eq:limit}
\end{equation}
by the \ac{slln} and the Continuous Mapping Theorem \citep{hypersphere}. Then, we can derive
\begin{equation*}
    \begin{aligned}
        &\lim_{K\rightarrow\infty} \left( \Ls_{\text{AnchorNCE}}(\rvg_1, \ldots, \rvg_M) - \lambda\log{K}\right) \\
        &= \mathop{\mathbb{E}}_{\substack{\rmX^i \sim p_{\text{data}}}} \left[-\frac{1}{M}\sum_{m=1}^M\rvg_m(\rvx^i_m)^{\top}\rva^i/\tau\right] 
        + \lim_{K\rightarrow\infty} \mathop{\mathbb{E}}_{\substack{ \rmX^i \sim p_{\text{data}} \\ \{\rmX^{j}\}_{j=1}^{K 
        } \overset{\mathrm{i.i.d.}}{\sim} p_{\text{data}}}} \left[  \lambda\log\left(\frac{1}{K}e^{(\rva^i)^{\top}\rva^i/\tau} + \frac{1}K{}\sum_{j=1}^{K}e^{(\rva^i)^{\top}\rva^{j}/\tau}\right)\right]\\
        &= \mathop{\mathbb{E}}_{\substack{\rmX^i \sim p_{\text{data}}}} \left[-\frac{1}{M}\sum_{m=1}^M\rvg_m(\rvx^i_m)^{\top}\rva^i/\tau\right] 
        + \mathop{\mathbb{E}}_{\substack{ \rmX^i \sim p_{\text{data}} \\ \{\rmX^{j}\}_{j=1}^{K 
        } \overset{\mathrm{i.i.d.}}{\sim} p_{\text{data}}}} \left[  \lim_{K\rightarrow\infty}\lambda\log\left(\frac{1}{K}e^{(\rva^i)^{\top}\rva^i/\tau} + \frac{1}{K}\sum_{j=1}^{K}e^{(\rva^i)^{\top}\rva^{j}/\tau}\right)\right]\\
        &= -\frac{1}{M\tau}\mathop{\mathbb{E}}_{\substack{\rmX^i \sim p_{\text{data}}}} \left[\sum_{m=1}^M\rvg_m(\rvx^i_m)^{\top}\rva^i\right] + \lambda\mathop{\mathbb{E}}_{\substack{\rmX^i \sim p_{\text{data}}}} \left[ \log \mathop{\mathbb{E}}_{\substack{ \rmX^{j}\sim p_{\text{data}} }}
    \left[ e^{(\rva^i)^{\top}\rva^{j}/\tau} \right]\right],
    \end{aligned}
\end{equation*}
where the exchange of limit and expectation follows from the \ac{dct} \citep{hypersphere}: the integrand converges almost surely by \cref{eq:limit} and it is dominated by a constant, as $e^{\rvu^{\top}\rvv/\tau} \in [e^{-1/\tau}, e^{1/\tau}]$ whenever $\|\rvu\|_2=\|\rvv\|_2=1$ and $\tau>0$.\\

For the convergence speed, we have
\begin{align}
&\left\vert  \lim_{K\rightarrow\infty} \left(\Ls_{\text{AnchorNCE}}(\rvg_1, \ldots, \rvg_M) - \lambda\log{K}\right) - \left(\Ls_{\text{AnchorNCE}}(\rvg_1, \ldots, \rvg_M)-\lambda\log{K}\right)\right\vert \nonumber\\
&= \left|\mathop{\mathbb{E}}_{\substack{\rmX^i \sim p_{\text{data}}}} \left[ \lambda\log \mathop{\mathbb{E}}_{\substack{ \rmX^{j}\sim p_{\text{data}} }} \left[ e^{(\rva^i)^{\top}\rva^{j}/\tau} \right]\right]-\left( \mathop{\mathbb{E}}_{\substack{ \rmX^i \sim p_{\text{data}} \nonumber\\ \{\rmX^{k}\}_{k=1}^{K} \overset{\mathrm{i.i.d.}}{\sim} p_{\text{data}}}} \left[  \lambda\log\left(e^{(\rva^i)^{\top}\rva^i/\tau} + \sum_{k=1}^{K}e^{(\rva^i)^{\top}\rva^{k}/\tau}\right)\right]-\lambda\log{K}\right)\right| \nonumber\\      
&= \left| \mathop{\mathbb{E}}_{\substack{ \rmX^i \sim p_{\text{data}} \\ 
\{\rmX^{k}\}_{k=1}^{K} \overset{\mathrm{i.i.d.}}{\sim} p_{\text{data}}}} \left[ \lambda \log \mathop{\mathbb{E}}_{\substack{ \rmX^{j}\sim p_{\text{data}}}} \left[ e^{(\rva^i)^{\top}\rva^{j}/\tau} \right] - \lambda\log\left( \frac{1}{K}e^{(\rva^i)^{\top}\rva^i/\tau}+\frac{1}{K}\sum_{k=1}^{K}e^{(\rva^i)^{\top}\rva^{k}/\tau} \right) \right] \right| \nonumber\\
& \leq \mathop{\mathbb{E}}_{\substack{ \rmX^i \sim p_{\text{data}} \\ \{\rmX^{k}\}_{k=1}^{K} \overset{\mathrm{i.i.d.}}{\sim} p_{\text{data}}}} \left[\left| \lambda\log \mathop{\mathbb{E}}_{\substack{ \rmX^{j}\sim p_{\text{data}}}} \left[e^{(\rva^i)^{\top}\rva^{j}/\tau}\right]- \lambda\log\left( \frac{1}{K}e^{(\rva^i)^{\top}\rva^i/\tau}+\frac{1}{K}\sum_{k=1}^{K}e^{(\rva^i)^{\top}\rva^{k}/\tau} \right)\right|\right] \nonumber\\
& \leq \lambda e^{1/\tau} \mathop{\mathbb{E}}_{\substack{ \rmX^i \sim p_{\text{data}} \\ \{\rmX^{k}\}_{k=1}^{K} \overset{\mathrm{i.i.d.}}{\sim} p_{\text{data}}}} \left[\left| \mathop{\mathbb{E}}_{\substack{ \rmX^{j}\sim p_{\text{data}}}} \left[ e^{(\rva^i)^{\top}\rva^{j}/\tau} \right] - \left( \frac{1}{K}e^{(\rva^i)^{\top}\rva^i/\tau}+\frac{1}{K}\sum_{k=1}^{K}e^{(\rva^i)^{\top}\rva^{k}/\tau} \right) \right|\right] \nonumber\displaybreak[4]\\
&\leq \lambda e^{1/\tau} \mathop{\mathbb{E}}_{\substack{ \rmX^i \sim p_{\text{data}} \\ \{\rmX^{k}\}_{k=1}^{K} \overset{\mathrm{i.i.d.}}{\sim} p_{\text{data}}}} \left[ \left| \mathop{\mathbb{E}}_{\substack{ \rmX^{j}\sim p_{\text{data}}}}\left[ e^{(\rva^i)^{\top}\rva^{j}/\tau} \right] - \frac{1}{K}\sum_{k=1}^{K}e^{(\rva^i)^{\top}\rva^{k}/\tau}\right| \right] +  \frac{\lambda}{K}e^{2/\tau},\label{eq:conv_rate}
\end{align}
where both $\rmX^j$ and $\rmX^k$ denote negative samples, and distinct superscripts are used to indicate that they originate from different sampling processes. The first inequality follows from Jensen's inequality. The second follows from the Mean Value Theorem: for a function $f$ that is continuous on $[a,b]$ and differentiable on $(a,b)$, there exists $c \in (a,b)$ such that $f(a)-f(b) = f'(c)(a-b)$. Hence, if $|f'| \leq U$ on that interval, then $|f(a)-f(b)| = |f'(c)|\,|a-b| \leq U\,|a-b|$. We apply this to $f=\log$ on the interval $[e^{-1/\tau},\,(1+\tfrac{1}{K})e^{1/\tau}]$, which contains the arguments of both logarithms. In this case, $f'(x) = 1/x \leq e^{1/\tau}$, so $U = e^{1/\tau}$. The third inequality follows from the triangle inequality, together with
\begin{equation*}
    \mathop{\mathbb{E}}_{\substack{ \rmX \sim p_{\text{data}} \\ \{\rmX^{k}\}_{k=1}^{K} \overset{\mathrm{i.i.d.}}{\sim} p_{\text{data}}}}\left[\left|\frac{1}{K}e^{(\rva^i)^{\top}\rva^i/\tau}\right|\right] = \frac{1}{K}e^{1/\tau},
\end{equation*}
which holds because $\|\rva^i\|_2 = 1$.\\

For a fixed positive sample $\mathbf{X}^i$, we can define the population mean $\mu:=\mathop{\mathbb{E}}_{\substack{ \rmX^{j}\sim p_{\text{data}}}}\left[ e^{(\rva^i)^{\top}\rva^{j}/\tau} \right]$, which is constant once $\mathbf{X}^i$ is fixed. Then,
\begin{equation*}
\mathop{\mathbb{E}}_{\substack{ \rmX^{j}\sim p_{\text{data}}}}\left[ e^{(\rva^i)^{\top}\rva^{j}/\tau} \right] - \frac{1}{K}\sum_{k=1}^{K}e^{(\rva^i)^{\top}\rva^{k}/\tau} = \mu - \frac{1}{K}\sum_{k=1}^{K}e^{(\rva^i)^{\top}\rva^{k}/\tau},
\end{equation*}
which is the difference between the population mean and a sample mean of $K$ independent draws.  If we define $Y_k = e^{(\rva^i)^{\top}\rva^{k}/\tau} - \mu$, then $\mathbb{E}[Y_k]=0$ and
\begin{equation*}
\left|\mu - \frac{1}{K}\sum_{k=1}^{K}e^{(\rva^i)^{\top}\rva^{k}/\tau}\right|= \left|-\left( \frac{1}{K}\sum_{k=1}^{K}(e^{(\rva^i)^{\top}\rva^{k}/\tau}-\mu) \right)\right| = \left|\frac{1}{K}\sum_{k=1}^{K}Y_k\right|.
\end{equation*}
Therefore, for fixed $\mathbf{X}^i$, the random variables $Y_k$ are i.i.d., have zero mean, and satisfy $|Y_k|\leq c$ and $\sigma_Y^2=\mathbb{E}[Y_k^2]\leq c^2$, where $c=e^{1/\tau}-e^{-1/\tau}$. These bounds follow from the fact that both $e^{(\rva^i)^{\top}\rva^{k}/\tau}$ and $\mu$ lie in $[e^{-1/\tau},e^{1/\tau}]$. 

If we now rewrite the expectation in \cref{eq:conv_rate} as 
\begin{equation}
    \mathbb{E}\left[\left|\frac{1}{K}\sum_{k=1}^{K}Y_k\right|\right] = \frac{\sigma_Y}{\sqrt{K}} \mathbb{E}\left[\left|\frac{1}{\sigma_Y\sqrt{K}}\sum_{k=1}^{K}Y_k\right|\right],
\end{equation}
we get an expectation of a random variable $W_K = \frac{1}{\sigma_Y\sqrt{K}}\sum_{k=1}^{K}Y_k$ with zero mean and unit variance. By the tail integral formula for nonnegative random variables, $\mathbb{E}[|W_K|]=\int_0^\infty P(|W_K|>x)\,dx$. Since
\begin{equation*}
    |W_K| =   \left|\frac{1}{\sigma_Y\sqrt{K}}\sum_{k=1}^{K}Y_k \right| \leq \frac{1}{\sigma_Y\sqrt{K}}\sum_{k=1}^{K}|Y_k| \leq \frac{Kc}{\sigma_Y\sqrt{K}} = \frac{c\sqrt{K}}{\sigma_Y},
\end{equation*}
the integrand vanishes for $x>c\sqrt{K}/\sigma_Y$, and therefore
\begin{equation}
\frac{\sigma_Y}{\sqrt{K}}\mathbb{E}\left[\left|\frac{1}{\sigma_Y\sqrt{K}}\sum_{k=1}^{K}Y_k\right|\right] = \frac{\sigma_Y}{\sqrt{K}}\int_0^{\tfrac{c\sqrt{K}}{\sigma_Y}}P\left(\left|\frac{1}{\sigma_Y\sqrt{K}}\sum_{k=1}^{K}Y_k\right|>x\right)dx,
\end{equation}

Now, following the Berry-Esseen Theorem as in \citet{hypersphere}, we have
\begin{equation}
\begin{aligned}
&\frac{\sigma_Y}{\sqrt{K}}\int_0^{\tfrac{c\sqrt{K}}{\sigma_Y}}P\left(\left|\frac{1}{\sigma_Y\sqrt{K}}\sum_{k=1}^{K}Y_k\right|>x\right)dx \\
&\leq \frac{\sigma_Y}{\sqrt{K}}\int_0^{\tfrac{c\sqrt{K}}{\sigma_Y}} \left(P\left(\left|\mathcal{N}(0,1)\right|>x\right)+\frac{C_c}{\sqrt{K}}\right)dx \\
& = \frac{\sigma_Y}{\sqrt{K}}\left(\frac{cC_c}{\sigma_Y}+\int_0^{\tfrac{c\sqrt{K}}{\sigma_Y}}P\left(\left|\mathcal{N}(0,1)\right|>x\right)dx\right) \\
& \leq \frac{\sigma_Y}{\sqrt{K}}\left( \frac{cC_c}{\sigma_Y}+\int_0^{\infty}P\left(\left|\mathcal{N}(0,1)\right|>x\right)dx \right)\\
& = \frac{\sigma_Y}{\sqrt{K}}\left( \frac{cC_c}{\sigma_Y} + \mathbb{E}[|\mathcal{N}(0,1)|]\right).
\end{aligned}
\end{equation}
Since the variance of $Y_k$ is bounded as $\sigma^2_Y = \mathbb{E}[Y^2_k]\leq c^2$, we have
\begin{equation}
\frac{\sigma_Y}{\sqrt{K}}\left( \frac{cC_c}{\sigma_Y} + \mathbb{E}[|\mathcal{N}(0,1)|]\right) \leq \frac{cC_c}{\sqrt{K}}+ \frac{c}{\sqrt{K}}\mathbb{E}[|\mathcal{N}(0,1)|] = \mathcal{O}(K^{-1/2}),
\end{equation}
\noindent
where the constant $C_c$ only depends on $c$. Therefore, the convergence speed of \cref{eq:thm1} can be expressed as
\begin{equation}
\left\vert \lim_{K\rightarrow\infty} \left(\Ls_{\text{AnchorNCE}}(\rvg_1, \ldots, \rvg_M) - \lambda\log{K}\right) - \left(\Ls_{\text{AnchorNCE}}(\rvg_1, \ldots, \rvg_M)-\lambda\log{K}\right)\right\vert \leq \frac{\lambda}{K}e^{2/\tau}+\mathcal{O}(K^{-1/2})
\end{equation} \\

\textbf{Proof of result~\hyperref[thm1_ap]{1}: the first term is minimized if and only if all modality-specific encoders are perfectly aligned.}\\

Recall that $\rva^i = \tilde{\rva}^i/\|\tilde{\rva}^i\|_2$, with $\tilde{\rva}^i=\tfrac{1}{M}\sum_{m=1}^{M}\rvg_m(\rvx^i_m)$, and that all modality-specific encoders output unit-norm representations. For any two unit-norm vectors $\rvu,\rvv$, we have $\rvu^{\top}\rvv \in [-1,1]$, with $\rvu^\top\rvv=1$ if and only if $\rvu=\rvv$, since
\begin{equation*}
\|\rvu - \rvv\|_2^2 = 2 - 2\rvu^{\top}\rvv.
\end{equation*}
Since $\tau>0$, the first term of \cref{eq:thm1} is therefore minimized if and only if
\begin{equation*}
\rvg_m(\rvx^i_m)^{\top}\rva^i = 1
\qquad
\forall m \in \{1,\ldots,M\},
\end{equation*}
that is, if and only if $\rvg_m(\rvx^i_m) = \rva^i$ for every modality $m$. In this case, $\tilde{\rva}^i$ is the average of $M$ identical unit-norm vectors, so it is itself unit-norm and the normalization leaves it unchanged. Hence
\begin{equation*}
\rva^i= \tilde{\rva}^i = \frac{1}{M}\sum_{m=1}^M \rvg_m(\rvx^i_m) = \rvg_1(\rvx^i_1) = \ldots = \rvg_M(\rvx^i_M).
\end{equation*}\\

\textbf{Proof of result~\hyperref[thm1_ap]{2}: if modality-specific encoders exist such that the anchors are perfectly uniform, they are the exact minimizers of the second term.}\\

We adapt the proof strategy of \citet{hypersphere}. Let $\mathcal{S}^{d}$ denote the unit hypersphere in $\mathbb{R}^{d+1}$ and $\mathcal{M}(\mathcal{S}^d)$ the set of Borel probability measures on $\mathcal{S}^d$.

\begin{definition}
$\forall\mu\in\mathcal{M}(\mathcal{S}^d)$, $\rvu\in\mathcal{S}^d$, we define the continuous and Borel measurable function
\begin{equation}
U_{\mu}(\rvu) \triangleq \int_{\mathcal{S}^d}e^{\rvu^{\top}\rvv/\tau}\text{\emph{d}}\mu(\rvv)
\end{equation}
with its range bounded in $[e^{-1/\tau}, e^{1/\tau}]$, since $\rvu^\top\rvv\in[-1,1]$ for all $\rvu,\rvv\in\mathcal S^d$.
\end{definition}

Let $T:\mathcal{X}_1\times\ldots\times\mathcal{X}_M\to\mathbb{R}^{d+1}$ and $Q:\mathbb{R}^{d+1}\to\mathcal{S}^{d}$ be
\begin{equation*}
\begin{gathered}
    T(\rvx_1,\ldots,\rvx_M)=\frac{1}{M}\sum_{m=1}^M\rvg_m(\rvx_m),
    \\
    Q(\rvx)=\rvx/\|\rvx\|_2.
\end{gathered}
\end{equation*}
Then, we can define the composed map $F=Q\circ T:\mathcal{X}_1\times\ldots\times\mathcal{X}_M\to\mathcal{S}^{d}$
so that $\rva^i = F(\rvx^i_1,\ldots,\rvx^i_M)$. Let $p_{\text{data}}\circ F^{-1}\in\mathcal M(\mathcal S^{d})$ denote the pushforward of $p_{\text{data}}$ under $F$. Since $\rva^j=F(\rvx_1^j,\ldots,\rvx_M^j)$ for an independent sample $\rmX^j\sim p_{\text{data}}$,  the inner expectation in the second term is, by the law of the unconscious statistician,
\begin{equation*}
\mathop{\mathbb{E}}_{\rmX^j\sim p_{\text{data}}}\!\left[e^{(\rva^i)^\top\rva^j/\tau}\right] = \int_{\mathcal S^{d}} e^{(\rva^i)^\top\rva^j/\tau}\,\text{d}(p_{\text{data}}\circ F^{-1})(\rva^j) = U_{p_{\text{data}}\circ F^{-1}}(\rva^i).
\end{equation*}
Therefore, the second term in \cref{eq:thm1} can be written as
\begin{equation}
\begin{aligned}
&\lambda\mathop{\mathbb{E}}_{\rmX^i\sim p_{\text{data}}}\left[\log \mathop{\mathbb{E}}_{\rmX^j\sim p_{\text{data}}}\left[e^{(\rva^i)^{\top}\rva^{j}/\tau}\right]\right]\\
&= \lambda\mathop{\mathbb{E}}_{\rmX^i\sim p_{\text{data}}}\left[\log U_{p_{\text{data}}\circ F^{-1}}(F(\rvx^i_1,\ldots,\rvx^i_M))\right] \label{eq:second-term-measure}
\end{aligned}
\end{equation}

Following \citet{hypersphere}, we now consider the relaxed problem obtained by minimizing over \emph{all} Borel probability measures on the hypersphere $\mathcal S^d$:
\begin{equation}
\min_{\mu\in\mathcal{M}(\mathcal{S}^d)}\int_{\mathcal{S}^d}\log U_{\mu}(\rvu)\,\text{d}\mu(\rvu). \label{eq:relaxed}
\end{equation}
The proof of \citet{hypersphere} applies verbatim to \cref{eq:relaxed}, since it is a statement purely about measures on the hypersphere $\mathcal S^d$ and does not reference how $\mu$ arises: the unique minimizer is the uniform (surface) measure $\sigma_d$ on $\mathcal{S}^d$. \\

Consequently, if modality-specific encoders $\rvg_1,\ldots,\rvg_M$ exist such that $p_{\text{data}}\circ F^{-1} = \sigma_{d_z-1}$, i.e., such that the induced anchors are perfectly uniform on the hypersphere $\mathcal{S}^{d_z-1}$, these encoders attain the minimum of the relaxed problem, and are therefore exact minimizers of
\begin{equation*}
\min_{\rvg_1,\ldots,\rvg_M} \lambda\mathop{\mathbb{E}}_{\rmX^i\sim p_{\text{data}}}\left[\log \mathop{\mathbb{E}}_{\rmX^j\sim p_{\text{data}}}\left[e^{(\rva^i)^{\top}\rva^{j}/\tau}\right]\right].
\end{equation*}

\subsection{Proof of Proposition 2}
\label{app_proof_thm2}

\textbf{Assumptions on the data generation process}\\
First, we formalize our assumptions on the underlying data generating process, which serves as the basis for our proposed methods and theoretical guarantees. Following \citet{ident_multimodal} and \citet{ident_multiview}, we formalize the multimodal data generation process as a latent variable model, extending their formulation to an arbitrary number of modalities $M$. Specifically, we assume a continuous latent variable $\rvz$ taking values in a $d$-dimensional space $\mathcal{Z}\subseteq\sR^{d}$, with probability density $p_{\rvz}$, which contains all the information required to generate observations across the $M$ modalities. We further assume that $\rvz$ can be decomposed into: (i) an invariant component $\rvs$, which captures information shared across all modalities, and (ii) $M$ modality-specific (private) components $\rvw_m$, one for each modality $m\in\{1,\ldots,M\}$. Note that this decomposition is a common underlying assumption in shared-private representation learning methods.

Each modality-specific observation $\rvx_m$, taking values in $\mathcal{X}_m\subseteq\sR^{D_m}$, is then generated by a modality-specific mixing function $\rvf_m:\mathcal{Z}_m\rightarrow\mathcal{X}_m$ applied to its corresponding latent representation $\rvz_m=(\rvs_m,\rvw_m)$, which takes values in $\mathcal{Z}_m \subseteq \mathbb{R}^{d_z}$:
\begin{equation} \tag{\ref{eq:gen_model}}
\rvz \sim p_{\rvz}, \qquad \rvx_m = \rvf_m(\rvz_m).
\end{equation}
We assume that the shared latent components satisfy $\rvs_m=\rvs_n$, almost everywhere for all $m\neq n$, where $m, n\in\{1,\dots,M\}$. Each mixing function $\rvf_m$ is assumed to be a \emph{smooth invertible mapping} with a smooth inverse, i.e., a diffeomorphism, between the latent representation and observation spaces. In general, different modalities are generated by distinct mixing functions applied to their respective modality-specific latent variables.

Following \citet{ident_unimodal}, \citet{ident_multimodal} and \citet{ident_multiview}, we formalize shared information across modalities, i.e., content invariance, through the conditional distribution $p_{\rvz_m|\rvz_n}$ for $m\neq n$, which characterizes the relationship between the corresponding latent variables.

\begin{assumption}[Content-invariance]\label{content_invariance}
The conditional density $p_{{\rvz_m|\rvz_n}}$ over $\mathcal{Z}_m\times\mathcal{Z}_n$, with $m\neq n$, takes the form 
\begin{equation}
p_{\rvz_m|\rvz_n}(\rvz_m|\rvz_n) = \delta(\rvs_m - \rvs_n)p_{\rvw_m}(\rvw_m)
\end{equation}
where $\delta(\cdot)$ is the Dirac delta function, i.e., $\rvs_m = \rvs_n$, a.e.
\end{assumption}

\begin{proposition}[Identifying shared information]
Assume the data generating process described in Section \ref{sec:gen_process}, where the different modalities are generated from \cref{eq:gen_model}, with $p_{\rvz_{m}|\rvz_{n}}$ as defined in Assumption \ref{content_invariance}. Further, assume that $p_{\rvz}$ is a smooth and continuous density on $\mathcal{Z}$ with $p_\rvz(\rvz)>0$ almost everywhere. Let $\rvg_m: \mathcal{X}_m\rightarrow(0,1)^{d_s}$, $m\in\{1,\ldots,M\}$, be smooth functions that minimize $\mathcal{L}_{\text{AnchorAlignEnt}}$ as defined in \cref{eq:anchoralignmaxent}. Then $\rvg_1, \ldots, \rvg_M$ block-identify the true shared variables in the sense of Definition \ref{def:blockident}.
\end{proposition}

\emph{Proof of Proposition 2.} We follow the proof outline of Theorem 4.4 in \citet{ident_unimodal} and Theorem 1 in \citet{ident_multimodal}, organizing the proof into three steps. \\

\noindent
\textbf{Step 1. Show that there exist $M$ smooth functions $\rvg_1, \ldots, \rvg_M$ that attain the global minimum of $\mathcal{L}_{\text{AnchorAlignEnt}}$.}\\

Following prior work \citep{ident_unimodal, ident_multimodal, ident_multiview}, we assume that, in the limit of infinite negative samples, the asymptotic \loss{} behaves as 
\begin{equation}
\begin{aligned}
&\mathcal{L}_{\text{AnchorAlignEnt}} = \mathop{\mathbb{E}}_{\mathbf{X}^i\sim p_{\text{data}}}\left[\frac{1}{M}\sum_{m=1}^M\|\rvg_m(\rvx^i_m) - \tilde{\rva}^i\|_2^2\right] - \lambda H(\tilde{\rva}),
\end{aligned}\tag{\ref{eq:anchoralignmaxent}}
\end{equation}
where $H(\tilde{\rva})$ denotes the differential entropy of the aggregated anchor. Note that \cref{eq:anchoralignmaxent} operates in the unnormalized embedding space, which in \cref{eq:anchorloss} is equivalent to replacing the dot product with cosine similarity. The global minimum of this objective is reached when the first term is minimized and the second term is maximized. The first term is minimized when all the modality-specific encoders are aligned, that is, 
\begin{equation*}
\tilde{\rva}^i = \frac{1}{M}\sum_{m=1}^M \rvg_m(\rvx^i_m) = \rvg_1(\rvx^i_1) = \ldots = \rvg_M(\rvx^i_M),
\end{equation*}
while the second term is maximized when the aggregated anchor $\tilde{\mathbf{a}}$ induced by the $M$ modality-specific encoders is uniformly distributed over $(0,1)^{d_s}$.\\

For each modality $m\in\{1,\ldots,M\}$, let $\rvf_{m,1:d_s}^{-1}$ be the inverse of the true mixing function $\rvf_m$ restricted, without loss of generality, to the dimensions encoding the shared information, which we index as $1:d_s$. Since each $\rvf_m$ is a diffeomorphism by assumption, its inverse exists and is smooth. Moreover, we have $\rvf_{m,1:d_s}^{-1}=\rvs_m$ by definition and, by Assumption \ref{content_invariance}, $\rvs_1=\rvs_2=\ldots=\rvs_M$ almost surely. That is, the shared latent variables are identical across modalities almost surely under the generative process described in Section \ref{sec:gen_process}. Next, we define a function $\rvd$ which maps each $\rvs_m$ to a uniform random variable on $(0,1)^{d_s}$ using the Darmois construction \citep{ident_unimodal, ident_multimodal, darmois1951analyse, hyvarinen2016unsupervised}. Note that the Darmois construction is determined entirely by the distribution of its input. Therefore, given Assumption \ref{content_invariance}, one single map can be used across all modalities. Since we assume that $p_{\rvz}$ is a smooth density, every function $\rvd$ is also smooth. \\

Finally, let $\rvg_m^* = \rvd \circ \rvf_{m,1:d_s}^{-1}$. Each $\rvg_m^*$ is smooth since it is a composition of two smooth functions. The corresponding aggregated anchor is given by  $\tilde{\rva}^* = \frac{1}{M}\sum_{m=1}^M \rvg_m^*(\rvx_m) = \rvd(\rvs_m)$. Any set of functions $\rvg_1^*, \ldots, \rvg_M^*$ constructed in this manner attains the global minimum of the objective $\mathcal{L}_{\text{AnchorAlignEnt}}$. In particular,
\begin{equation*}
\begin{aligned}
\mathcal{L}_{\text{AnchorAlignEnt}}(\rvg^*_1, \ldots, \rvg^*_M) &= \mathop{\mathbb{E}}_{\mathbf{X}^i\sim p_{\text{data}}}\left[\frac{1}{M}\sum_{m=1}^M\|\rvg^*_m(\rvx^i_m) - \tilde{\rva}^{i*}\|_2^2\right] - \lambda H(\tilde{\rva}^*)\\
&= \mathop{\mathbb{E}}_{\mathbf{X}^i\sim p_{\text{data}}}\left[\frac{1}{M}\sum_{m=1}^M\|\rvd(\rvs^i_m) - \tilde{\rva}^{i*}\|_2^2\right] - \lambda H(\tilde{\rva}^*)\\
&=0.
\end{aligned}
\end{equation*}

Applying the same Darmois construction $\rvd$ to the identical shared variables $\rvs_1=\rvs_2=\ldots=\rvs_M$ yields identical representations across all modalities. Consequently, their average coincides with each individual representation, making the first term equal to zero. Since the modality-specific representations are uniformly distributed on $(0,1)^{d_s}$, their aggregated anchor is also uniformly distributed on this domain. Therefore, the second term is also zero, as the uniform distribution on the unit hypercube has zero entropy \citep{ident_unimodal, ident_multimodal}.\\

Now, let $\rvg_m:\mathcal{X}_m\rightarrow(0,1)^{d_s}$, $m\in \{1,\ldots,M\}$ be any set of smooth functions that attains the global minimum of $\mathcal{L}_{\text{AnchorAlignEnt}}$, i.e.
\begin{equation*}
\begin{aligned}
\mathcal{L}_{\text{AnchorAlignEnt}}(\rvg_1, \ldots, \rvg_M) = \mathop{\mathbb{E}}_{\mathbf{X}^i\sim p_{\text{data}}}\left[\frac{1}{M}\sum_{m=1}^M\|\rvg_m(\rvx^i_m) - \tilde{\rva}^i\|_2^2\right] - \lambda H(\tilde{\rva}) = 0.
\end{aligned}
\end{equation*}

Let $\rvh_m \coloneqq \rvg_m \circ \rvf_m : \mathcal{Z}_m \rightarrow (0,1)^{d_s}$. For every $m$, $\rvh_m$ is smooth, as it is the composition of two smooth functions. Writing $\rvx_m=\rvf_m(\rvz_m)$, this implies that
\begin{align}
\mathop{\mathbb{E}}_{\mathbf{X}^i\sim p_{\text{data}}}\left[\frac{1}{M} \sum_{m=1}^M\|\rvh_m(\rvz^i_m)-\tilde{\rva}^i\|_2^2\right] = 0, \label{eq:aligment_proof}\\
H(\tilde{\rva}) = 0.
\end{align}

Therefore, $\rvh_1(\rvz_1)=\ldots=\rvh_M(\rvz_M)=\rva$ must hold almost everywhere with respect to the generative model described in Section \ref{sec:gen_process}. Moreover, all aligned encoder outputs, and consequently $\tilde{\rva}$, must be uniformly distributed on $(0,1)^{d_s}$.\\

\noindent
\textbf{Step 2. Show that any set of functions that minimize $\mathcal{L}_{\text{AnchorAlignEnt}}$ depends only on the (invariant) shared information across modalities.}\\

Following \citet{ident_unimodal} and \citet{ident_multimodal}, suppose \emph{for a contradiction} that $\rvh_m(\rvz_m)\coloneqq\rvh_m(\rvs_m,\rvw_m) : \mathcal{Z}_m \rightarrow (0,1)^{d_s}$ depends on at least one component of the private latent $\rvw_m$. In particular, suppose that the partial derivative of $\rvh_m(\rvz_m)$ w.r.t. some modality-specific variable $w_{m,k}$ is non-zero for some point $(\rvs_m^*, \rvw_m^*)\in\mathcal{Z}_m$. That is,
\begin{gather*}
    \exists k \in \{1,\ldots,d_w\}, (\rvs_m^*, \rvw_m^*)\in\mathcal{Z}_s\times\mathcal{Z}_w,\\ 
    \text{s.t.} \quad \frac{\partial\rvh_m(\rvz_m)}{\partial w_{m,k}} \neq 0.
\end{gather*}

Since $\rvh_m$ is smooth, its first partial derivatives are continuous. Therefore, if $\frac{\partial\rvh_m(\rvz_m)}{\partial w_{m,k}}$ is non-zero at $(\rvs_m^*,\rvw_m^*)$, it remains non-zero in a neighborhood of this point. This neighborhood is non-empty and open, and since $p_{\rvz}(\rvz)>0$ almost everywhere, it has positive probability under the generative model. Hence, there exists an open set of points in which
\begin{equation}\label{eq:contradiction}
    |\rvh_m(\rvz_m)-\rvh_n(\rvz_n)|= |\rvh_m(\rvs_m, \rvw_m)-\rvh_n(\rvs_n, \rvw_n)|>0, \quad m,n\in\{1,\ldots,M\},\quad m\neq n.
\end{equation}

Since the modality-specific latent variables are independent and $\rvh_m(\rvz_m)$ cannot depend on $\rvw_n$ under our assumed generative model, \cref{eq:contradiction} contradicts the invariance condition in \cref{eq:aligment_proof}. Therefore, any encoder that minimizes $\mathcal{L}_{\text{AnchorAlignEnt}}$ cannot depend on modality-specific information.\\

\noindent
\textbf{Step 3. Show that the mappings $\rvh_m$ are invertible.}\\

To this end, following \citet{ident_unimodal} and \citet{ident_multimodal}, we make use of Proposition 5 in \citet{zimmermann2021contrastive}.  We know that $\mathcal{Z}_s$ and $(0,1)^{d_s}$ are simply connected and oriented $\mathcal{C}^1$ manifolds of the same dimension, and the maps $\rvh_1,\ldots,\rvh_M$ are smooth, and hence differentiable, as shown in Step 1. Moreover, $p_{\rvs}$ is a regular density, and the pushforward of $p_{\rvs}$ through each $\rvh_m$ is the uniform distribution on $(0,1)^{d_s}$, which is also a regular density. Hence, $\rvh_1,\ldots,\rvh_M$ are bijections \citep{zimmermann2021contrastive, ident_unimodal, ident_multimodal}.\\

This concludes the proof. We have shown that any smooth functions $\rvg_1,\ldots,\rvg_M$ that attain the global minimum of $\mathcal{L}_{\text{AnchorAlignEnt}}$ satisfy $\rvg_m(\rvx_m)=\rvh_m(\rvs_m)$, with $\rvh_m$ a bijection for every $m$. Therefore, the shared information $\rvs$ is \emph{block-identified} by $\rvg_1,\ldots,\rvg_M$ in the sense of Definition \ref{def:blockident}.

\section{Experimental details}
\label{app:exp_details_1}

\subsection{Datasets}

\begin{table}[t]
    \centering
    \vspace{-0.05in}
    \caption{\textbf{Dataset statistics.} Modalities and number of samples per split. For MS-COCO and RoCOCO, we report images / captions. RoCOCO counts include the adversarial samples of one set.}
    \label{tab:datasets}
    \small
    \begin{tabular}{rcccc}
        \toprule
        \textbf{Dataset} & \textbf{Modalities} & \textbf{Train} & \textbf{Val} & \textbf{Test} \\
        \midrule
        Clevr-4~\citep{vaze2023no}     & image, text
                    & 72,011  & 8,001  & 19,988 \\
        \midrule
        VGGSound~\citep{chen2020vggsound}    & audio, video
                    & 164,389 & 18,225 & -- \\
        VGGSounder~\citep{zverev2025vggsounder}  & audio, video
                    & --      & --     & 12,014 \\
        \midrule
        MS-COCO~\citep{lin2014microsoft}     & image, text
                    & 113,287 / 566,747
                    & 5,000 / 25,010
                    & 5,000 / 25,010 \\
        RoCOCO (I2T)~\citep{park2024rococo} & image, text
            & -- & -- & 5,000 / 50,010 \\
        RoCOCO (T2I)~\citep{park2024rococo} & image, text
            & -- & -- & 10,000 / 25,010 \\
        \midrule
        VAST-150K~\citep{chen2023vast}   & video, audio, text
                    & 140,208 & --  & -- \\
        MSR-VTT~\citep{xu2016msr}     & video, audio, text
                    & --      & --     & 1,000 \\
        ActivityNet~\citep{caba2015activitynet} & video, audio, text
                    & --      & --     & 3,987 \\
        VATEX~\citep{wang2019vatex}       & video, audio, text
                    & --      & --     & 1,224 \\
        \bottomrule
    \end{tabular}
    \vspace{-0.05in}
\end{table}

\begin{figure*}[t]
    \centering
    \begin{subfigure}[t]{0.48\textwidth}
        \centering
        \includegraphics[width=\linewidth]{figures/clevr4/clevr4_exp_1.pdf}
    \end{subfigure}
    \hfill
    \begin{subfigure}[t]{0.48\textwidth}
        \centering
        \includegraphics[width=\linewidth]{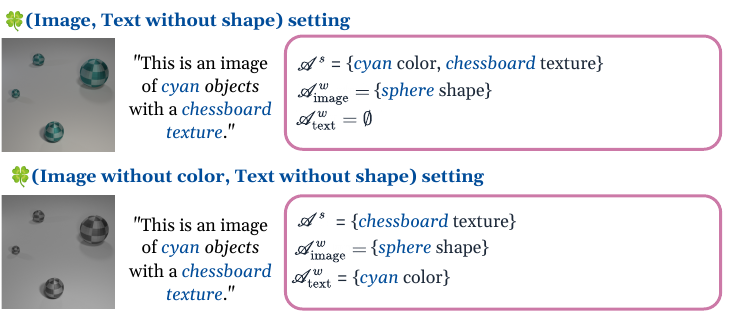}
    \end{subfigure}

    \caption{\textbf{Four Clevr-4 settings.}
    Shared attributes $\mathscr{A}^s$ and private attributes $\mathscr{A}^w_m$ for each modality $m$.}
    \label{fig:clevr4_exp_full}
    \vspace{-0.1in}
\end{figure*}

This section describes the datasets used in our experiments, whose modalities and split sizes are summarized in \Cref{tab:datasets}. Since not all video clips remain available, the table reports the number of clips we obtained. 

\vspace{0.08in} \noindent \textbf{Clevr-4.} Clevr-4~\citep{vaze2023no} is a synthetic image dataset in which all objects in an image share the same \emph{shape}, \emph{color}, and \emph{texture}, each sampled independently from 10 classes. Clevr-4 additionally annotates the number of objects, which we do not use. To obtain a second modality, we pair each image with a template-based caption describing its attributes:
\begin{center}
  \texttt{This is an image of \{color\} \{shape\}s
  with a \{texture\} texture.}
\end{center}
To control which information is shared across modalities, we optionally remove color from images via grayscale conversion and shape from captions, in which case the caption becomes 
\begin{center}
  \texttt{This is an image of \{color\} objects
  with a \{texture\} texture.}
\end{center}
Removing an attribute from one modality makes it private to the other. This yields the four settings summarized in~\Cref{fig:clevr4_exp_full}.

\vspace{0.08in} \noindent \textbf{VGGSounder.} VGGSounder~\cite{zverev2025vggsounder} is a multi-label audio-visual benchmark that re-annotates the test set of VGGSound~\citep{chen2020vggsound}, a dataset of 10-second YouTube clips covering 309 classes. Each label of each clip is marked as audible, visible, or both. We train all models on the VGGSound training set and evaluate them on VGGSounder, which does not overlap with the training data. Following \citet{zverev2025vggsounder}, we use the subset without background music. To obtain modality-specific targets, we partition the labels by their modality annotation. Labels that are audible but not visible, visible but not audible, or both define the audio-private, video-private, and shared tasks, respectively. Examples are shown in~\Cref{fig:vggsounder_qualitative_results_1,fig:vggsounder_qualitative_results_2}.

\vspace{0.08in} \noindent \textbf{RoCOCO.} RoCOCO~\citep{park2024rococo} is a robustness benchmark built on the MS-COCO 5K test split~\citep{lin2014microsoft} that adds adversarial captions and images to the retrieval galleries. We train all models on the MS-COCO Karpathy training split~\citep{karpathy2015deep} and evaluate on its test split and RoCOCO. \textbf{\emph{Adversarial captions}} replace one word of the original with a random, same-concept, different-concept, or danger-related word, yielding the Rand-voca, Same-concept, Diff-concept, and Danger sets. For image-to-text retrieval, each set extends the 25,010 original captions with 25,000 adversarial ones. \textbf{\emph{Adversarial images}} blend a randomly chosen test image into each test image with weight $\rho$ on the original (Mix), or paste a patch of it covering a $(1-\rho)$ fraction of the area (Patch), where $\rho \in \{0.9, 0.8\}$. For text-to-image retrieval, each of the four perturbations adds 5,000 images to the 5,000 originals, and results are averaged over three generation seeds. Examples are shown in~\Cref{fig:rococo_exp}.

\vspace{0.08in} \noindent \textbf{VAST-150K.} VAST-150K~\citep{cicchetti2025gramian, cicchetti2026triangle} is a 150K-sample subset of VAST-27M~\citep{chen2023vast}, a large-scale dataset of open-domain video clips with automatically generated captions, released by Cicchetti et al.~\citep{cicchetti2025gramian}. Each sample pairs a video clip, \ie, its frames and audio track, with a single caption, yielding three modalities: video, audio, and text. Following GRAM~\citep{cicchetti2025gramian} and TRIANGLE~\citep{cicchetti2026triangle}, we evaluate zero-shot retrieval on MSR-VTT~\citep{xu2016msr}, ActivityNet~\citep{caba2015activitynet}, and VATEX~\citep{wang2019vatex}. For ActivityNet and VATEX, we use the test videos released by CalMRL~\citep{pmlr-v306-liu26ex}, keeping only those with an audio track.

\subsection{Evaluation}

\noindent \textbf{Linear probing.} We evaluate the learned shared and private representations with linear probes. Each representation is standardized with training-set statistics, and a logistic-regression probe ($\ell_2$-regularized, $C{=}1$, at most $1000$ iterations) is fit on the training split and evaluated on the test split. On Clevr-4, we train one probe per attribute, \ie, color, shape, and texture, each with $10$ classes, and report accuracy (\%). On VGGSounder, we train a single probe on the $309$ VGGSound classes and evaluate it on the clips without background music against the labels of each task, \ie, shared, audio-private, or video-private, reporting micro F1@1 (\%) computed with the official VGGSounder benchmark code~\cite{zverev2025vggsounder}.

\vspace{0.08in} \noindent \textbf{Retrieval.} On MS-COCO and RoCOCO, we rank gallery items by the cosine similarity between the shared representations of the query and gallery modalities, except for FactorCL, which uses its critic. Following the original RoCOCO evaluation code~\cite{park2024rococo}, we report R@1 (\%), the relative drop (\%)
\begin{equation}
  \text{drop} = \frac{\text{R@1}_{\text{MS-COCO}}
  - \text{R@1}_{\text{RoCOCO}}}{\text{R@1}_{\text{MS-COCO}}} \times 100,
\end{equation}
and RSMS (\%), \ie, the percentage of queries for which an adversarial sample is ranked first. On VAST-150K, we follow the protocol of GRAM~\cite{cicchetti2025gramian} and TRIANGLE~\cite{cicchetti2026triangle} and report zero-shot R@1 for text-to-video and video-to-text retrieval, where each video comprises its frames and audio track. \loss{} scores each text--video pair by the cosine similarity between the text representation and the anchor of the frame and audio representations, whereas the two baselines use their volume- and area-based scores, respectively. For all methods, the top-$50$ candidates are then re-ranked with the VAST matching head~\cite{chen2023vast}.

\subsection{Implementation}
\label{app:implementation}

\begin{table}[t]
  \centering
  \vspace{-0.05in}
  \caption{\textbf{Common settings.} Used by all factorization methods unless stated otherwise in~\Cref{app:implementation}. Hidden dimensions and layers apply to all adapters and decoders. On Clevr-4, shared and private dimensions are varied jointly.}
  \small
  \begin{tabular}{rccc}
    \toprule
    & \textbf{Clevr-4}~\citep{vaze2023no} & \textbf{VGGSounder}~\citep{zverev2025vggsounder} & \textbf{RoCOCO}~\citep{park2024rococo} \\
    \midrule
    Embedding dimension ($\mathbf{e}_m$)      & $1024$ & 1024 & $512$ \\
    Shared dimension ($\hat{\mathbf{s}}_m$)     & $\{4, 8, 16, 32, 64\}$ & $256$ & $256$ \\
    Private dimension ($\hat{\mathbf{w}}_m$)    & $\{4, 8, 16, 32, 64\}$ & $64$  & $256$ \\
    Hidden dimension     & $256$ & $512$ & $512$ \\
    Number of layers          & $3$   & $3$   & $3$   \\
    Learning rate   & $10^{-4}$ & $5\times10^{-5}$ & $5\times10^{-5}$ \\
    Batch size ($B)$      & $64$  & $512$ & $1024$ \\
    Epochs          & $10$  & $10$  & $200$ \\
    Residual (shared adapter) & No & No & Yes \\
    \bottomrule
  \end{tabular}
  \label{tab:hparams}
  \vspace{-0.1in}
\end{table}

\begin{table*}[t]
  \centering
  \vspace{-0.05in}
  \caption{\textbf{Method-specific hyperparameters.} For each hyperparameter, we report the sweep range and the value used per dataset, where ``all'' denotes that every value of the sweep is shown in~\Cref{fig:av_single_f1}. Hyperparameter names follow each method's original paper or code.}
  \small
  \setlength{\tabcolsep}{2.5pt}
  \begin{tabular}{rccccc}
    \toprule
    \textbf{Method} & \textbf{Hyperparameter} & \textbf{Sweep} & \textbf{Clevr-4}~\citep{vaze2023no} & \textbf{VGGSounder}~\citep{zverev2025vggsounder} & \textbf{RoCOCO}~\cite{park2024rococo} \\
    \midrule
    \rowcolor{ourpurple}
    \method{} (ours)
      & $\lambda$ & $\{1, 4, 8, 16, 32, 64, 128\}$ & $1$
      & $\{1, 4, 8, 16, 32, 64, 128\}$ & $4$ \\
    \midrule
    FactorCL~\citep{liang2023factorized}
      & CLUB updates per step & -- & $1$ & $1$ & $1$ \\
    \midrule
    \multirow{4}{*}{DisentangledSSL~\citep{wang2025an}}
      & $\kappa$ & -- & $1000$ & $1000$ & $1000$ \\
      & $\beta$ & $\{0.01, 0.1, 1, 10, 100\}$ & $0.01$ & all & $0.01$ \\
      & $\lambda$ & $\{0, 0.001, 0.1\}$ & $0.001$ & $0.1$ & $0.001$ \\
      & Epochs (stage 1 / 2) & -- & $10$ / $6$ & $10$ / $6$
      & $200$ / $120$ \\
    \midrule
    \multirow{3}{*}{JointOpt~\citep{wang2025an}}
      & $\kappa$ & -- & $1000$ & $1000$ & $1000$ \\
      & $a$ & $\{0.01, 0.1, 1, 10, 100\}$ & $0.1$ & all & $1$ \\
      & $\lambda$ & $\{0, 0.001, 0.1\}$ & $0.001$ & $0.1$ & $0.001$ \\
    \midrule
    \multirow{3}{*}{RePercENT~\citep{rizou2026repercent}}
      & $\kappa$ & -- & $3000$ & $3000$
      & $3000$ \\
      & $\beta$ & -- & $2$ & $2$
      & $2$ \\
      & $(\alpha, \lambda)$ & $\{(2,1), (6,1), (6,4), (10,4)\}$
      & $(2, 1)$ & all & $(2, 1)$ \\
    \bottomrule
  \end{tabular}
  \label{tab:hparams_methods}
  \vspace{-0.05in}
\end{table*}

\noindent \textbf{Inputs and encoders.} All methods operate on embeddings precomputed with frozen encoders: \texttt{CLIP-RN50}~\citep{radford2021learning} on Clevr-4, \texttt{PE-AV-Base}~\citep{vyas2026pushing} on VGGSounder, and \texttt{CLIP-ViT-B/32}~\citep{radford2021learning} on RoCOCO. We use the projected embeddings, with \emph{two exceptions}: CLIP with fine-tuned projection heads uses the \texttt{CLS} (image) and \texttt{EOS} (text) token embeddings taken before CLIP's projections, and RePercENT~\citep{rizou2026repercent} uses the token sequences before pooling and projection. Inputs \emph{follow each encoder's official preprocessing}: on Clevr-4 and RoCOCO, images are resized and center-cropped to $224\times224$; on VGGSounder, 16 uniformly spaced frames per clip are resized to $336\times336$, and the mono waveform is resampled to 48\,kHz.

\vspace{0.08in} \noindent \textbf{Architecture and training.} 
All factorization methods use lightweight MLPs with ReLU activations as adapters and decoders, and \Cref{tab:hparams} lists the common settings. RePercENT~\citep{rizou2026repercent} instead uses \emph{its own attention-based architecture} with $2$ cross-attention heads, $1$ latent head, and a depth of $2$. On VGGSounder, FactorCL~\citep{liang2023factorized} and RePercENT~\citep{rizou2026repercent} use $256$ dimensions for both representations, as they do not support different dimensions. On RoCOCO, the shared adapters include a residual connection with a linear projection shortcut~\citep{resnet}, and the last MLP layer is initialized to zero so that training starts from this linear projection. For \method{}, the reconstruction targets are the $\ell_2$-normalized embeddings, and following CLIP~\citep{radford2021learning}, the temperature $\tau$ is learned, initialized to $0.07$, and lower-bounded by $0.01$. We optimize all trainable parameters with Adam~\citep{adam} using a single learning rate, and average all results over $5$ seeds. \Cref{tab:hparams_methods} lists the method-specific hyperparameters.

\vspace{0.08in}
\noindent
\textbf{Baselines.} Unlike \method{}, \emph{all factorization baselines use two augmented views} per sample as positives, which are precomputed before training by augmenting the raw inputs and encoding them with the frozen encoders. On Clevr-4 and RoCOCO, images are augmented with random resized crops with minimum scale $0.8$, horizontal flips, and brightness and contrast jitter of $0.2$, applied with probability $0.8$. For text, each view deletes one random template word on Clevr-4, while on RoCOCO, it is a different caption of the same image. On VGGSounder, each video view applies one of a random resized crop with minimum scale $0.8$, a horizontal flip, a color jitter, or a slight Gaussian blur, identically across all frames, and each audio view applies either a random gain within $\pm 3$\,dB or a zeroed segment of up to $0.25$\,s.

We integrate the official implementations of all factorization baselines into our pipeline, keeping their objectives and training procedures with the settings in~\Cref{tab:hparams,tab:hparams_methods}. FactorCL~\citep{liang2023factorized} uses concatenation-based critics to estimate mutual information during training and score similarity at retrieval. DisentangledSSL~\citep{wang2025an} trains the shared and private representations in two stages, for as many epochs as \method{} and $60\%$ of them, respectively, while JointOpt~\citep{wang2025an} optimizes both jointly. On VGGSounder, we fix the best-performing $\lambda = 0.1$ for both and vary $\beta$ and $a$, respectively. For RePercENT~\citep{rizou2026repercent}, we sweep $(\alpha, \lambda)$ with $\alpha \geq \lambda$, as required by its optimality guarantee. CLIP with fine-tuned projection heads trains the projections and temperature with symmetric InfoNCE for 100 epochs (learning rate $5\times10^{-5}$, weight decay $0.1$, one warmup epoch).

\begin{figure*}[t]
    \centering
    \includegraphics[width=0.75\textwidth]{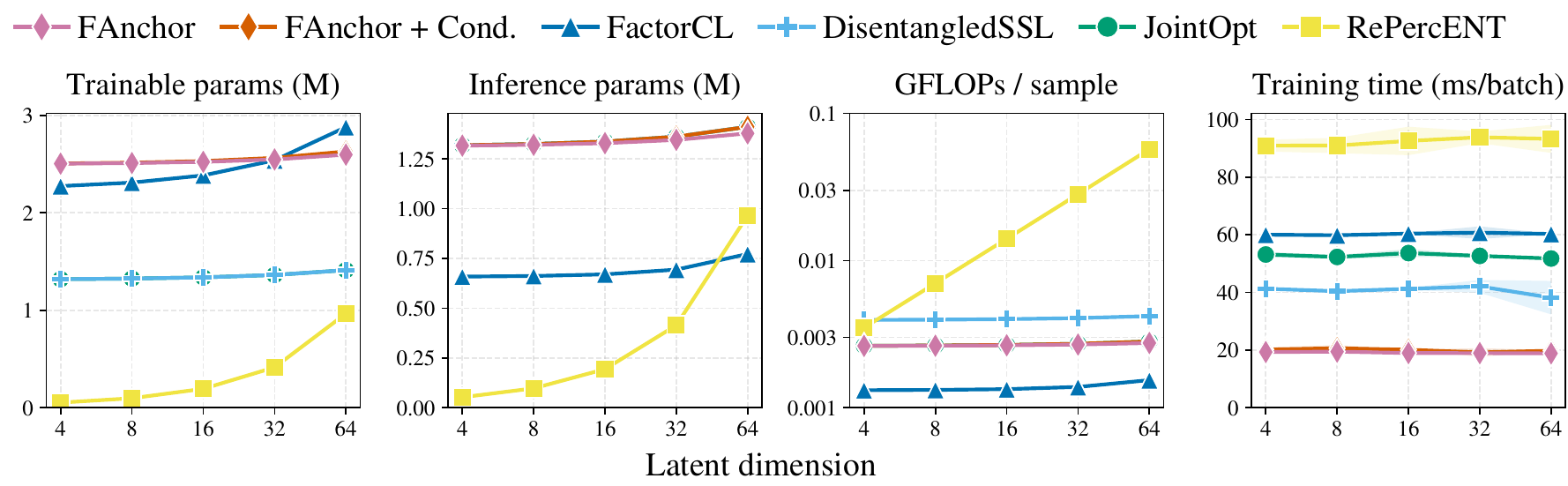}
    \caption{\textbf{Computational efficiency on Clevr-4.} Trainable and inference parameters, GFLOPs, and training time across latent dimensions.}
    \label{fig:clevr4_compute_cost}
    \vspace{-0.16in}
\end{figure*}

\vspace{0.08in} \noindent \textbf{VAST-150K.} Following the pipeline of TRIANGLE~\citep{cicchetti2026triangle}, we fine-tune the pretrained VAST encoders~\citep{chen2023vast}, without their fusion layers, on VAST-150K with \loss{} for 5 epochs, using $\lambda = 16$, a batch size of 256, a learning rate of $2\times10^{-5}$, and a warmup ratio of $0.215$. GRAM~\citep{cicchetti2025gramian} and TRIANGLE~\citep{cicchetti2026triangle} are fine-tuned in the same way with their original objectives. All methods use 2 frames per clip for training and 8 for evaluation (16 on VATEX~\citep{wang2019vatex}).

\subsection{Computational cost}

\begin{table}[t]
    \centering
    \vspace{-0.05in}
    \caption{\textbf{Computational efficiency on VGGSounder.}
    Trainable and inference parameters, GFLOPs, and training time (mean $\pm$ std).}
    \label{tab:cost_vggsound}
    \small
    \begin{tabular}{rcccc}
        \toprule
        \textbf{Method} & \textbf{Trainable params (M)} & \textbf{Inference params (M)} & \textbf{GFLOPs/sample} & \textbf{Training time (ms/batch)} \\
        \midrule
        \rowcolor{ourpurple}
        \methodb{} (ours) & 7.35 & 3.68 & 0.0073 & $19.5 \pm 0.4$ \\
        \rowcolor{ourpurple}
        \methodb{} \textbf{+ Cond.} (ours) & 7.62 & 3.94 & 0.0079 & $20.1 \pm 0.1$ \\
        \midrule
        FactorCL~\citep{liang2023factorized} & 6.83 & 3.15 & 0.0063 & $524.8 \pm 0.1$ \\
        DisentangledSSL~\citep{wang2025an} & 3.94 & 3.94 & 0.0115 & 29.5 $\pm$ 3.9 \\
        JointOpt~\citep{wang2025an} & 3.94 & 3.94 & 0.0079 & $38.1 \pm 1.8$ \\
        RePercENT~\citep{rizou2026repercent} & 6.85 & 6.85 & 0.1849 & $108.0 \pm 1.6$ \\
        \bottomrule
    \end{tabular}
\end{table}

\begin{table}[t]
    \centering
    \vspace{-0.05in}
    \caption{\textbf{Computational efficiency on RoCOCO.}
    Trainable and inference parameters, GFLOPs, and training time (mean $\pm$ std).}
    \label{tab:cost_rococo}
    \small
    \begin{tabular}{rcccc}
        \toprule
        \textbf{Method} & \textbf{Trainable params (M)} & \textbf{Inference params (M)} & \textbf{GFLOPs/sample} & \textbf{Training time (ms/batch)} \\
        \midrule
        \rowcolor{ourpurple}
        \methodb{} (ours) & 5.52 & 2.89 & 0.0058 & $20.6 \pm 1.0$ \\
        \rowcolor{ourpurple}
        \methodb{} \textbf{+ Cond.} (ours) & 5.78 & 3.15 & 0.0063 & $20.6 \pm 0.5$\\
        \midrule
        FactorCL~\citep{liang2023factorized} & 6.57 & 2.89 & 0.0058 & $2140.2 \pm 1.5$ \\
        DisentangledSSL~\citep{wang2025an} & 3.15 & 3.15 & 0.0094 & 35.6 $\pm$ 2.3 \\
        JointOpt~\citep{wang2025an} & 3.15 & 3.15 & 0.0063 & $49.8 \pm 1.6$ \\
        RePercENT~\citep{rizou2026repercent} & 6.06 & 6.06 & 0.1541 & $146.4 \pm 1.4$ \\
        \bottomrule
    \end{tabular}
    \vspace{-0.1in}
\end{table}

\noindent \textbf{Compute resources.} Experiments run on NVIDIA A100, H100, or H200 GPUs, depending on availability, with one GPU per run except on VAST-150K, which uses 4 GPUs per run. All reported timings are measured on A100 80GB GPUs. In total, all experiments, including baselines and ablations, required approximately 2{,}300 GPU hours, more than half of which on VAST-150K. Linear probing used about 160 CPU hours on Clevr-4, while on VGGSounder it ran on the GPU.

\begin{wraptable}[11]{r}{0.48\linewidth}
    \centering
    \vspace{-0.1in}
    \setlength{\tabcolsep}{2.5pt}
    \caption{\textbf{Computational efficiency on VAST-150K.} Loss time covers the forward and backward pass of the objective; training time covers a full training step, including the encoders. Times are reported as mean $\pm$ std.}
    \label{tab:cost_vast}
    \small
    \begin{tabular}{rcc}
        \toprule
        \textbf{Method}
            & \textbf{Loss time (ms)}
            & \textbf{Training time (ms/batch)} \\
        \midrule
        \rowcolor{ourpurple}
        \lossb{} (ours)
            & 1.36 $\pm$ 0.26
            & 3724 $\pm$ 39 \\
        \midrule
        GRAM~\citep{cicchetti2025gramian}
            & 3.07 $\pm$ 0.26
            & 3763 $\pm$ 33 \\
        TRIANGLE~\citep{cicchetti2026triangle}
            & 4.97 $\pm$ 0.20
            & 3756 $\pm$ 36 \\
        \bottomrule
    \end{tabular}
\end{wraptable}

\vspace{0.08in} \noindent \textbf{Computational efficiency.} We report trainable and inference parameters, GFLOPs per sample, and training time per batch, excluding the frozen encoders, on Clevr-4 in~\Cref{fig:clevr4_compute_cost}, VGGSounder in~\Cref{tab:cost_vggsound}, and RoCOCO in~\Cref{tab:cost_rococo}.

In terms of parameters, \method{} has more trainable parameters than DisentangledSSL and JointOpt due to its decoders, which are used only during training. At inference, however, it requires among the fewest parameters, second only to FactorCL on VGGSounder and on par with it on RoCOCO, while RePercENT requires the most. On Clevr-4, FactorCL and RePercENT require fewer inference parameters, but the latter grows with the latent dimension, whereas \method{} remains nearly constant.

In terms of GFLOPs, \method{} is among the cheapest methods, second only to FactorCL on Clevr-4 and VGGSounder and on par with it on RoCOCO, while RePercENT requires an order of magnitude more, as it applies cross-attention over full token sequences rather than pooled embeddings. Note, however, that FactorCL retrieves with an MLP critic applied to every query--gallery pair rather than a dot product, a cost not reflected in GFLOPs per sample.

In terms of training time, \method{} is the fastest across all datasets, and hierarchical conditioning adds negligible cost. FactorCL is the slowest, as it trains several critic networks that score every pair of samples in a batch.

\begin{takeaway}
Overall, \method{} has the shortest training time and remains among the cheapest in GFLOPs across all datasets.
\end{takeaway}

\vspace{0.08in} 
\noindent 
\textbf{Efficiency of \loss{}.} On VAST-150K in~\Cref{tab:cost_vast}, \loss{} computes the loss $2.3$ and $3.7$ times faster than GRAM and TRIANGLE, respectively, consistent with its linear scaling in the number of modalities discussed in~\Cref{sec:exp_multi_align}. However, since the full training step is dominated by the encoders, the overall training time remains comparable across objectives.

\section{Additional results}
\label{app:exp_details}

\begin{figure*}[t]
    \centering

    \begin{subfigure}{0.92\textwidth}
        \centering
        \includegraphics[width=\linewidth]{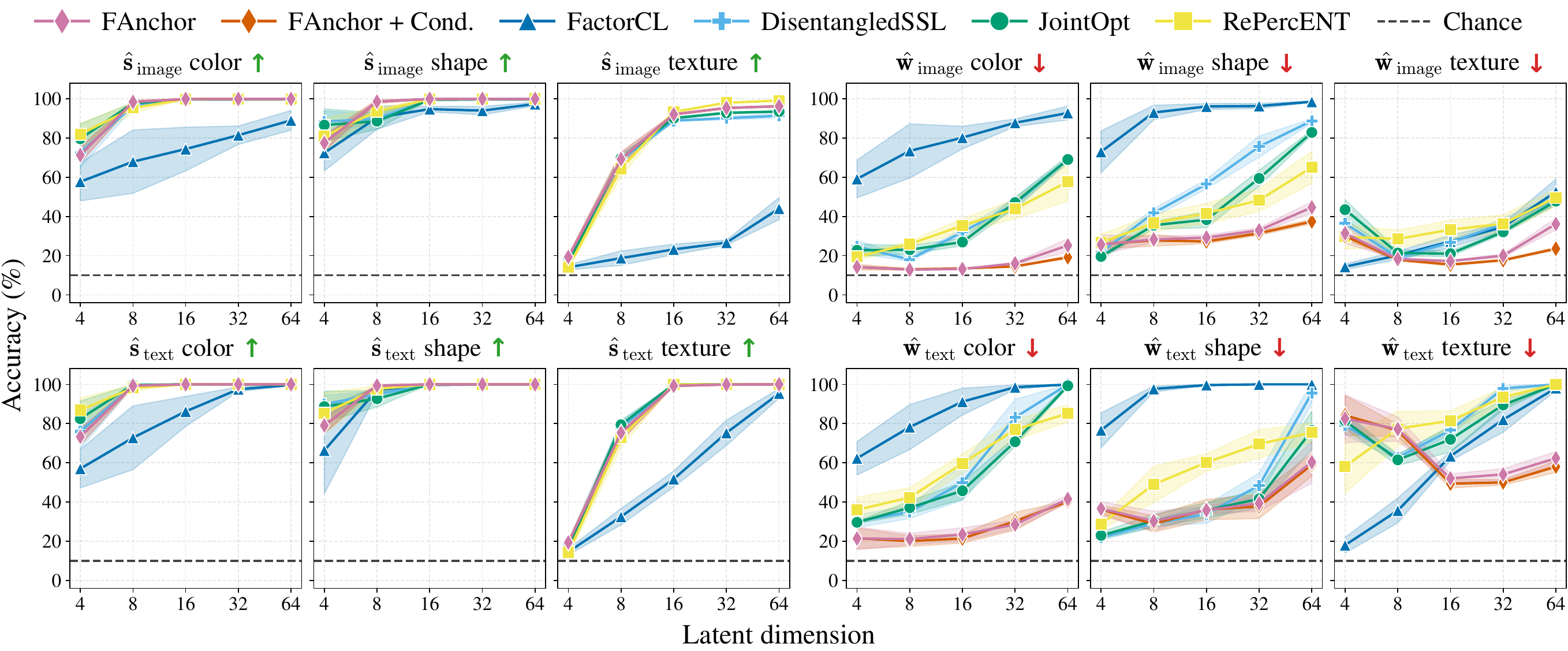}
        \caption{\textbf{(Image, Text) setting}}
        \label{fig:clevr4_image_text}
    \end{subfigure}

    \begin{subfigure}{0.92\textwidth}
        \centering
        \includegraphics[width=\linewidth]{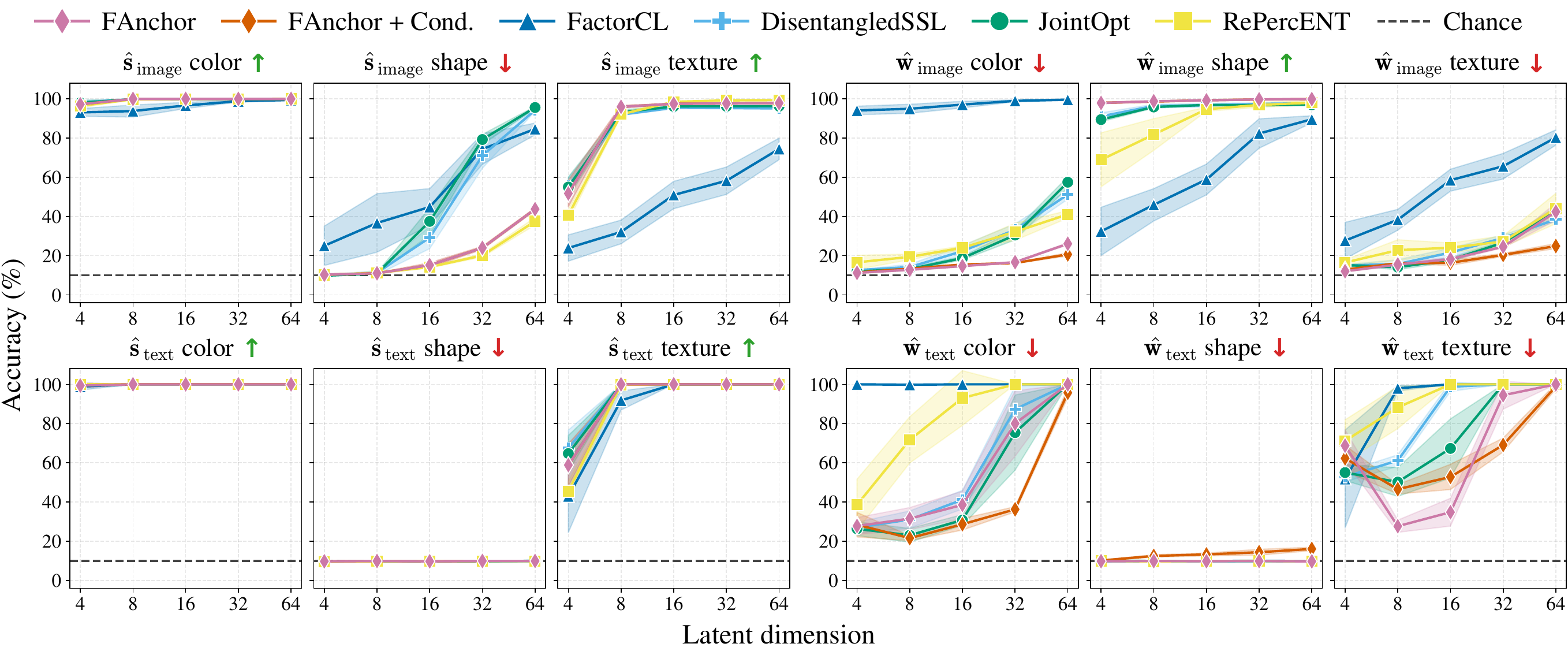}
        \caption{\textbf{(Image, Text without shape) setting}}
        \label{fig:clevr4_image_text_no_shape}
    \end{subfigure}

    \begin{subfigure}{0.92\textwidth}
        \centering
        \includegraphics[width=\linewidth]{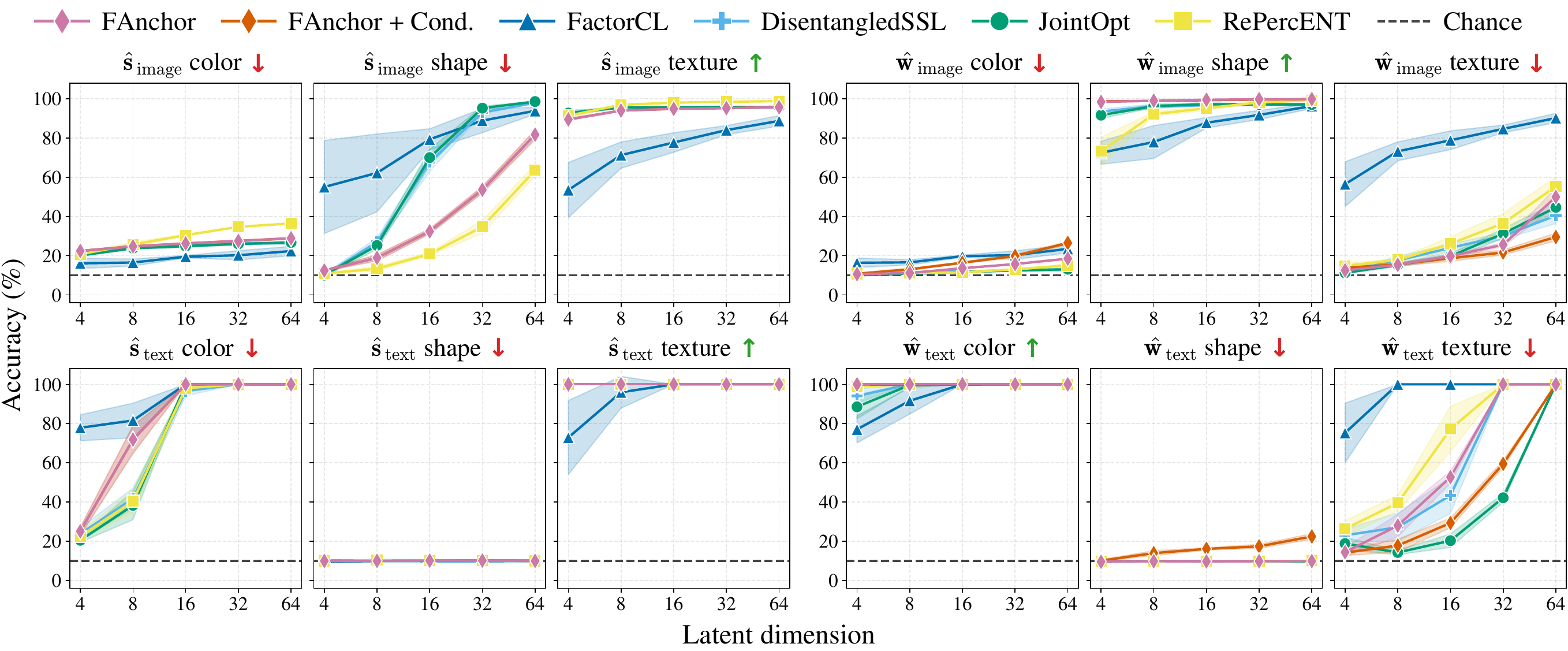}
        \caption{\textbf{(Image without color, Text without shape) setting}}
        \label{fig:clevr4_image_no_color_text_no_shape}
    \end{subfigure}

    \caption{\textbf{Factorization on Clevr-4.} Linear-probe accuracy (\%) for predicting color, shape, and texture from the shared (left) and private (right) representations across latent dimensions. \up/\down: higher/lower is better; dashed lines: chance (10\%).}
    \label{fig:clevr4}
\end{figure*}

\subsection{Clevr-4}

\noindent \textbf{Factorization in the remaining settings.} \Cref{fig:clevr4} reports the remaining three Clevr-4 settings. Overall, the trends observed in~\Cref{sec:clevr4} still hold, including the increase in leakage with latent dimension for all methods, likely due to excess capacity. We therefore focus on observations specific to each setting.

\vspace{0.08in} 
\textbf{(Image, Text).} \Cref{fig:clevr4_image_text} shows this setting, where all three attributes are shared. Our shared representations preserve color and shape from dimension $8$ and texture from $16$ onward, matching the strongest baselines, whereas FactorCL lags behind, particularly on texture. Our private representations exhibit substantially lower leakage of shared information than the baselines, and this difference grows with latent dimension. The texture leakage into $\hat{\mathbf{w}}_{\text{text}}$ at low dimensions is consistent with our Matryoshka-like reconstruction objective: since $\hat{\mathbf{s}}_{\text{text}}$, and hence the anchor, capture texture only partially, $\hat{\mathbf{w}}_{\text{text}}$ is encouraged to retain it. As the shared representations become more informative, this leakage decreases. Finally, \method{} + Cond., which leaves the shared representations unchanged, further reduces private leakage, mainly in $\hat{\mathbf{w}}_{\text{image}}$.

\vspace{0.08in} 
\textbf{(Image, Text without shape).} \Cref{fig:clevr4_image_text_no_shape} shows this setting, where shape is private to images, while color and texture are shared. In terms of information preservation, our shared representations capture the shared attributes from dimension $8$ onward, and our private image representations capture shape at all dimensions, on par with the strongest baselines. In terms of leakage, our shared representations exhibit low private-to-shared leakage of shape into $\hat{\mathbf{s}}_{\text{image}}$, comparable to RePercENT, whereas the other baselines leak substantially more from dimension $16$ onward. Our private representations also achieve the lowest shared-to-private leakage, in $\hat{\mathbf{w}}_{\text{text}}$ at most dimensions and, for color, in $\hat{\mathbf{w}}_{\text{image}}$ at all dimensions.

\vspace{0.08in} 
\textbf{(Image without color, Text without shape).} \Cref{fig:clevr4_image_no_color_text_no_shape} shows this setting, where only texture is shared, while shape is private to images and color to text. In terms of information preservation, our shared and private representations already capture their respective attributes from dimension $4$, whereas baselines require more capacity, particularly FactorCL. In terms of leakage, our shared representations exhibit low private-to-shared leakage of shape into $\hat{\mathbf{s}}_{\text{image}}$, second only to RePercENT. However, like all methods, they encode the text-private color in $\hat{\mathbf{s}}_{\text{text}}$ from dimension $16$ onward. For shared-to-private leakage, \method{} + Cond. leaks the least texture into $\hat{\mathbf{w}}_{\text{image}}$ and the second least into $\hat{\mathbf{w}}_{\text{text}}$, after JointOpt.

\begin{figure*}[t]
    \centering
    \begin{subfigure}[t]{\textwidth}
        \centering
        \includegraphics[width=0.935\linewidth]{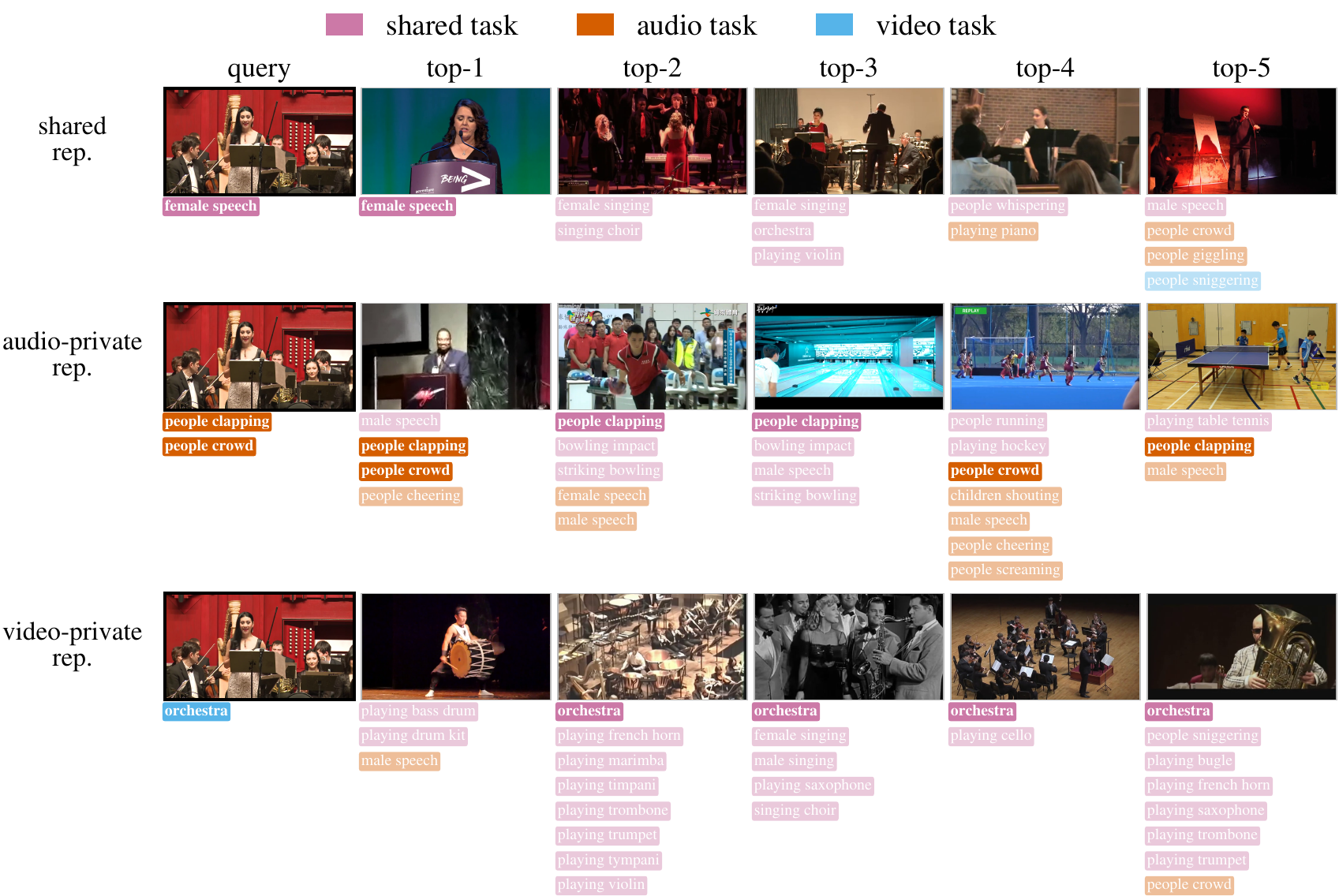}
        \caption{}
        \label{fig:vggsounder_qualitative_results_1}
    \end{subfigure}

    \begin{subfigure}[t]{\textwidth}
        \centering
        \includegraphics[width=0.935\linewidth]{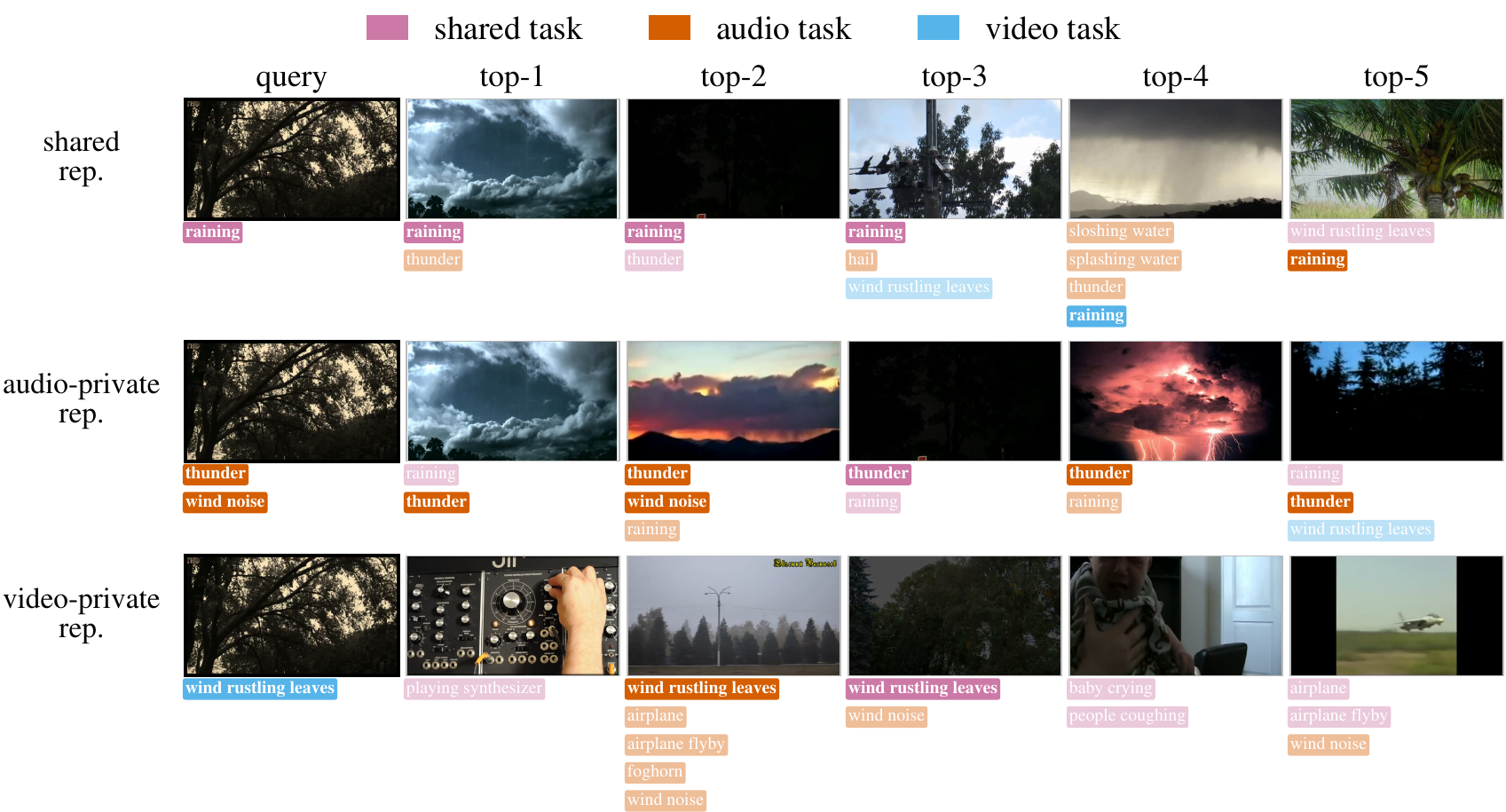}
        \caption{}
        \label{fig:vggsounder_qualitative_results_2}
    \end{subfigure}

    \caption{\textbf{Factorization on VGGSounder: Qualitative examples.}
    Top-5 nearest neighbors of a query clip in the shared (top), audio-private (middle), and video-private (bottom) representations of \method{}. Labels are colored by task (shared, audio, or video). In each row, the query shows only the labels of the corresponding task, and matching neighbor labels are highlighted, even if annotated with a different task in the neighbor. Ideally, each representation retrieves clips sharing these labels.}
    \label{fig:vggsounder_qualitative_results}
\end{figure*}

\subsection{VGGSounder}

\noindent \textbf{FactorCL~\citep{liang2023factorized}.} We omit FactorCL from~\Cref{fig:av_single_f1}, as it achieves substantially lower information preservation than all other methods on VGGSounder: its audio-private, video-private, and shared representations reach only 11.4\%, 18.4\%, and 38.7\% F1@1 on their respective tasks, compared with at least 14.9\%, 23.1\%, and 48.0\% for the others under the same protocol.  

\vspace{0.08in} 
\noindent \textbf{Qualitative examples.} \Cref{fig:vggsounder_qualitative_results} shows the top-5 nearest neighbors of two query clips in the shared, audio-private, and video-private representations of \method{}, retrieved by cosine similarity on the VGGSounder test set. In both cases, the three representations retrieve largely different neighbors, each capturing a different aspect of the query.

In \Cref{fig:vggsounder_qualitative_results_1}, a woman sings in front of an orchestra while the audience applauds. The shared representation retrieves stage performances or talks, four of which feature a woman, capturing what both modalities agree on. All audio-private neighbors share clapping or crowd sounds with the query despite visually unrelated scenes, such as bowling, hockey, and table tennis, indicating that this representation captures audible but not visible information. Similarly, four video-private neighbors show orchestral performances, one of which is also labeled ``people crowd'', as crowds often accompany such performances. 

In \Cref{fig:vggsounder_qualitative_results_2}, the storm query has the shared label ``raining'', the audio-only labels ``thunder'' and ``wind noise'', and the video-only label ``wind rustling leaves''. This example illustrates the \emph{entanglement} discussed in \Cref{sec:vggsounder}: besides ``raining'', the shared representation retrieves ``thunder'' in three neighbors, whereas besides ``thunder'', the audio-private representation retrieves ``raining'' in all five. Since rain and thunder often co-occur, they cannot be fully separated. In contrast, the video-private representation retrieves ``wind rustling leaves'' in two neighbors and ``raining'' in none, indicating a clearer separation from the shared information than the audio-private representation.

\begin{table*}[t]
\centering
\caption{\textbf{Robustness of shared representations.} Text-to-image retrieval on clean MS-COCO 5K and RoCOCO with adversarial images added to the retrieval gallery. \up/\down: higher/lower is better. Best in \textbf{bold}, second-best \underline{underlined}.}
\label{tab:rococo_t2i}
\footnotesize
\setlength{\tabcolsep}{2.35pt}
\begin{tabular}{r c ccc ccc ccc ccc}
\toprule
& \textbf{COCO 5K}
& \multicolumn{3}{c}{\textbf{Mix} ($\rho_\text{mix}=0.9$)}
& \multicolumn{3}{c}{\textbf{Mix} ($\rho_{\text{mix}}=0.8$)}
& \multicolumn{3}{c}{\textbf{Patch} ($\rho_{\text{patch}}=0.9$)}
& \multicolumn{3}{c}{\textbf{Patch} ($\rho_{\text{patch}}=0.8$)} \\
\cmidrule(lr){2-2}
\cmidrule(lr){3-5}
\cmidrule(lr){6-8}
\cmidrule(lr){9-11}
\cmidrule(lr){12-14}
\textbf{Text to Image}
& \textbf{R@1}\up
& \textbf{R@1}\up & \textbf{drop}\down & \textbf{RSMS}\down
& \textbf{R@1}\up & \textbf{drop}\down & \textbf{RSMS}\down
& \textbf{R@1}\up & \textbf{drop}\down & \textbf{RSMS}\down
& \textbf{R@1}\up & \textbf{drop}\down & \textbf{RSMS}\down \\
\midrule
\rowcolor{ourpurple}
\methodb{} (ours)
& \underline{37.78}
& \textbf{27.92} & \textbf{26.10} & \textbf{29.14}
& \textbf{31.67} & \textbf{16.18} & \textbf{19.34}
& \textbf{32.16} & \textbf{14.87} & \textbf{17.76}
& \textbf{33.88} & \textbf{10.34} & \textbf{13.07} \\
\midrule
CLIP {\scriptsize(zero-shot)}~\citep{radford2021learning}
& 30.44
& 20.74 & 31.87 & \underline{34.41}
& 23.40 & 23.13 & 26.21
& 23.10 & 24.11 & 28.18
& 24.77 & 18.63 & 23.09 \\

CLIP {\scriptsize(proj. FT)}~\citep{radford2021learning}
& \textbf{42.23}
& \underline{27.08} & 35.87 & 41.08
& \underline{30.43} & 27.94 & 34.43
& \underline{30.08} & 28.76 & 34.87
& \underline{32.32} & 23.46 & 29.78 \\
\midrule

FactorCL~\citep{liang2023factorized}
& 30.67
& 19.00 & 38.05 & 43.94
& 21.20 & 30.89 & 37.79
& 21.04 & 31.41 & 38.59
& 22.69 & 26.04 & 33.05 \\

DisentangledSSL~\citep{wang2025an}
& 36.28
& 25.31 & 30.22 & 34.60
& 29.03 & \underline{19.96} & \underline{24.37}
& 29.72 & 18.07 & \underline{21.85}
& 31.78 & \underline{12.40} & \underline{15.73} \\

JointOpt~\citep{wang2025an}
& 36.60
& 25.76 & \underline{29.61} & 34.50
& 29.29 & 19.99 & 24.78
& 30.00 & \underline{18.04} & 22.54
& 32.04 & 12.47 & 16.32 \\

RePercENT~\citep{rizou2026repercent}
& 26.27
& 16.36 & 37.70 & 43.64
& 18.59 & 29.24 & 36.42
& 18.28 & 30.40 & 37.46
& 19.80 & 24.61 & 31.93 \\
\bottomrule
\end{tabular}
\end{table*}

\begin{figure*}[t]
    \centering
    \includegraphics[width=0.935\textwidth]{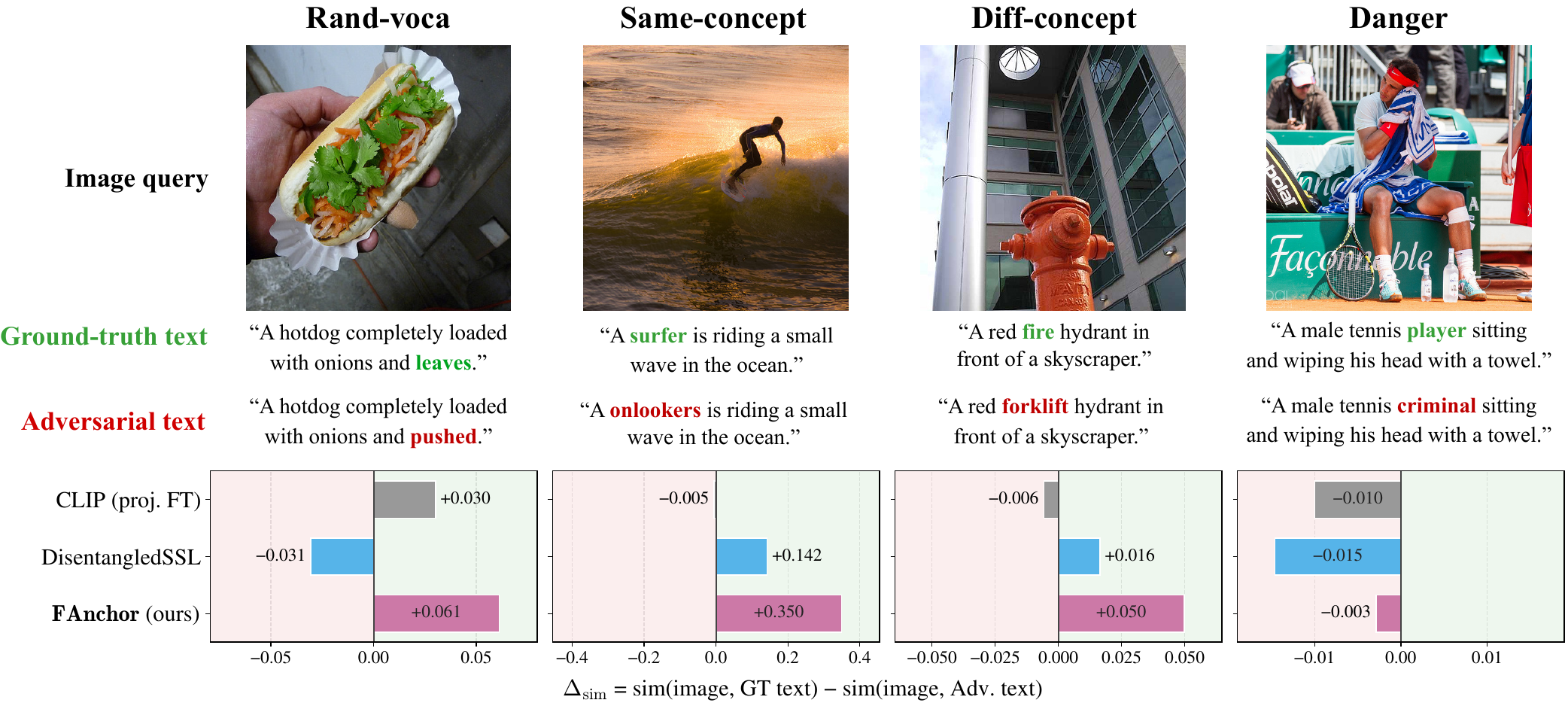}
    \caption{\textbf{Robustness of shared representations.} Examples of image queries on RoCOCO with ground-truth and adversarial captions across four distractor types. \textbf{\textcolor{upgreen}{Green}} and \textbf{\textcolor{downred}{red}} highlight the \textbf{\textcolor{upgreen}{original}} and \textbf{\textcolor{downred}{replaced}} words, respectively. Bars show the similarity difference $\Delta_{\mathrm{sim}}$ between the ground-truth and adversarial captions, \emph{where positive values indicate a preference for the ground truth}.}
    \label{fig:rococo_exp}
\end{figure*} 

\subsection{RoCOCO}

\noindent \textbf{Robustness in text-to-image retrieval.} \Cref{tab:rococo_t2i} reports text-to-image retrieval results with adversarial images added to the retrieval gallery. The trends observed for image-to-text retrieval in~\Cref{sec:exp_robust} still hold. On clean MS-COCO, \method{} ranks second in R@1 ($37.78\%$), behind only CLIP with fine-tuned projection heads ($42.23\%$), and outperforms all factorization baselines. On RoCOCO, it achieves the highest R@1 and the lowest relative drop and RSMS in every setting. Our relative drop ranges from $10.34\%$ to $26.10\%$ and our RSMS from $13.07\%$ to $29.14\%$, whereas the RSMS of fine-tuned CLIP is $11.94$ to $17.11$ points higher, and that of DisentangledSSL and JointOpt $2.66$ to $5.46$ points higher. Interestingly, although zero-shot CLIP achieves a lower R@1 than CLIP with fine-tuned projection heads, it is more robust, with a lower relative drop and RSMS in all four settings (\eg, an RSMS of $34.41\%$ vs.\ $41.08\%$ for $\rho_{\text{mix}} = 0.9$).

\vspace{0.08in} \noindent \textbf{Qualitative examples.} \Cref{fig:rococo_exp} shows randomly selected image queries from RoCOCO with their ground-truth and adversarial captions for each distractor type, together with the similarity difference $\Delta_{\text{sim}} = \mathrm{sim}(\text{image}, \text{GT text}) - \mathrm{sim}(\text{image}, \text{Adv.\ text})$ for CLIP with fine-tuned projection heads, DisentangledSSL, and \method{}. Positive values indicate a preference for the ground-truth caption. \method{} yields positive values in all cases except Danger, where all methods are slightly negative, although \method{} remains closest to neutral. These examples illustrate the quantitative trends in~\Cref{tab:rococo_i2t}.

\begin{table*}[t]
\centering
\caption{{\boldmath\textbf{Alignment beyond two modalities.}} Zero-shot text-to-video (T2V) and video-to-text (V2T) retrieval (R@1, \%) on MSR-VTT, ActivityNet, and VATEX, with videos given as frames and audio. 
VAST denotes the pretrained model without fine-tuning. 
Results are reported as \textbf{mean $\pm$ standard deviation} over 5 seeds.
\up: higher is better. Best in \textbf{bold}, second-best \underline{underlined}.}
\label{tab:vast_std}
\footnotesize
\setlength{\tabcolsep}{5pt}
\begin{tabular}{rcccccc}
\toprule
&
\multicolumn{2}{c}{\textbf{MSR-VTT}}
& \multicolumn{2}{c}{\textbf{ActivityNet}}
& \multicolumn{2}{c}{\textbf{VATEX}} \\
\cmidrule(lr){2-3}
\cmidrule(lr){4-5}
\cmidrule(lr){6-7}
\textbf{Method}
& \textbf{T2V}\up & \textbf{V2T}\up
& \textbf{T2V}\up & \textbf{V2T}\up
& \textbf{T2V}\up & \textbf{V2T}\up \\
\midrule
VAST~\citep{chen2023vast}
& 49.20 & 46.90
& 51.80 & 47.80
& 75.70 & 73.00 \\
\midrule
\rowcolor{ourpurple}
\lossb \, (ours)
& \textbf{52.86 $\pm$ 0.18} & \textbf{51.34 $\pm$ 0.37}
& \underline{56.48 $\pm$ 0.24} & 50.92 $\pm$ 0.26
& \textbf{79.64 $\pm$ 0.19} & \underline{74.74 $\pm$ 0.19}\\
\midrule
GRAM~\citep{cicchetti2025gramian}
& \underline{52.58 $\pm$ 0.46} & 50.14 $\pm$ 0.57
& \textbf{59.28 $\pm$ 0.33} & \underline{52.72 $\pm$ 0.43}
& \underline{79.28 $\pm$ 0.45} & 74.42 $\pm$ 0.44 \\

TRIANGLE~\citep{cicchetti2026triangle}
& 49.86 $\pm$ 0.15 & \underline{50.86 $\pm$ 0.36}
& 56.06 $\pm$ 0.30 & \textbf{53.44 $\pm$ 0.48}
& 77.56 $\pm$ 0.31 & \textbf{74.78 $\pm$ 0.42} \\
\bottomrule
\end{tabular}
\vspace{-0.1in}
\end{table*}

\subsection{VAST-150K}

\noindent \textbf{Standard deviations and significance tests.} \Cref{tab:vast_std} reports the mean and standard deviation of R@1 over $5$ different seeds for \loss{}, GRAM, and TRIANGLE. We compare \loss{} with each baseline using an uncorrected two-sided Student's $t$-test per dataset and retrieval direction ($p < 0.05$), with the conclusions summarized in~\Cref{sec:exp_multi_align}.

\subsection{Ablations}

\begin{figure*}[t]
    \centering

    \begin{subfigure}{\textwidth}
        \centering
        \includegraphics[width=0.87\linewidth]{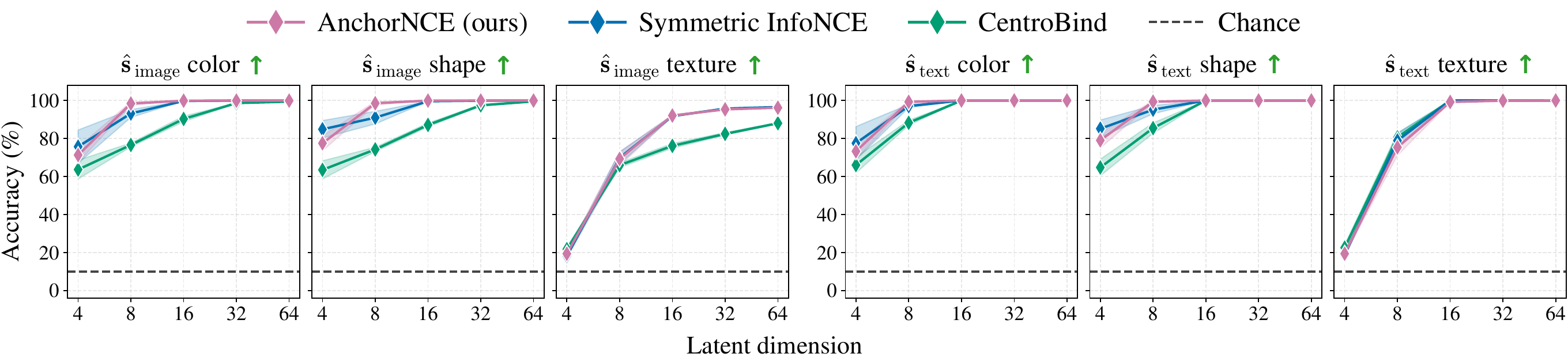}
        \caption{\textbf{(Image, Text) setting}}
        \label{fig:ablation_cl_loss_lam1_shared_image_text}
    \end{subfigure}

        \begin{subfigure}{0.87\textwidth}
        \centering
        \includegraphics[width=\linewidth]{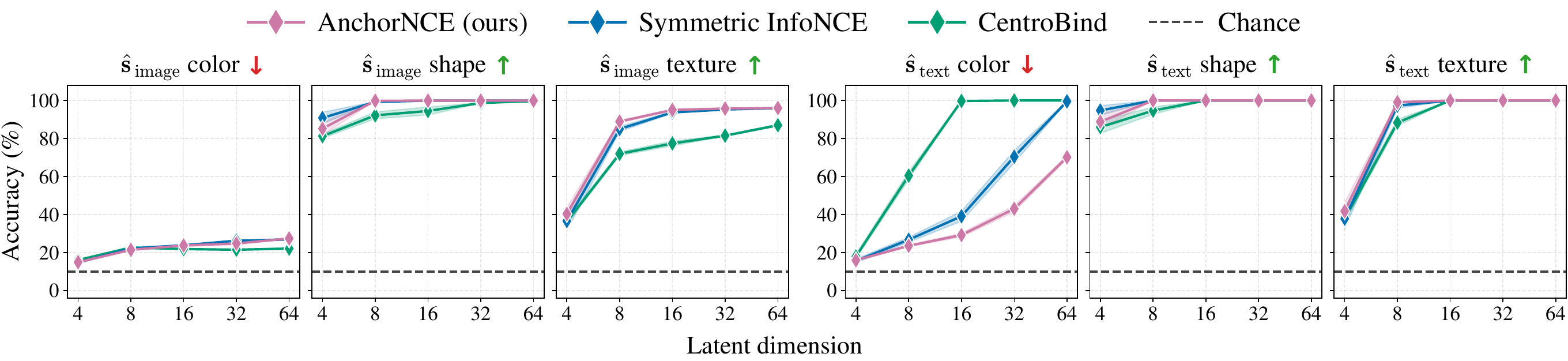}
        \caption{\textbf{(Image without color, Text) setting}}
        \label{fig:ablation_cl_loss_lam1_shared_image_no_color_text}
    \end{subfigure}

    \begin{subfigure}{0.87\textwidth}
        \centering
        \includegraphics[width=\linewidth]{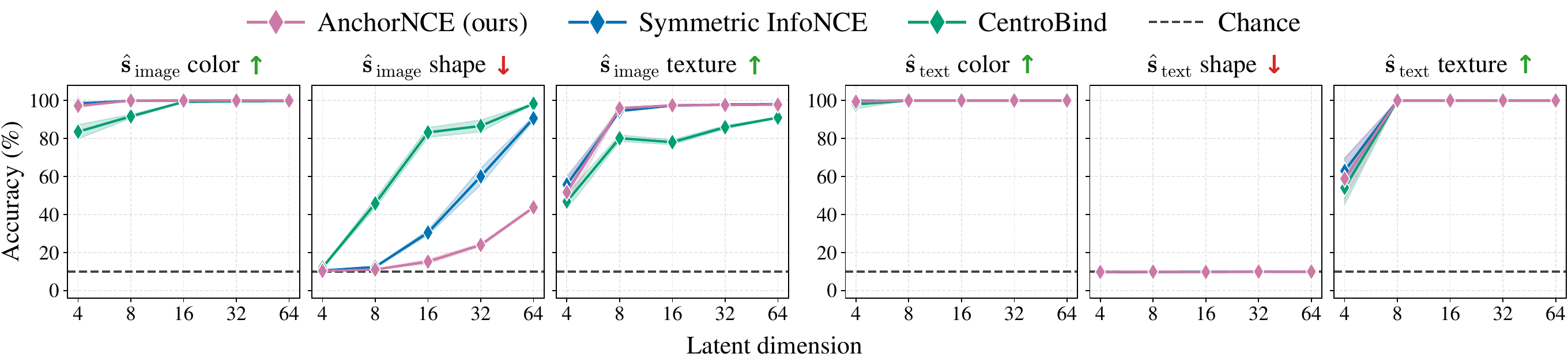}
        \caption{\textbf{(Image, Text without shape) setting}}
         \label{fig:ablation_cl_loss_lam1_shared_image_text_no_shape}
    \end{subfigure}
    
    \caption{\textbf{Factorization on Clevr-4: Contrastive objective ablation.} Linear-probe accuracy (\%) for predicting color, shape, and texture \emph{from shared representations} across four settings and latent dimensions. \up/\down: higher/lower is better; dashed lines: chance (10\%).}
    \label{fig:ablation_cl_loss_lam1_shared_all_settings}
\end{figure*}

\begin{figure*}[t]
    \ContinuedFloat
    \centering

    \begin{subfigure}{0.87\textwidth}
        \centering
        \includegraphics[width=\linewidth]{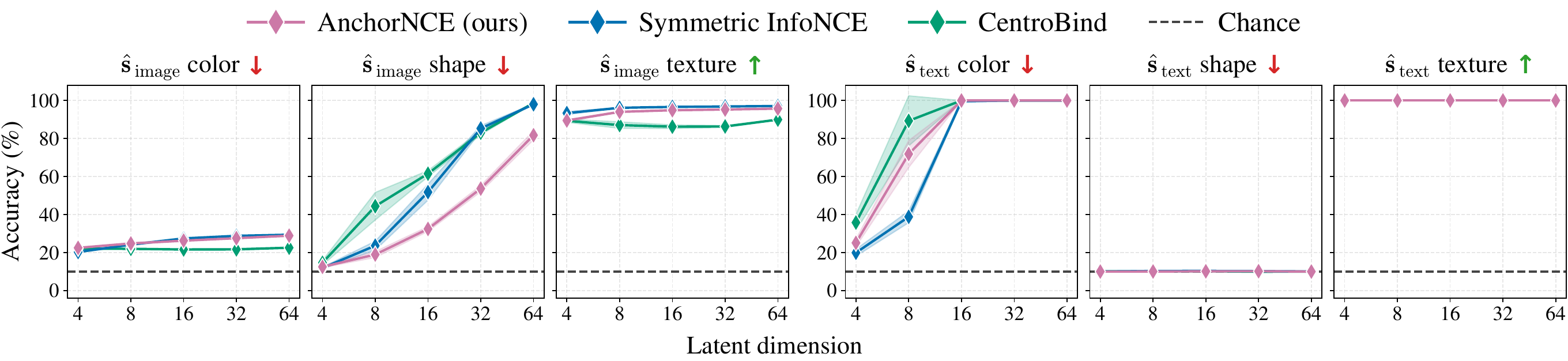}
        \caption{\textbf{(Image without color, Text without shape) setting}}
        \label{fig:ablation_cl_loss_lam1_shared_image_no_color_text_no_shape}
    \end{subfigure}
    \caption[]{\textbf{Factorization on Clevr-4: Contrastive objective ablation (continued).}}
\end{figure*}

\begin{figure}[t]
    \centering
    \includegraphics[width=0.5\textwidth]{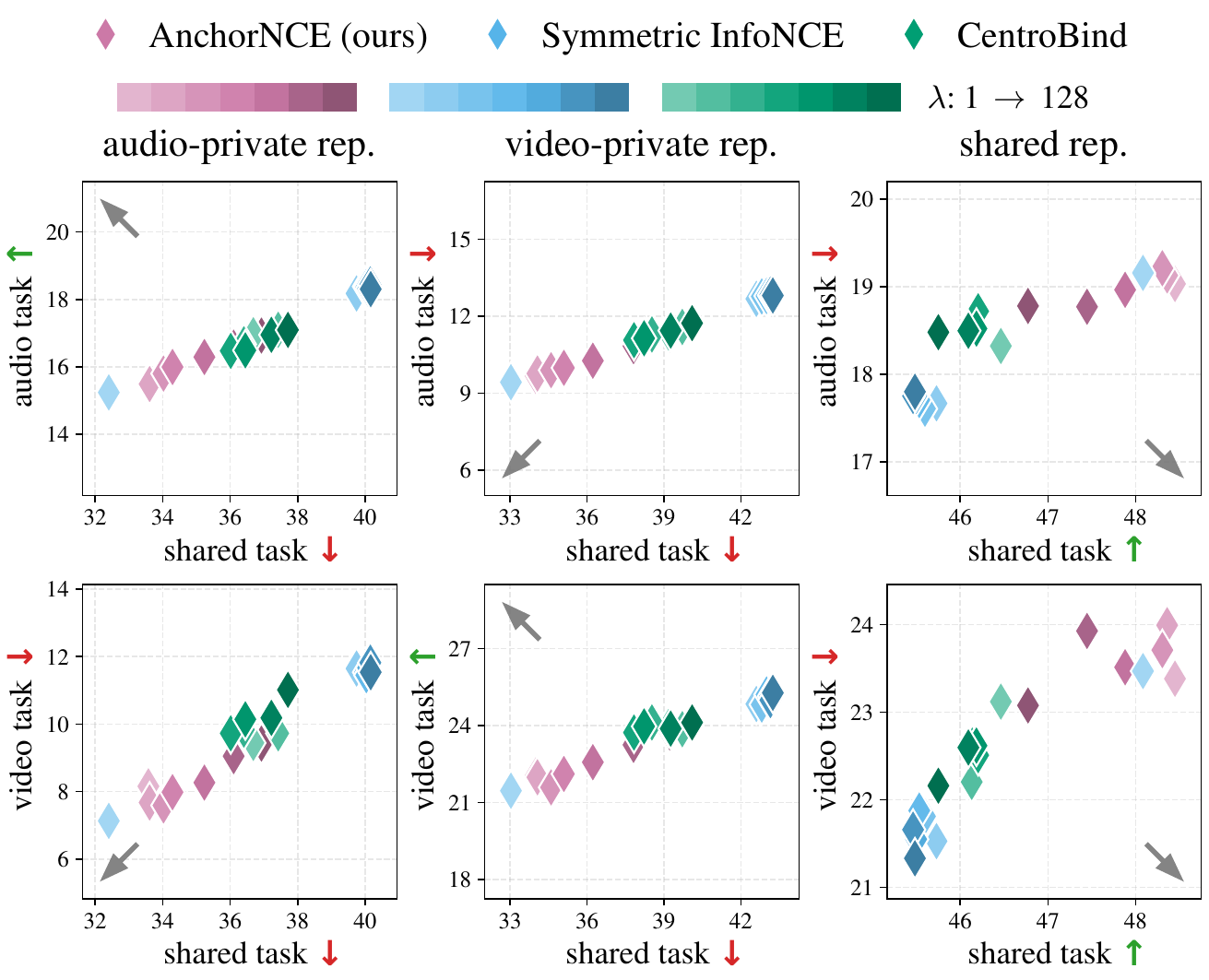}
    \caption{\textbf{Factorization on VGGSounder: Contrastive objective ablation.} Linear-probe F1@1 (\%) on shared and private tasks using audio-private (left), video-private (middle), and shared (right) representations. Each point corresponds to one hyperparameter configuration; color gradients indicate $\lambda$ values from $1$ to $128$ for all objectives. \up/\down: higher/lower is better; \optarrow: optimal regions.}
    \label{fig:vggsounder_cl_loss_f1_at1_disenFalse}
\end{figure}

\subsubsection{Contrastive objective} 

To isolate the effect of \loss{}, we compare it with two alternative contrastive objectives for learning the shared representations. (i)~Symmetric InfoNCE~\cite{infonce}, the standard pairwise objective used by CLIP, which directly contrasts the shared representations of the two modalities, as introduced in~\Cref{sec:background}. (ii)~CentroBind~\cite{jeong2024anchors}, which, like \loss{}, aligns each modality with the anchor $\mathbf{a}$, but applies a symmetric InfoNCE loss between each $\hat{\mathbf{s}}_m$ and $\mathbf{a}$:
\begin{equation}
\mathcal{L}_{\text{CentroBind}} = \frac{1}{M} \sum_{m=1}^{M} \mathcal{L}_{\text{SymInfoNCE}}(\hat{\mathbf{s}}_m, \mathbf{a}).
\end{equation}
In all cases, we keep the architecture and the private objective of \method{} unchanged and only replace $\mathcal{L}_{\text{shared}}$ in~\cref{eq:general_loss}. For a fair comparison, we apply the exponent $\lambda$ to both denominators, as in~\cref{eq:anchorloss}. Both perform best with $\lambda = 1$, \ie, their original formulation, which we use on Clevr-4 and RoCOCO. On VGGSounder, we report all values.

\vspace{0.08in} \noindent 
\textbf{Clevr-4.} \Cref{fig:ablation_cl_loss_lam1_shared_all_settings} reports linear-probe accuracy on the shared representations learned with each objective across the four Clevr-4 settings. In terms of information preservation, \loss{} and symmetric InfoNCE capture the shared attributes comparably, whereas CentroBind requires more capacity in general. In terms of leakage, \loss{} exhibits the lowest private-to-shared leakage in nearly all cases, except for color in $\hat{\mathbf{s}}_{\text{text}}$ in the (Image without color, Text without shape) setting of~\Cref{fig:ablation_cl_loss_lam1_shared_image_no_color_text_no_shape}, where all objectives fully encode color from dimension $16$ onward. In the (Image, Text without shape) setting of~\Cref{fig:ablation_cl_loss_lam1_shared_image_text_no_shape}, for example, shape is predicted from $\hat{\mathbf{s}}_{\text{image}}$ with only $43.7\%$ accuracy at dimension $64$ under \loss{}, compared with $90.6\%$ under symmetric InfoNCE and $98.3\%$ under CentroBind, which leaks the most overall.

\begin{table*}[t]
\centering
\caption{\textbf{Robustness of shared representations.} Image-to-text retrieval on clean MS-COCO 5K and RoCOCO with adversarial captions. All methods use \method{} with different contrastive objectives. \up/\down: higher/lower is better. Best in \textbf{bold}, second-best \underline{underlined}.}
\label{tab:rococo_i2t_loss}
\footnotesize
\setlength{\tabcolsep}{2.35pt}
\begin{tabular}{r c ccc ccc ccc ccc}
\toprule
& \textbf{COCO 5K}
& \multicolumn{3}{c}{\textbf{Rand-voca}}
& \multicolumn{3}{c}{\textbf{Same-concept}}
& \multicolumn{3}{c}{\textbf{Diff-concept}}
& \multicolumn{3}{c}{\textbf{Danger}} \\
\cmidrule(lr){2-2}
\cmidrule(lr){3-5}
\cmidrule(lr){6-8}
\cmidrule(lr){9-11}
\cmidrule(lr){12-14}
\textbf{Image to Text}
& \textbf{R@1}\up
& \textbf{R@1}\up & \textbf{drop}\down & \textbf{RSMS}\down
& \textbf{R@1}\up & \textbf{drop}\down & \textbf{RSMS}\down
& \textbf{R@1}\up & \textbf{drop}\down & \textbf{RSMS}\down
& \textbf{R@1}\up & \textbf{drop}\down & \textbf{RSMS}\down \\
\midrule
\rowcolor{ourpurple}
\lossb{} (ours)
& \underline{50.71}
& \textbf{46.32} & \textbf{8.66} & \textbf{9.47}
& \textbf{45.90} & \textbf{9.48} & \textbf{11.08}
& \textbf{47.60} & \textbf{6.13} & \textbf{7.14}
& \textbf{47.52} & \textbf{6.28} & \textbf{6.77} \\
\midrule
Symmetric InfoNCE~\citep{infonce}
& \textbf{55.92}
& \underline{42.50} & 23.99 & 30.05
& \underline{44.22} & 20.93 & 26.44
& \underline{45.60} & 18.45 & 23.50
& \underline{45.26} & \underline{19.07} & 22.79 \\

CentroBind~\citep{jeong2024anchors}
& 43.70
& 34.32 & \underline{21.48} & \underline{25.06}
& 34.86 & \underline{20.23} & \underline{23.66}
& 37.00 & \underline{15.34} & \underline{18.12}
& 35.14 & 19.60 & \underline{21.58} \\
\bottomrule
\end{tabular}
\end{table*}

\begin{table*}[t]
\centering

\caption{\textbf{Robustness of shared representations.} Text-to-image retrieval on clean MS-COCO 5K and RoCOCO with adversarial images. All methods use \method{} with different contrastive objectives. \up/\down: higher/lower is better. Best in \textbf{bold}, second-best \underline{underlined}.}

\label{tab:rococo_t2i_loss}
\footnotesize
\setlength{\tabcolsep}{2.35pt}
\begin{tabular}{r c ccc ccc ccc ccc}
\toprule
& \textbf{COCO 5K}
& \multicolumn{3}{c}{\textbf{Mix} ($\rho_{\text{mix}}=0.9$)}
& \multicolumn{3}{c}{\textbf{Mix} ($\rho_{\text{mix}}=0.8$)}
& \multicolumn{3}{c}{\textbf{Patch} ($\rho_{\text{patch}}=0.9$)}
& \multicolumn{3}{c}{\textbf{Patch} ($\rho_{\text{patch}}=0.8$)} \\
\cmidrule(lr){2-2}
\cmidrule(lr){3-5}
\cmidrule(lr){6-8}
\cmidrule(lr){9-11}
\cmidrule(lr){12-14}
\textbf{Text to Image}
& \textbf{R@1}\up
& \textbf{R@1}\up & \textbf{drop}\down & \textbf{RSMS}\down
& \textbf{R@1}\up & \textbf{drop}\down & \textbf{RSMS}\down
& \textbf{R@1}\up & \textbf{drop}\down & \textbf{RSMS}\down
& \textbf{R@1}\up & \textbf{drop}\down & \textbf{RSMS}\down \\
\midrule
\rowcolor{ourpurple}
\lossb{} (ours)
& \underline{37.78}
& \textbf{27.92} & \textbf{26.10} & \textbf{29.14}
& \textbf{31.67} & \textbf{16.18} & \textbf{19.34}
& \textbf{32.16} & \textbf{14.87} & \textbf{17.76}
& \textbf{33.88} & \textbf{10.34} & \textbf{13.07} \\
\midrule
Symmetric InfoNCE~\citep{infonce}
& \textbf{40.57}
& \underline{25.19} & 37.91 & 43.08
& \underline{28.35} & 30.12 & 36.78
& \underline{28.72} & 29.20 & 35.57
& \underline{31.36} & 22.70 & 29.02 \\
CentroBind~\citep{jeong2024anchors}
& 33.14
& 22.09 & \underline{33.35} & \underline{38.39}
& 24.77 & \underline{25.27} & \underline{31.22}
& 25.20 & \underline{23.97} & \underline{29.63}
& 27.16 & \underline{18.06} & \underline{23.76} \\
\bottomrule
\end{tabular}
\vspace{-0.1in}
\end{table*}

\vspace{0.08in} \noindent 
\textbf{VGGSounder.} \Cref{fig:vggsounder_cl_loss_f1_at1_disenFalse} reports the preservation--leakage trade-off of each objective for $\lambda \in \{1, \dots, 128\}$. For the private representations (left, middle), all objectives lie along the same diagonal as in~\Cref{sec:vggsounder}, and \loss{} yields the lowest shared-to-private leakage for most values of $\lambda$. For the shared representations (right), the objectives again lie along a diagonal, with \loss{} achieving the highest shared-task F1@1 for nearly all values of $\lambda$.

\vspace{0.08in} \noindent 
\textbf{RoCOCO.} \Cref{tab:rococo_i2t_loss,tab:rococo_t2i_loss} report image-to-text and text-to-image retrieval for each objective. On clean MS-COCO, symmetric InfoNCE achieves the highest R@1, followed by \loss{}. On RoCOCO, however, \loss{} achieves the highest R@1 and the lowest relative drop and RSMS in every setting and direction. Symmetric InfoNCE yields an RSMS $2.4$ to $3.4$ times higher for image-to-text and $1.5$ to $2.2$ times higher for text-to-image retrieval, and CentroBind, despite also aligning each modality with the anchor, $1.3$ to $3.2$ times higher.

\begin{takeaway} 
Overall, \loss{} preserves shared information comparably to symmetric InfoNCE, and yields the lowest leakage in most cases and the most robust shared representations among the contrastive objectives.
\end{takeaway}

\begin{figure*}[t]
    \centering

    \begin{subfigure}{\textwidth}
        \centering
        \includegraphics[width=0.87\linewidth]{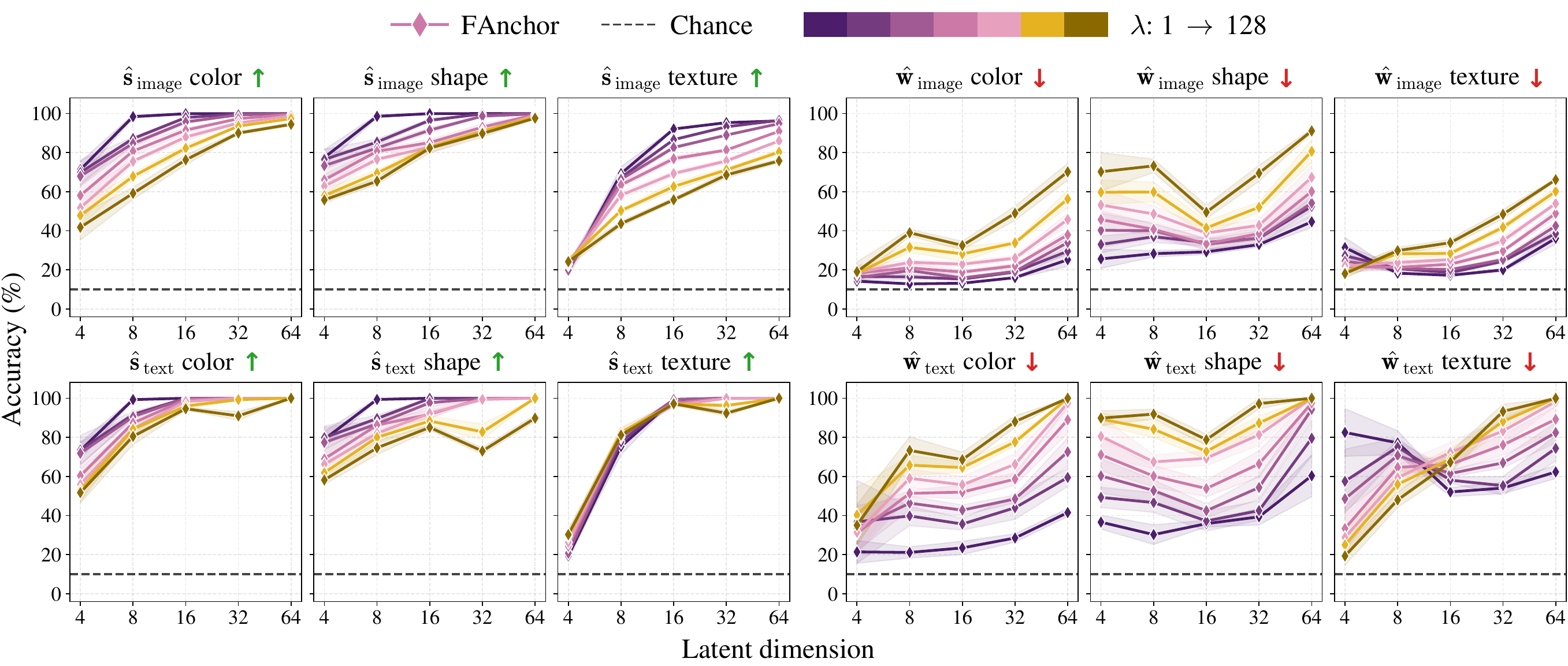}
        \caption{\textbf{(Image, Text) setting}}
        \label{fig:clevr4_image_text_all_lam}
    \end{subfigure}
    
    \caption{\textbf{Factorization on Clevr-4: $\lambda$ ablation.}
    Linear-probe accuracy (\%) for predicting color, shape, and texture
    from the shared (left) and private (right) representations across latent dimensions.
    \up/\down: higher/lower is better; dashed lines: chance (10\%).}
    \label{fig:clevr4_all_lam}
\end{figure*}

\begin{figure*}[t]
    \ContinuedFloat
    \centering

    \begin{subfigure}{0.87\textwidth}
        \centering
        \includegraphics[width=\linewidth]{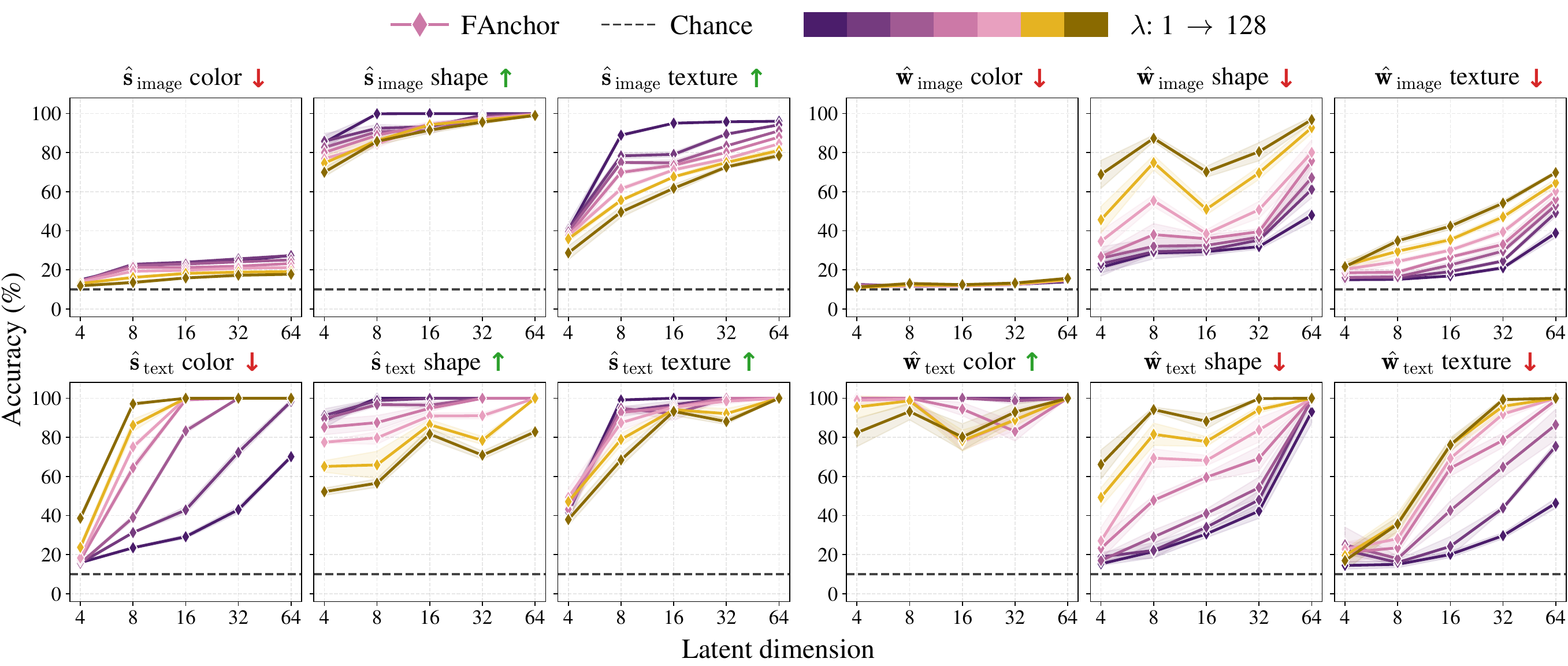}
        \caption{\textbf{(Image without color, Text) setting}}
        \label{fig:clevr4_image_no_color_text_all_lam}
    \end{subfigure}

    \begin{subfigure}{0.87\textwidth}
        \centering
        \includegraphics[width=\linewidth]{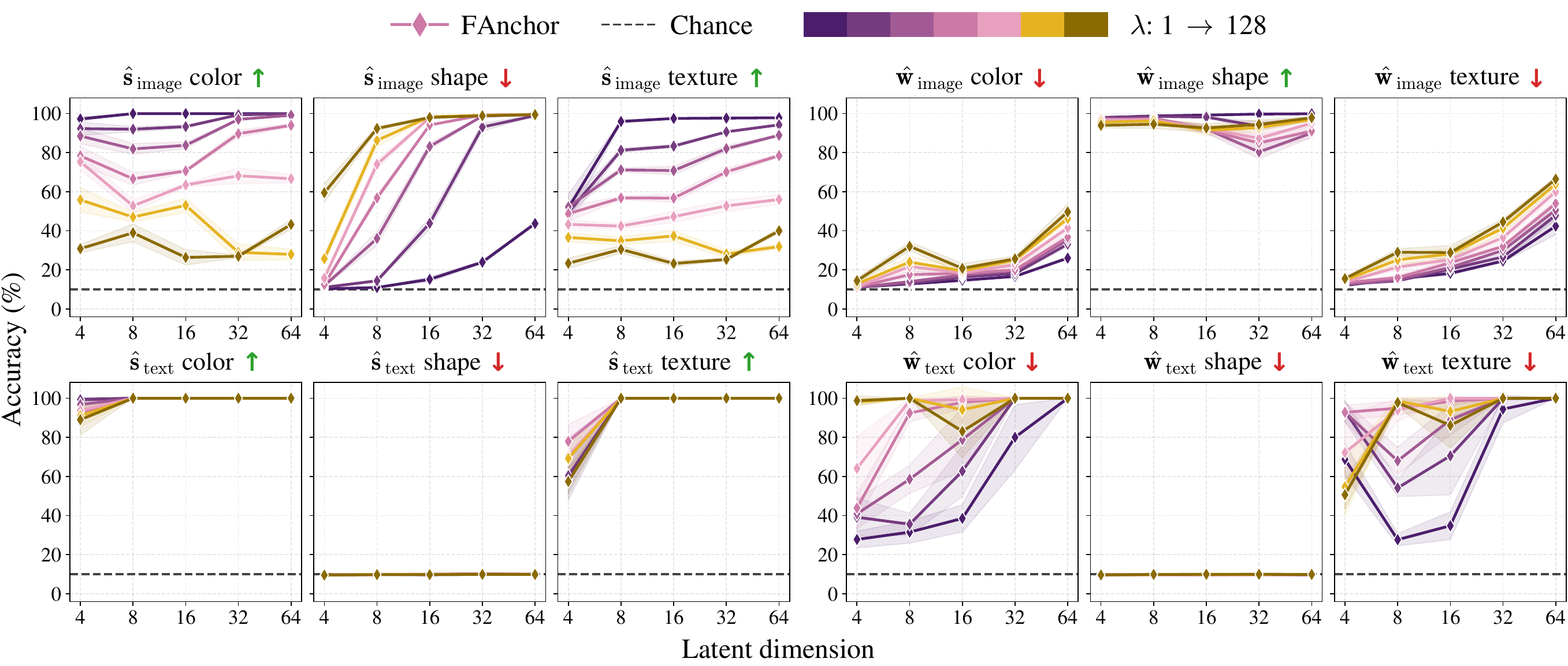}
        \caption{\textbf{(Image, Text without shape) setting}}
        \label{fig:clevr4_image_text_no_shape_all_lam}
    \end{subfigure}

    \begin{subfigure}{0.87\textwidth}
        \centering
        \includegraphics[width=\linewidth]{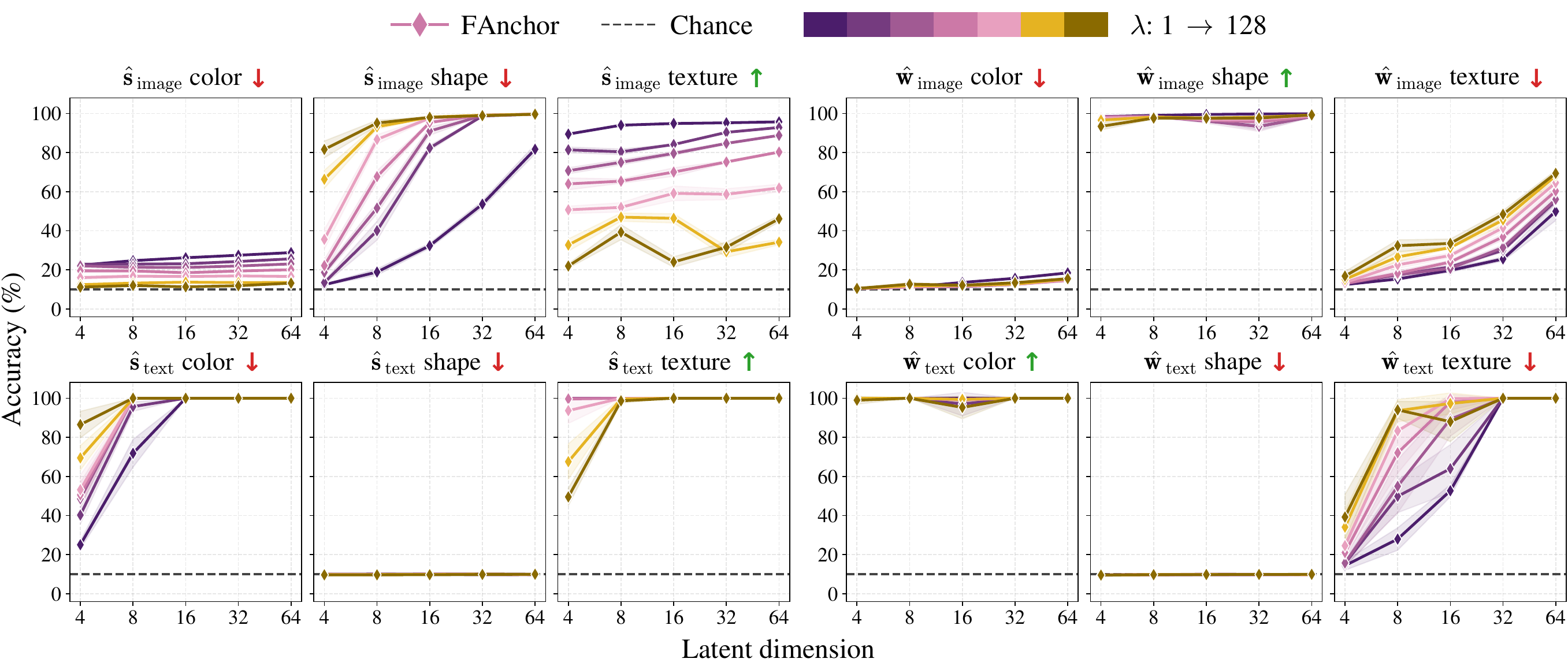}
        \caption{\textbf{(Image without color, Text without shape) setting}}
        \label{fig:clevr4_image_no_color_text_no_shape_all_lam}
    \end{subfigure}
    \caption[]{\textbf{Factorization on Clevr-4: $\lambda$ ablation (continued).}}
\end{figure*}

\subsubsection{$\lambda$ in \loss{}}

\noindent \textbf{Clevr-4.} \Cref{fig:clevr4_all_lam} reports the effect of $\lambda$, which balances alignment and uniformity in \loss{}, across the four Clevr-4 settings. Increasing $\lambda$ makes the shared representations capture more modality-specific and less shared information, \eg, in the (Image, Text without shape) setting of~\Cref{fig:clevr4_image_text_no_shape_all_lam}, $\hat{\mathbf{s}}_{\text{image}}$ encodes shape instead of color as $\lambda$ grows. Since our Matryoshka-like reconstruction objective first reconstructs each embedding from the anchor, the private representations compensate for the missing shared information, increasing shared-to-private leakage as well. The same mechanism explains the preservation--leakage diagonal on VGGSounder in~\Cref{sec:vggsounder}: since shared and modality-specific information are entangled there, the shared information absorbed by the private representations at larger $\lambda$ also improves their private-task performance.

\begin{figure*}[t]
    \centering

    \begin{subfigure}{\textwidth}
        \centering
        \includegraphics[width=0.87\linewidth]{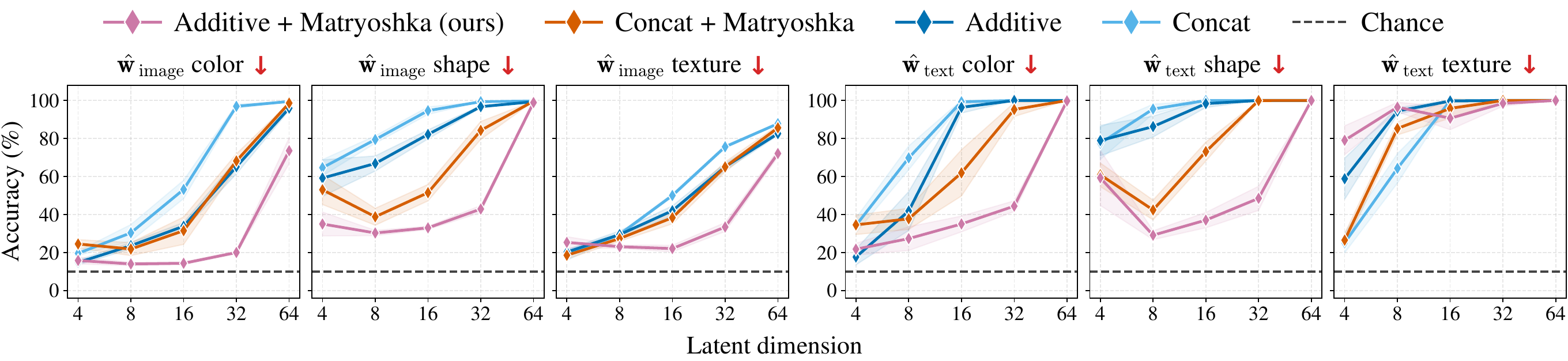}
        \caption{\textbf{(Image, Text) setting}}
        \label{fig:selected_recon_type_private_image_text}
    \end{subfigure}

    \begin{subfigure}{0.87\textwidth}
        \centering
        \includegraphics[width=\linewidth]{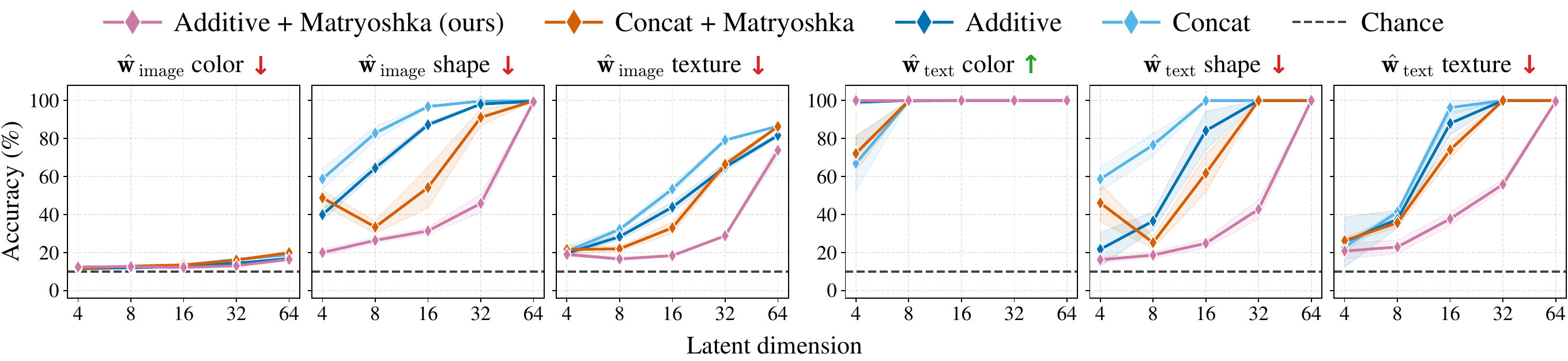}
        \caption{\textbf{(Image without color, Text) setting}}
        \label{fig:selected_recon_type_private_image_no_color_text}
    \end{subfigure}

    \begin{subfigure}{0.87\textwidth}
        \centering
        \includegraphics[width=\linewidth]{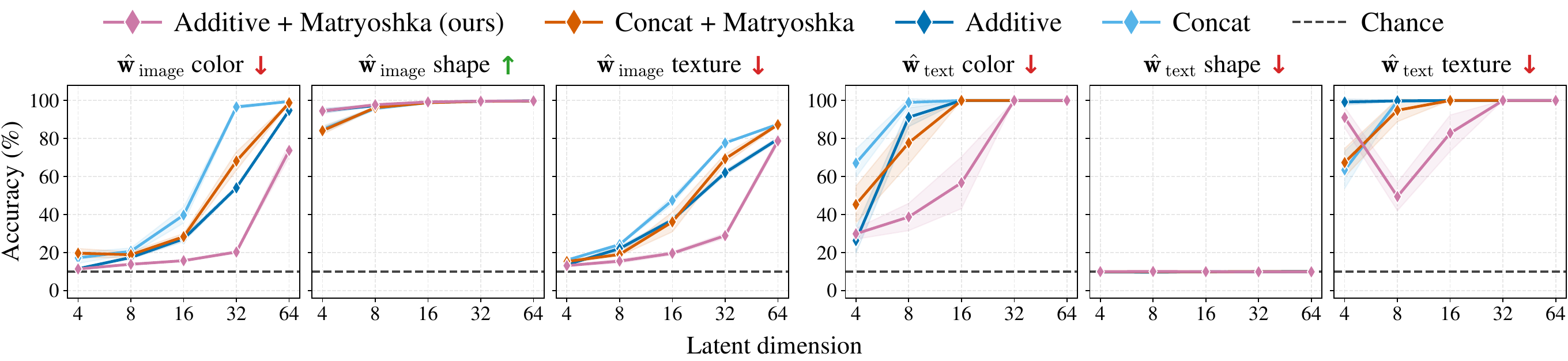}
        \caption{\textbf{(Image, Text without shape) setting}}
         \label{fig:selected_recon_type_private_image_text_no_shape}
    \end{subfigure}

    \begin{subfigure}{0.87\textwidth}
        \centering
        \includegraphics[width=\linewidth]{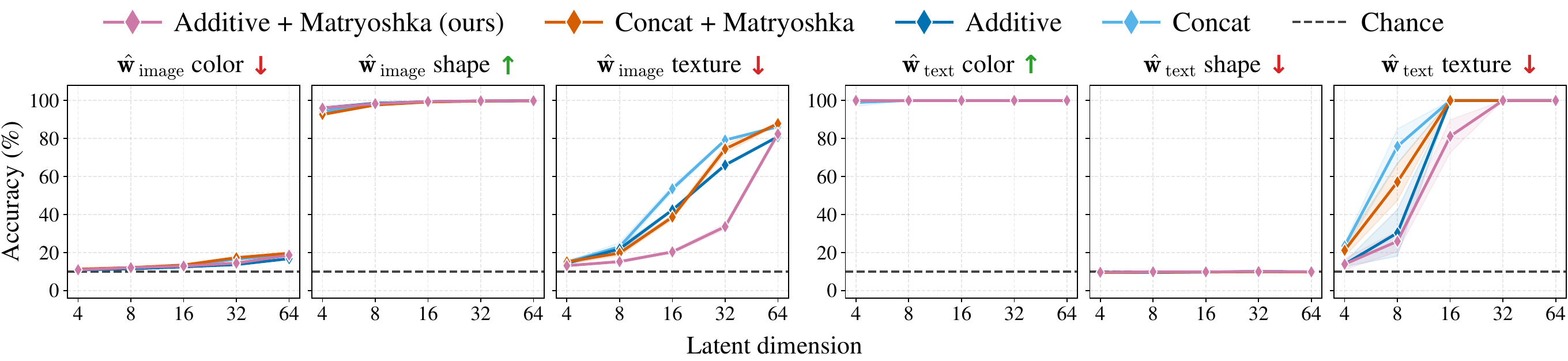}
        \caption{\textbf{(Image without color, Text without shape) setting}}
        \label{fig:selected_recon_type_private_image_no_color_text_no_shape}
    \end{subfigure}

    \caption{\textbf{Factorization on Clevr-4: Reconstruction objective ablation.} Linear-probe accuracy (\%) for predicting color, shape, and texture \emph{from private representations} across four settings and latent dimensions. \up/\down: higher/lower is better; dashed lines: chance (10\%).}
    \label{fig:clevr4_selected_recon_type_all_settings}
\end{figure*}

\subsubsection{Reconstruction objective}

We ablate the two design choices of our reconstruction objective in~\cref{eq:loss_rec}: how the anchor and the private representation are combined, and the Matryoshka-like coarse term. In all cases, we keep the shared adapters and \loss{} unchanged. We compare our additive combination in~\Cref{sec:private} with a concatenation-based one, where a single decoder takes both inputs and the coarse reconstruction is obtained by replacing the private representation with zeros:
\begin{align}
  \text{Additive:}\quad
  & \hat{\mathbf{e}}^a_m
    = \text{Decoder}^a_m\big(\text{sg}(\mathbf{a}); \phi^a_m\big),
  && \hat{\mathbf{e}}_m
    = \hat{\mathbf{e}}^a_m
    + \text{Decoder}^w_m(\hat{\mathbf{w}}_m; \phi^w_m), \\
  \text{Concat:}\quad
  & \hat{\mathbf{e}}^a_m
    = \text{Decoder}_m\big([\text{sg}(\mathbf{a}); \mathbf{0}];
      \phi_m\big),
  && \hat{\mathbf{e}}_m
    = \text{Decoder}_m\big([\text{sg}(\mathbf{a});
      \hat{\mathbf{w}}_m]; \phi_m\big).
\end{align}
Each combination is trained either with the coarse term $\|\hat{\mathbf{e}}^a_m - \mathbf{e}^{\text{tgt}}_m\|_2^2$ and the refined term $\|\hat{\mathbf{e}}_m - \mathbf{e}^{\text{tgt}}_m\|_2^2$, as in~\cref{eq:loss_rec}, or with the refined term only, yielding four variants: \emph{Additive + Matryoshka} (ours), \emph{Concat + Matryoshka}, \emph{Additive}, and \emph{Concat}.

\vspace{0.08in}
\noindent
\textbf{Clevr-4.} \Cref{fig:clevr4_selected_recon_type_all_settings} reports linear-probe accuracy on the private representations for each variant across the four Clevr-4 settings. All variants capture the private attributes comparably, but ours exhibits the lowest shared-to-private leakage in nearly all cases. Specifically, the Matryoshka-like objective generally reduces leakage with either combination, and the additive combination leaks less than concatenation regardless of the objective, showing that both components contribute. Intuitively, the coarse term forces the anchor decoder to explain as much of the embedding as possible, leaving only the residual modality-specific information to the private representation, and the additive combination helps decouple their contributions, in line with prior work showing that additive decoders can identify blocks of latents under certain conditions~\citep{lachapelle2023additive}.

\begin{figure*}[t]
    \centering
    \includegraphics[width=0.87\textwidth]{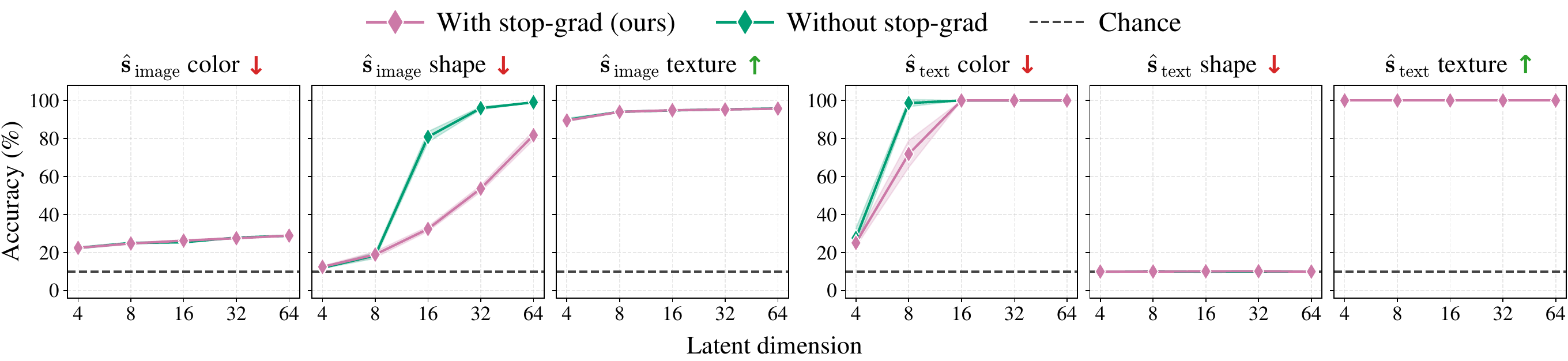}
    \caption{\textbf{Factorization on Clevr-4 (Image without color, Text
    without shape): Stop-gradient on the anchor ablation.} Linear-probe accuracy
    (\%) for predicting color, shape, and texture \emph{from shared
    representations}. \up/\down: higher/lower is better; dashed lines:
    chance (10\%).}
    \label{fig:clevr4_stop_grad_shared_image_no_color_text_no_shape}
\end{figure*}

\subsubsection{Stop-gradient on the anchor}

\noindent
\textbf{Clevr-4.} In~\cref{eq:decoder_a}, we apply a stop-gradient to the anchor so that the shared adapters are trained only with \loss{}. We ablate this choice by letting the reconstruction objective in~\cref{eq:loss_rec} also update the shared adapters. \Cref{fig:clevr4_stop_grad_shared_image_no_color_text_no_shape} reports linear-probe accuracy on the shared representations in the (Image without color, Text without shape) setting. Both variants preserve texture equally well, but removing the stop-gradient substantially increases private-to-shared leakage, \eg, shape is predicted from $\hat{\mathbf{s}}_{\text{image}}$ with $80.8\%$ accuracy at dimension $16$, compared with only $32.4\%$ with it. Hence, the stop-gradient prevents the reconstruction objective from introducing private information into the shared representations.

\begin{figure*}[t]
    \centering
    \includegraphics[width=0.87\textwidth]{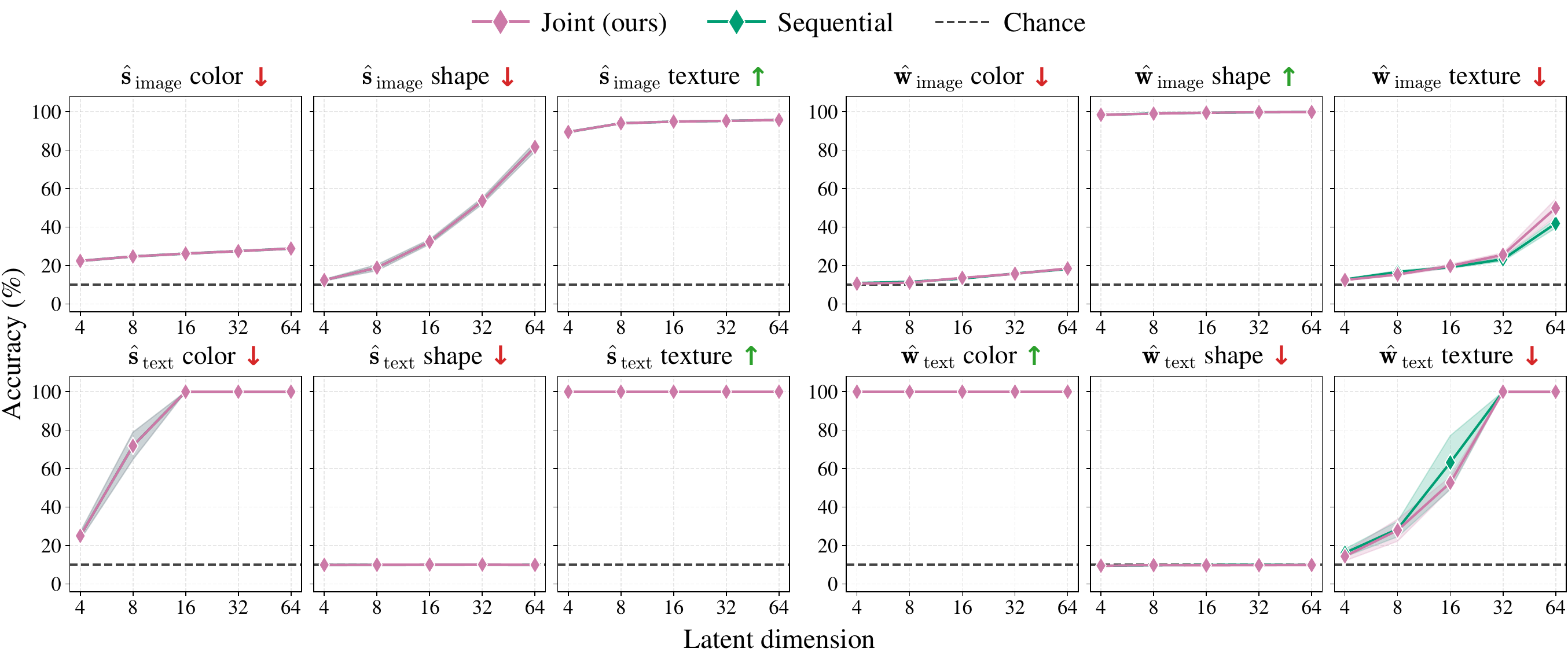}
    \caption{\textbf{Factorization on Clevr-4 (Image without color, Text
    without shape): Joint versus sequential training ablation.} Linear-probe accuracy
    (\%) for predicting color, shape, and texture from
    the shared (left) and private (right) representations. \up/\down: higher/lower is better; dashed lines:
    chance (10\%).}
    \label{fig:ablation_two_stage_image_no_color_text_no_shape}
\end{figure*}

\begin{figure}[t]
    \centering
    \includegraphics[width=0.5\textwidth]{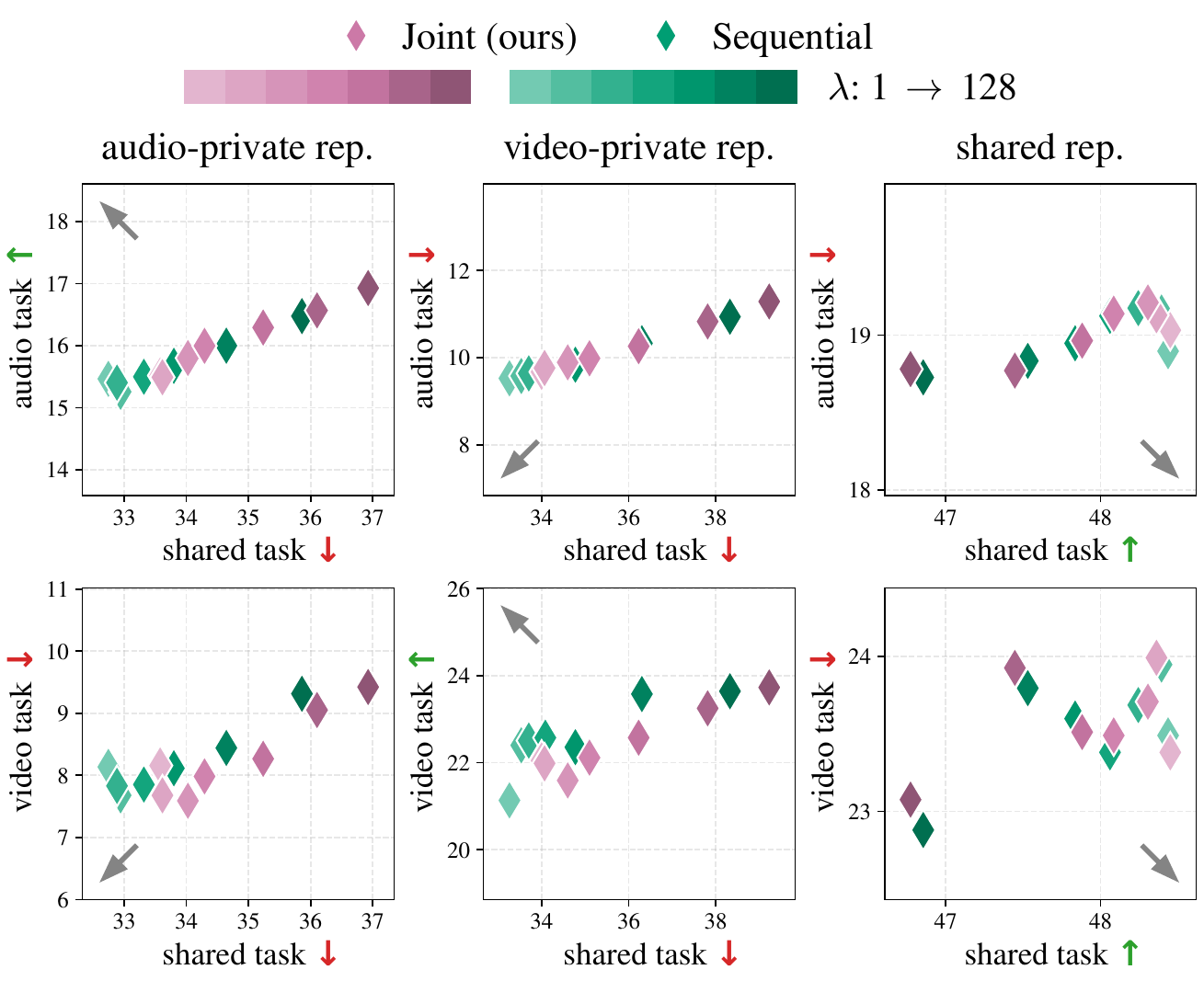}
    \caption{\textbf{Factorization on VGGSounder: Joint versus sequential training ablation.} Linear-probe F1@1 (\%) on shared and private tasks using audio-private (left), video-private (middle), and shared (right) representations. Each point corresponds to one hyperparameter configuration; color gradients indicate $\lambda$ values from $1$ to $128$ for both schemes. \up/\down: higher/lower is better; \optarrow: optimal regions.}
    \label{fig:two_stage_f1_at1_disenFalse}
\end{figure}

\begin{figure*}[t]
    \centering
    \includegraphics[width=0.55\textwidth]{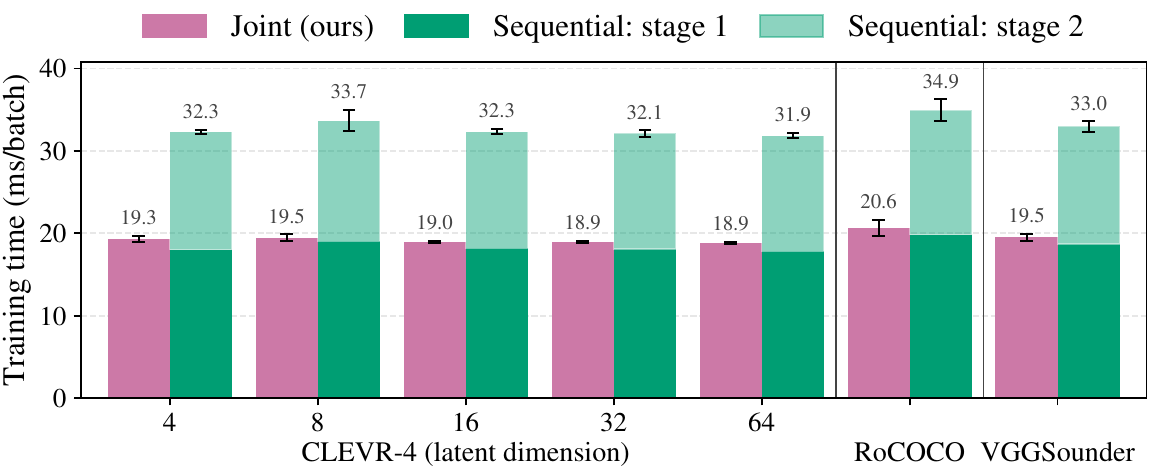}
    \caption{\textbf{Training time (ms/batch) of joint versus sequential training.} Mean over 5 different seeds on Clevr-4 across latent dimensions, RoCOCO, and VGGSounder; error bars show the standard deviation. Sequential bars stack stage~1 (shared) and stage~2 (private), each trained for the same number of epochs as joint training. Numbers above bars: total time.}
    \label{fig:compute_cost_joint_vs_sequential}
\end{figure*}

\subsubsection{Joint versus sequential training} 

With the stop-gradient in~\cref{eq:decoder_a}, the shared adapters receive gradients only from \loss{}, even when all components are trained jointly via~\cref{eq:general_loss}. As an alternative, sequential training first trains the shared adapters, then freezes them and trains the private adapters and decoders, each stage for the same number of epochs.

\vspace{0.08in}
\noindent
\textbf{Clevr-4.} In the (Image without color, Text without shape) setting, \Cref{fig:ablation_two_stage_image_no_color_text_no_shape} shows nearly identical shared and private representations for both schemes. This is expected for the shared adapters, which receive the same gradients in both.

\vspace{0.08in}
\noindent
\textbf{VGGSounder.} As shown in~\Cref{fig:two_stage_f1_at1_disenFalse}, the shared representations are again nearly identical, whereas the sequential scheme yields slightly lower shared-to-private leakage in the private representations. This may be attributed to the private adapters being optimized while the anchor is already trained and fixed.

\vspace{0.08in}
\noindent
\textbf{Training time.} \Cref{fig:compute_cost_joint_vs_sequential} shows that the sequential scheme takes $1.6$ to $1.8$ times longer than joint training across all datasets, as it runs two full training schedules instead of one. The inference cost is identical for both schemes. Hence, joint training achieves comparable factorization quality at a lower training cost.

\end{document}